%% file: neurips_2022.tex
\documentclass{article}

\usepackage[final]{neurips_2022}
\usepackage{enumitem}
\usepackage{tabularx}
\usepackage{multirow}
\usepackage{pifont}
\usepackage{xcolor}
\usepackage{tikz}
\usepackage{wrapfig}

\newcommand{\cmark}{\textcolor{green!80!black}{\ding{51}}}
\newcommand{\xmark}{\textcolor{red}{\ding{55}}}

\usepackage[utf8]{inputenc} 
\usepackage[T1]{fontenc}    
\usepackage{hyperref}       
\usepackage{url}            
\usepackage{booktabs}       
\usepackage{amsfonts}       
\usepackage{nicefrac}       
\usepackage{microtype}      
\usepackage{xcolor}         
\usepackage{amsmath}
\usepackage{graphicx}
\usepackage[english]{babel}
\usepackage{amssymb}
\usepackage{amsthm}
\usepackage{bbm} 
\usepackage{tabularx}
\usepackage{makecell}

\newtheorem{theorem}{Theorem}
\newtheorem{corollary}{Corollary}

\definecolor{darkpastelgreen}{rgb}{0.01, 0.75, 0.24}
\definecolor{cobalt}{rgb}{0.0, 0.28, 0.67}

\title{JAWS: Auditing Predictive Uncertainty Under Covariate Shift}

\author{%
  Drew Prinster \\
  Department of Computer Science\\
  Johns Hopkins University\\
  Baltimore, MD 21211 \\
  \texttt{drew@cs.jhu.edu} \\
   \And
   Anqi Liu \\
   Department of Computer Science \\
   Johns Hopkins University\\
  Baltimore, MD 21211 \\
  \texttt{aliu@cs.jhu.edu} \\
   \And
   Suchi Saria \\
   Department of Computer Science \\
   Johns Hopkins University\\
  Baltimore, MD 21211 \\
  \texttt{ssaria@cs.jhu.edu} \\
}

\begin{document}

\maketitle
\begin{abstract}
We propose \textbf{JAWS}, a series of wrapper methods for distribution-free uncertainty quantification tasks under covariate shift, centered on the core method \textbf{JAW}, the \textbf{JA}ckknife+ \textbf{W}eighted with data-dependent likelihood-ratio weights. JAWS also includes computationally efficient \textbf{A}pproximations of JAW using higher-order influence functions: \textbf{JAWA}. Theoretically, we show that JAW relaxes the jackknife+'s assumption of data exchangeability to achieve the same finite-sample coverage guarantee even under covariate shift. JAWA further approaches the JAW guarantee in the limit of the sample size or the influence function order under common regularity assumptions. Moreover, we propose a general approach to repurposing predictive interval-generating methods and their guarantees to the reverse task: estimating the probability that a prediction is erroneous, based on user-specified error criteria such as a safe or acceptable tolerance threshold around the true label. We then propose \textbf{JAW-E} and \textbf{JAWA-E} as the repurposed proposed methods for this \textbf{E}rror assessment task. Practically, JAWS outperform state-of-the-art predictive inference baselines in a variety of biased real world data sets for interval-generation and error-assessment predictive uncertainty auditing tasks.




\end{abstract}



\section{Introduction}
\label{sec:intro}


\textbf{Auditing the uncertainty under data shift} Principled quantification of predictive uncertainty is crucial for enabling users to calibrate how much they should or should not trust a given prediction \citep{thiebes2021trustworthy, ghosh2021uncertainty, tomsett2020rapid, bhatt2021uncertainty}. Uncertainty-based predictor auditing can be considered a type of uncertainty quantification performed \textit{post-hoc}, for example by a regulator without detailed knowledge of a predictor's architecture and with limited resources \citep{schulam2019can}. Data shift poses a major challenge to uncertainty quantification due to violation of the common assumption that the training and test data are exchangeable, or more specifically independent and identically distributed (i.i.d.) \citep{ovadia2019can,ulmer2020trust,zhou2021amortized,chan2020unlabelled}. Therefore, it is essential to develop convenient tools for users or regulators to audit the uncertainty of a given prediction even when training data is biased.


\textbf{Predictor auditing: Interval generation} In this work we distinguish between two types of predictive uncertainty auditing. We describe the first type as \textit{interval generation}, which refers to a common goal in the distribution-free uncertainty quantification literature: to generate a predictive confidence interval (or set) that covers the true label with at least a user-specified probability. For instance, an auditor might ask for predictive intervals that contain the true label with at least, say 90\% frequency.

\textbf{Predictor auditing: Error assessment} While predictive interval generation has been a central focus of the distribution-free uncertainty quantification literature \citep{angelopoulos2021gentle}, in some applications the reverse computation may be more actionable: estimating the probability that a prediction is erroneous or not, based on user-specified error critieria such as a safe or acceptable tolerance region around the true label. We thus refer to this task as \textit{error assessment}.
For instance, take the setting of chemical or radiation therapy dose prediction for cancer treatment, where administering a dose within approximately $\pm 10\%$ of the optimal dose is considered safety-critical (see Appendix \ref{app:dose_example} for details). Whereas predictive interval generation could fail to provide safety assurance (e.g., if the predictive confidence interval is larger than the safe tolerance region), error assessment would give a worst-case probability of the prediction being safe. Similar examples could be formulated in other applications, such as incision planning in surgical robotics and autonomous vehicle navigation.

\textbf{Coverage} We assume a standard regression setup with a multiset of training data $\{(X_1, Y_1), ..., (X_n, Y_n)\}$ and a test point $(X_{n+1}, Y_{n+1})$ with unknown label $Y_{n+1}$, where $(X_i, Y_i) \in \mathbb{R}^d \times \mathbb{R}$ for all $i \in \{1, ..., n+1\}$. Also, we denote a predictor as $\widehat{\mu} = \mathcal{A}(\{(X_1, Y_1), ..., (X_n, Y_n)\})$, where $\mathcal{A}$ is a model-fitting algorithm. For a predictive interval (or set) $\widehat{C}_{n, \alpha}^{\text{audit}} : \mathbb{R}^d \rightarrow \{\text{subsets of } \mathbb{R}\}$, a \textit{coverage guarantee} gives a lower bound to the probability that the interval covers the true test label: 
\begin{align}
    \mathbb{P}\big\{Y_{n+1} \in \widehat{C}_{n, \alpha}^{\text{audit}}(X_{n+1})\big\} \geq 1 - \alpha.
    \label{eq:coverage_intro}
\end{align}
The coverage guarantee provides the basis for both interval-generation and error-assessment auditing, though it is important to note that in this work we focus on marginal rather than conditional coverage (see \citep{foygel2021limits} for more details on this distinction). Standard conformal prediction methods \citep{vovk2005algorithmic,shafer2008tutorial,vovk2013transductive} along with the jackknife+ and related methods \citep{barber2021predictive}, which we refer to together as ``predictive inference'' methods, provide a framework for generating predictive intervals with finite-sample guaranteed coverage. 

\textbf{Exchangeability} Standard conformal prediction and the jackknife+ rely on two crucial notions of \textit{exchangeability}: data exchangeability, that is that the training and test data are all exchangeable (e.g., i.i.d.); and secondly that the model-fitting algorithm $\mathcal{A}$ treats the data symmetrically \citep{barber2022conformal}. In common situations of dataset shift, however, the data exchangeability assumption is violated. Empirically, the coverage performance of standard conformal prediction methods can suffer under data shift \citep{tibshirani2019conformal, podkopaev2021distribution}. 

%
\begin{figure}[ht]
\centering
\begin{tabular}{cc}
     \includegraphics[width=0.4\textwidth]{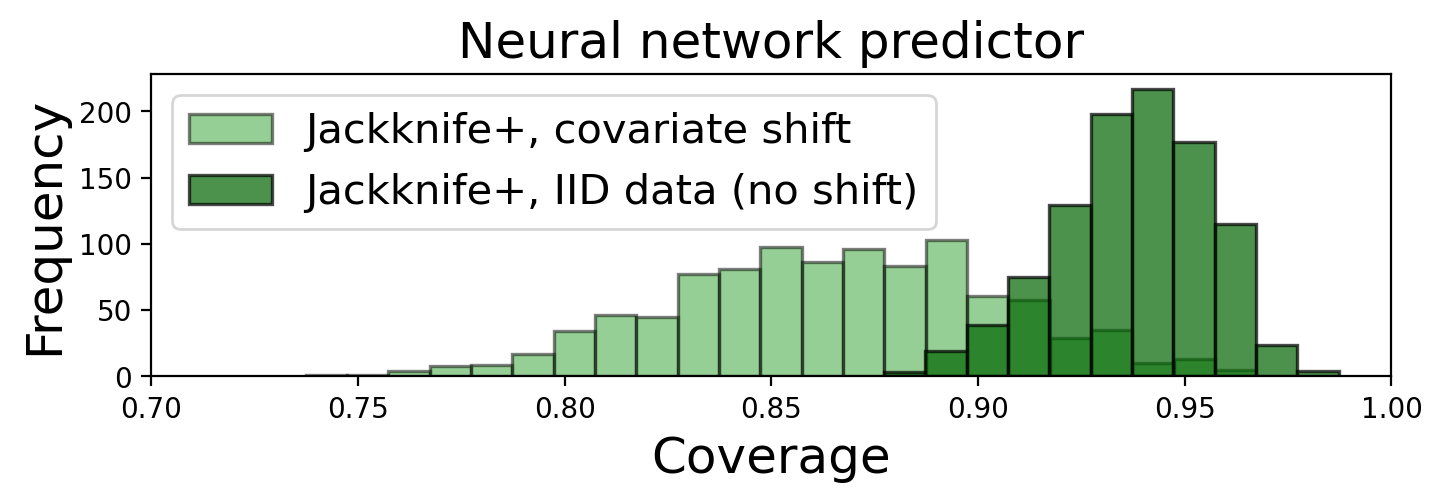} & 
      \includegraphics[width=0.4\textwidth]{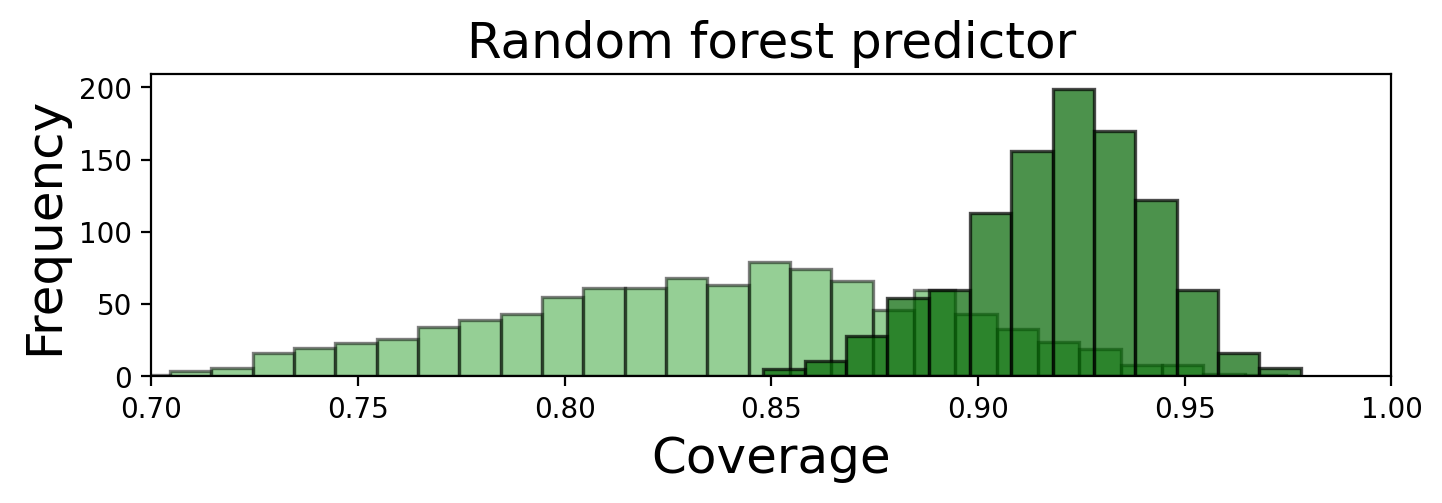}
\end{tabular}
\caption{Jackknife+ loses coverage on the airfoil dataset under covariate shift (details in Section \ref{sec:exps}).}
    \label{Fig:loss_cover}
\end{figure}


In this work, we build on the jackknife+ method due to its beneficial compromise between the statistical and computational limitations of other conformal prediction methods \citep{barber2021predictive}. However, jackknife+ coverage performance can still degrade under data shift, such as shown in Figure \ref{Fig:loss_cover}, and in some applications its computational requirements can still be limiting. To address these concerns and make extensions to error assessment, we develop JAWS, a series of wrapper methods for distribution-free uncertainty quantification under covariate shift (see Table \ref{tab:properties} for key properties).

\begin{table}[ht]
\caption{Summary of key properties for JAWS methods (details in Section \ref{sec:ProposedApproach}).}
\label{tab:properties}
\centering
\begin{tabular}{|c | c | c | c | c| } 
\hline
&  & \multicolumn{2}{c|}{\textbf{Guarantee (under covariate shift)}} &   \\
\cline{3-4}
\textbf{Method} & \textbf{Task} & \textbf{Finite sample} & \textbf{Asymptotic} & \textbf{Avoids retraining}  \\ 
\hline
JAW & Interval generation & \cmark & \cmark & \xmark \\
\hline
JAWA & Interval generation & \xmark & \cmark & \cmark \\ 
\hline
JAW-E & Error assessment & \cmark & \cmark & \xmark \\
\hline
JAWA-E & Error assessment & \xmark & \cmark & \cmark \\ 
\hline
\end{tabular}
\end{table}

\textbf{Our contributions can be summarized as follows}:
\begin{enumerate}[leftmargin=13pt,topsep=0pt,itemsep=0mm]
    \item We develop JAW: a jackknife+ method with data-dependent likelihood-ratio weights for predictive interval generation under covariate shift. We show that JAW achieves the same rigorous, finite-sample coverage guarantee as jackknife+ \citep{barber2021predictive} while relaxing the data exchangeability assumption to allow for covariate shift.
    \item We develop JAWA: a sequence of computationally efficient approximations to JAW that uses higher-order influence functions to avoid retraining. Under assumptions outlined in \cite{giordano2019higher} regarding the regularity of the data, Hessian of the objective (local strong convexity), and the existence and boundedness of higher order derivatives, we provide an asymptotic guarantee for the JAWA coverage in the limit of the sample size or influence function order. 
    \item We propose a general approach to repurposing any distribution-free predictive inference method to the error assessment task, with rigorous guarantees for the coverage probability estimation. Our approach applies to methods that assume exchangeable data and to methods like JAW and JAWA that allow for covariate shift---JAW-E and JAWA-E refer to the error assessment versions.
    \item We demonstrate superior empirical performance of JAWS over other distribution-free predictive inference baselines on a variety of benchmark datasets under covariate shift.
\end{enumerate}



\section{Background and related work}


\subsection{Standard conformal prediction}
\label{intro:conformal_pred}

Conformal prediction has grown into a broad research field since arising in the 1990s \citep{vovk2005algorithmic, shafer2008tutorial, balasubramanian2014conformal, angelopoulos2021gentle}. Standard conformal prediction methods generate a prediction interval (or set) with a finite-sample coverage guarantee as in \eqref{eq:coverage_intro}, which is \textit{distribution-free} in the sense that the guarantee applies to any exchangeable data distribution \citep{lei2014distribution,lei2018distribution}. With the exchangeability assumptions in Section \ref{sec:intro}, standard conformal prediction methods rely on a pre-fit score function $\widehat{S}: \mathbb{R}^d \times \mathbb{R}\rightarrow \mathbb{R}$ (in regression, the absolute-value residual score $\widehat{S}(x, y) = |y - \widehat{\mu}(x)|$ is commonly used). A conformal prediction interval at confidence level $1-\alpha$ is then determined by a corresponding quantile on a multiset of (exchangeable) score values.

Split conformal and full conformal are two main types of standard conformal prediction, and each bears its own limitation \citep{vovk2005algorithmic,shafer2008tutorial}. Split conformal generates scores on labeled holdout data and is computationally efficient due to not requiring retraining, but sample splitting to obtain the holdout set can reduce model accuracy \citep{papadopoulos2008inductive,lei2018distribution,vovk2012conditional}. On the other hand, full conformal prediction avoids the holdout set requirement, but at the heavy computational cost of retraining the model on every possible target value (or, in practice, on a fine grid of target values) \citep{ndiaye2019computing,zeni2020conformal}. 

\subsection{Covariate shift}


Under the \textit{covariate shift} assumption, the $Y | X$ distribution is assumed to be the same between training and test data but the marginal $X$ distributions may change \citep{sugiyama2007covariate,shimodaira2000improving}:
\begin{align}
    (X_i, Y_i) \stackrel{\text{i.i.d.}}{\sim} P_X \times P_{Y|X}, i = 1, ..., n ; \qquad (X_{n+1}, Y_{n+1}) \sim \tilde{P}_X \times P_{Y|X}, \ \text{independently}.
    \label{eq:cs}
\end{align}
Rich literature exist in this domain---see Appendix \ref{app:cov_shift_background} for more details. Uncertainty quantification is relatively less explored under covariate shift, though recent work \citep{ovadia2019can,zhou2021amortized,chan2020unlabelled} emphasizes its importance, especially in deep learning.

\subsection{Conformal prediction under covariate shift and beyond exchangeability}

\label{sec:weighted_conformal}

\cite{tibshirani2019conformal} develop the idea of \textit{weighted exchangeability} for adapting conformal prediction to the covariate shift setting. Random variables $V_1, ..., V_n$ are weighted exchangeable with weight functions $w_1, ..., w_n$ if their joint density $f$ can be factorized as $f(v_1, ..., v_n) = \prod_{i=1}^nw_i(v_i) \cdot g(v_1, ..., v_n)$, where $g$ is independent of ordering on its inputs. For covariate shift as in \eqref{eq:cs}, if $\tilde{P}_X$ is absolutely continuous with respect to $P_X$, then the data $\{(X_i, Y_i)\}$ are weighted exchangeable with weights given by the likelihood ratio $w(X_i) = \text{d}\tilde{P}_X(X_i) / \text{d}P_X(X_i)$ \citep{tibshirani2019conformal}.

If $\{v_i\}$ represents a set of scores for standard conformal prediction, then we can represent the empirical distribution of $\{v_i\}$ as $\frac{1}{n+1}\sum_{i=1}^n\delta_{v_i} + \frac{1}{n+1}\delta_{\infty}$, where $\delta_{v_i}$ denotes a point mass at $v_i$ \citep{barber2022conformal}. By extension, weighted conformal prediction uses the \textit{weighted} empirical distribution defined as $\sum_{i=1}^np_i^w(x)\delta_{v_i} + p_{n+1}^w(x)\delta_{\infty}$, with weights given by
\begin{align}
    p_i^w(x) = \frac{w(X_i)}{\sum_{j=1}^nw(X_j) + w(x)}, i = 1, ..., n  ; \qquad \text{and} \qquad p_{n+1}^w(x) = \frac{w(x)}{\sum_{j=1}^nw(X_j) + w(x)},
    \label{eq:weights}
\end{align}
where $p_i^w(x)$ can be thought of as a normalized likelihood ratio weight for each $i \in \{1, ..., n+1\}$. Corollary 1 in \citep{tibshirani2019conformal} provides the coverage guarantee of weighted conformal prediction that takes the form of \eqref{eq:coverage_intro} but relaxes the exchangeable data assumption to allow for covariate shift. However, weighted split and weighted full conformal inherit the same statistical and computational limitations, respectively, from their standard (exchangeable) variants.

The recent work of \citep{barber2022conformal} provides a novel extension of conformal prediction and the jackknife+ to unknown violations of the exchangeability assumption, including a ``nonexchangeable jackknife+'' defined with fixed weights. The key difference between the nonexchangeable jackknife+ in \cite{barber2022conformal} and our proposed JAW method is that \cite{barber2022conformal} use fixed weights to compensate for unknown exchangeability violations (not limited to covariate shift) but at the expense of a bounded but generally nonzero “coverage gap” (drop in guaranteed coverage relative to if the data were exchangeable), whereas our JAW method with data-dependent weights assumes covariate shift but does not suffer from any similar coverage gap. See Appendix \ref{app:comparison_barber_etal_2022} for more details.

\subsection{Jackknife+} 
The jackknife+ \citep{barber2021predictive}, which is closely related to cross conformal prediction \citep{vovk2018cross}, offers a compromise between the statistical limitation of split conformal and the computational limitation of full conformal, at the cost of a slightly weaker coverage guarantee.
The jackknife+ predictive interval can most easily be understood as a modification to a predictive interval from the classic jackknife resampling method \citep{miller1974jackknife,steinberger2018conditional,steinberger2016leave}. For a set of point masses $\{\delta_{v_i}\}$ at values $v_1, ..., v_n$, let  $Q^-_{\beta}\{\frac{1}{n+1}\delta_{v_i}\}$ denote the level $\beta$ quantile on the empirical distribution $\sum_{i=1}^n\frac{1}{n+1}\delta_{v_i} + \frac{1}{n+1}\delta_{-\infty}$ and let  $Q^+_{\beta}\{\frac{1}{n+1}\delta_{v_i}\}$ denote the level $\beta$ quantile on the empirical distribution $\sum_{i=1}^n\frac{1}{n+1}\delta_{v_i} + \frac{1}{n+1}\delta_{\infty}$.
Then, denoting the model trained without the $i$th point as $\widehat{\mu}_{-i} = \mathcal{A}\big(\big\{(X_1, Y_1), ..., (X_{i-1}, Y_{i-1}), (X_{i+1}, Y_{i+1}), ..., (X_n, Y_n)\big\}\big)$ and the leave-one-out residual $R_i^{LOO} = |Y_i - \widehat{\mu}_{-i}(X_i)|$, the jackknife prediction interval can be written as
\begin{align}
    \widehat{C}_{n, \alpha}^{\text{jackknife}}(X_{n+1}) & = \Big[Q^-_{\alpha}\Big\{\tfrac{1}{n+1}\delta_{\widehat{\mu}(X_{n+1})- R_i^{LOO}}\Big\}, \ Q^+_{1-\alpha}\Big\{\tfrac{1}{n+1}\delta_{\widehat{\mu}(X_{n+1}) + R_i^{LOO}}\Big\}\Big].
\label{eq:jackknife}
\end{align}
In contrast, we obtain the jackknife+ predictive interval in \cite{barber2021predictive} by replacing the full model prediction $\widehat{\mu}(X_{n+1})$ in \eqref{eq:jackknife} with $\widehat{\mu}_{-i}(X_{n+1})$:
\begin{align}
    \widehat{C}_{n, \alpha}^{\text{jackknife+}}(X_{n+1}) = \Big[Q^-_{\alpha}\Big\{\tfrac{1}{n+1}\delta_{\widehat{\mu}_{-i}(X_{n+1})- R_i^{LOO}}\Big\}, \ Q^+_{1-\alpha}\Big\{\tfrac{1}{n+1}\delta_{\widehat{\mu}_{-i}(X_{n+1}) + R_i^{LOO}}\Big\}\Big].
    \label{eq:jackknife+}
\end{align}
\citep{barber2021predictive} prove that, with the same exchangeability assumptions as in standard conformal prediction, the jackknife+ prediction interval satisfies 
\begin{align}
    \mathbb{P}\{Y_{n+1} \in \widehat{C}_{n, \alpha}^{\text{jackknife+}}(X_{n+1})\} \geq 1 - 2\alpha.
    \label{eq:jackknife_coverage}
\end{align}

\subsection{Approximating leave-one-out models with higher-order influence functions}

Influence functions (IFs) \citep{cook1977detection} have a long history in robust statistics for estimating the dependence of parameters on sample data. Recently, IFs have become more widespread in machine learning for uses including model interpretability \cite{koh2017understanding} and approximating classic resampling-based uncertainty quantification methods including bootstrap \citep{schulam2019can}, jackknife, and leave-$k$-out cross validation \citep{giordano2019swiss, giordano2019higher}. In each of these cases, IFs enable approximation of the parameters that would be obtained if the model were retrained on resampled data by instead estimating the effect of a corresponding reweighting. In prior work, \cite{alaa2020discriminative} proposed approximating the leave-one-out models required by the jackknife+ with higher-order IFs, but their work assumes exchangeable or i.i.d. train and test data.

Let $\hat{\theta}$ denote the fitted parameters for predictor $\widehat{\mu}$ trained on the full training data.
Given Assumptions 1-4 in \cite{giordano2019higher}---which require that $\hat{\theta}$ is a local minimum of the objective function, that the objective is $k+1$ times continuously differentiable with bounded norms, and that the objective is strongly convex in the neighborhood of $\hat{\theta}$---then the $k$-th order leave-one-out IF refers to the $k$-th order directional derivative of the model parameters $\hat{\theta}$ with respect to the data weights, in the direction of the leave-one-out change in weights (See \ref{app:IFs_background} for more details).
With each of these $k$th order leave-one-out IFs for $k \in \{1, ..., K\}$, denoted with condensed notation $\delta^k_{-i} \hat{\theta}$, we can construct a $K$-th order Taylor series approximation to estimated the leave-one-out model parameters $\hat{\theta}_{-i}\ $:
\begin{align}\hat{\theta}^{\text{IF-}K}_{-i} := \hat{\theta} + \sum_{k=1}^{K}\frac{1}{k!}\delta^k_{-i} \hat{\theta}.
\label{eq:kth_order}
\end{align}
In this work we implement the algorithm proposed by \cite{giordano2019higher} to compute higher-order IFs, a recursive procedure based on foreward-mode automatic differentiation \citep{maclaurin2015autograd} for memory efficiency in computing higher-order directional derivatives. Our introduction of IFs is highly simplified---we refer to Appendix \ref{app:IFs_background} and to \cite{giordano2019higher} for more details. 

\subsection{Error assessment}
Whereas conformal prediction and related methods generate prediction intervals that control the error probability (miscoverage level $\alpha$) at a user-specified level, we refer to the reverse task as \textit{error assessment}: estimating the probability that a prediction is erroneous or not, based on user-specified error criteria. For instance, a user might define an error as any deviation between the prediction $\widehat{\mu}(X_{n+1})$ and the true label $Y_{n+1}$ greater than some acceptable tolerance threshold $\tau$: that is, when $|Y_{n+1} - \widehat{\mu}(X_{n+1})| > \tau$. In Section \ref{sec:error_shift}, we present a general approach to repurposing predictive inference methods with validity under covariate shift to error assessment.

We note that for score functions that are monotonic in $y$, such as $\widehat{S}(x, y) = y - \widehat{\mu}(x)$, guarantees for this error assessment task can be obtained using conformal predictive \textit{distributions} as described by \cite{vovk2017nonparametric} (also see \cite{vovk2020computationally, vovk2018conformal, xie2022homeostasis}). In regression tasks assuming exchangeable data, CPDs generate a probability distribution for the label over $\mathbb{R}$. However, CPDs require that score functions be monotonic in $y$, whereas we allow for certain non-monotone conformity scores such as the commonly used absolute-value residual $|y - \widehat{\mu}(x)|$; Moreover, CPDs assume exchangeable data, whereas our approach extends to covariate shift.

\section{Proposed approach and theoretical results}
\label{sec:ProposedApproach}




\subsection{JAW: Jackknife+ weighted with data-dependent weights}
\label{sec:coverage_theory}


We present \textbf{JAW}, the \textbf{JA}ckknife+ \textbf{W}eighted with data-dependent likelihood-ratio weights, defined by the following predictive interval:
\begin{align} 
    \widehat{C}_{n, \alpha}^{\text{JAW}}(X_{n+1}) = \Big[ & Q^-_{\alpha}\big\{p_i^w(X_{n+1})\cdot\delta_{\widehat{\mu}_{-i}(X_{n+1})- R_i^{LOO}}\big\}, \ Q^+_{1-\alpha}\big\{p_i^w(X_{n+1})\cdot\delta_{\widehat{\mu}_{-i}(X_{n+1}) + R_i^{LOO}}\big\}\Big],
    \label{JAW}
\end{align}
where $R_i^{LOO} = \big|\widehat{\mu}_{-i}(X_i) - Y_i\big|$, where the $p_i^w(x)$ are as defined in \eqref{eq:weights}, where $Q^-_{\alpha}\big\{p_i^w(X_{n+1})\cdot\delta_{\widehat{\mu}_{-i}(X_{n+1}) - R_i^{LOO}}\big\}$ denotes the level $\alpha$ quantile of the empirical distribution $\sum_{i=1}^n\big[p_i^w(X_{n+1})\cdot\delta_{\widehat{\mu}_{-i}(X_{n+1}) - R_i^{LOO}}\big] + p_{n+1}^w(X_{n+1})\cdot\delta_{-\infty}$, and where $Q^+_{1-\alpha}\big\{p_i^w(X_{n+1})\cdot\delta_{\widehat{\mu}_{-i}(X_{n+1}) + R_i^{LOO}}\big\}$ is the level $1-\alpha$ quantile for $\sum_{i=1}^n\big[p_i^w(X_{n+1})\cdot\delta_{\widehat{\mu}_{-i}(X_{n+1}) + R_i^{LOO}}\big] + p_{n+1}^w(X_{n+1})\cdot\delta_{\infty}$. 

We show that $\widehat{C}_{n, \alpha}^{\text{JAW}}(X_{n+1})$ satisfies the same coverage guarantee as the jackknife+ except relaxing the data exchangeability assumption to allow for covariate shift, which we state formally as follows:
\begin{theorem}\textit{Assume data under covariate shift from \eqref{eq:cs}. If $\tilde{P}_X$ is absolutely continuous with respect to $P_X$, then the JAW interval in \eqref{JAW} satisfies}
\label{thm:1}
\begin{align}
    \mathbb{P}\big\{Y_{n+1} \in \widehat{C}_{n, \alpha}^{\text{JAW}}(X_{n+1})\big\} \geq 1 - 2\alpha
\end{align}
\end{theorem}

\textbf{Remark 1.} The results from \cite{tibshirani2019conformal} do not directly imply Theorem \ref{thm:1}. Applying the approach from \cite{tibshirani2019conformal} to the jackknife+ entails treating $\widehat{\mu}_{-i}(X_{n+1}) \pm R_i^{LOO}$ as implicit nonconformity scores. \cite{tibshirani2019conformal} assume that the nonconformity scores for the training data $\{V_1, ..., V_n\}$ are weighted exchangeable with the nonconformity score for the test point $V_{n+1}$. But, observe that for $i\in \{1, ..., n\}$, $\widehat{\mu}_{-i}$ is trained on $n-1$ datapoints, whereas $\widehat{\mu}_{-(n+1)} = \widehat{\mu}$ is trained on $n$ datapoints. Thus, no reweighting can make $\widehat{\mu}_{-i}$ equivalent to $\widehat{\mu}$ in distribution, and therefore $\widehat{\mu}_{-i}(X_{n+1}) \pm R_i^{LOO}$ and $\widehat{\mu}(X_{n+1}) \pm R_{n+1}^{LOO}$ are not weighted exchangeable.

\textbf{Proof sketch:} Our proof technique for Theorem \ref{thm:1} extends the jackknife+ coverage guarantee proof in \cite{barber2021predictive} to the covariate shift setting for JAW using likelihood ratio weights as in \cite{tibshirani2019conformal}. The outline is as follows:

\begin{enumerate}[leftmargin=13pt,topsep=0pt]
    \item[] \textit{Setup}: Following \cite{barber2021predictive}, we define a set of leave-\textit{two}-out models $\{\tilde{\mu}_{-(i, j)}\}$. We then generalize the notion of ``strange'' points described in \cite{barber2021predictive} to covariate shift. 
    \item \textit{Bounding the total normalized weight of strange points}: We establish deterministically that the total normalized weight of strange points cannot exceed $2\alpha$.
    \item \textit{Weighted exchangeability using the leave-two-out models}: Using the leave-two-out model construction, we leverage weighted exchangeability to show that the probability that a test point $n+1$ is strange is thus bounded by $2\alpha$. 
    \item \textit{Connection to JAW}: Lastly, we show that the JAW interval can only fail to cover the test label value $Y_{n+1}$ if $n + 1$ is a strange point.
    
\end{enumerate}

While JAW assumes access to oracle likelihood ratio weights, in practice this information often has to be estimated. See Appendix \ref{app:jaw_e} for a discussion and experiments of JAW with estimated weights. 

\subsection{JAWA: Using higher-order influence functions to approximate JAW without retraining}

For computationally efficient JAW \textbf{A}pproximations that avoid retraining $n$ leave-one-out models, we propose the JAWA sequence, which approximates the leave-one-out models required by JAW using higher-order influence functions. For each training point $i \in \{1, ..., n\}$, define the $K$-th order influence function approximation to the leave-one-out refit parameters $\hat{\theta}_{-i}$, obtained from Algorithm 4 in \cite{giordano2019higher}, as given by equation \eqref{eq:kth_order}, and let $\widehat{\mu}_{-i}^{\text{IF-}K}$ be the model with with these approximated parameters $\hat{\theta}^{\text{IF-}K}_{-i}$ for each $i \in \{1, ..., n\}$. Then, the prediction interval for the $K$-th order JAWA (i.e., for JAWA-$K$) is given by 
\begin{align}
    \widehat{C}_{n, \alpha}^{\text{JAWA-}K}(X_{n+1}) = \Big[& Q^-_{\alpha}\big\{p_i^w(X_{n+1})\cdot\delta_{\widehat{\mu}_{-i}^{\text{IF-}K}(X_{n+1})- R_i^{\text{IF-}K, LOO}}\big\}, \nonumber \\ 
    & Q^+_{1-\alpha}\big\{p_i^w(X_{n+1})\cdot\delta_{\widehat{\mu}_{-i}^{\text{IF-}K}(X_{n+1}) + R_i^{\text{IF-}K, LOO}}\big\}\Big],
    \label{JAWA}
\end{align}

with $R_i^{\text{IF-}K, LOO} = \big|\widehat{\mu}_{-i}^{\text{IF-}K}(X_i) - Y_i\big|$, $p_i^w(x)$ as in \eqref{eq:weights}, and quantiles defined analogously to JAW.

We now provide an asymptotic coverage guarantee for $\widehat{C}_{n, \alpha}^{\text{JAWA-}K}(X_{n+1})$ that holds either in the limit of the sample size or in the limit of the influence function order, under regularity conditions formally described in \cite{giordano2019higher}. These assumptions concern the regularity and continuity of the training data, local convexity of the objective (or that the Hessian is strongly positive definite), and the existence and boundedness of the objective's 1st through $K+1$th order directional derivatives.

\begin{theorem}\textit{Assume data under covariate shift from \eqref{eq:cs} and that $\tilde{P}_X$ is absolutely continuous with respect to $P_X$. Let Assumptions 1 - 4 and either Condition 2 or Condition 4 from \cite{giordano2019higher} hold uniformly for all $n$. Then, in the limit of the training sample size $n \rightarrow \infty$ or in the limit of the influence function order $K \rightarrow \infty$, the JAWA-$K$ interval in \eqref{JAWA} satisfies}
\label{thm2}
\begin{align}
    \mathbb{P}\big\{Y_{n+1} \in \widehat{C}_{n, \alpha}^{\text{JAWA-}K}(X_{n+1})\big\} \geq 1 - 2\alpha.
\end{align}
\end{theorem}

We leave the proof to Appendix \ref{prf:2}, but we note that the result follows by combining Propositions 1 and 3 in \cite{giordano2019higher} with the JAW coverage guarantee that we present in Theorem \ref{thm:1}.

\subsection{Error assessment under covariate shift}
\label{sec:error_shift}

\begin{figure}[hbt]
\begin{center}
\includegraphics[width=0.8\textwidth]{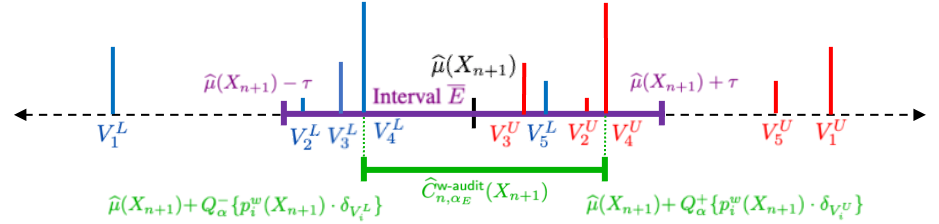}
\end{center}
\caption{Illustration of approach to repurposing a predictive inference method ``w-audit'' to error assessment. The interval $\overline{E} = [\widehat{\mu}(X_{n+1})-\tau, \ \widehat{\mu}(X_{n+1})+\tau]$ is shown in violet, the lower score values $\{V_i^L\}$ in blue, the upper score values $\{V_i^U\}$ in red, and the interval $\widehat{C}_{n, \alpha_E}^{\text{w-audit}}(X_{n+1})$ in green. Each vertical line at a location $V_i$ on the real line represents a point mass $\delta_{V_i}$ with height corresponding to the normalized likelihood ratio weight $p_i^w(X_{n+1})$.} 
\label{fig:thm4}
\end{figure}

We now propose a general approach to repurposing predictive inference methods with validity under covariate shift from predictive interval generation to the reverse task: estimating the probability that a prediction is erroneous or not, based on user-specified error criteria. 
For example, consider a user that defines a prediction $\widehat{\mu}(X_{n+1})$ as erroneous, relative to the true label $Y_{n+1}$, if it is farther than some acceptable tolerance threshold $\tau$ from $Y_{n+1}$: i.e., if $|Y_{n+1} - \widehat{\mu}(X_{n+1})| > \tau$. For this common regression error criterion, our approach to adapting a method such as JAW \eqref{JAW} or weighted split conformal prediction \citep{tibshirani2019conformal} to error assessment reduces to first defining the set of labels that would \textit{not} be considered erroneous, $\overline{E} = [\widehat{\mu}(X_{n+1})-\tau, \ \widehat{\mu}(X_{n+1})+\tau]$, and then finding the method's largest predictive interval contained within $\overline{E}$, call it $\widehat{C}_{n, \alpha_E}^{\text{w-audit}}(X_{n+1})$. The coverage guarantee for $\widehat{C}_{n, \alpha_E}^{\text{w-audit}}(X_{n+1})$ then yields a lower bound on $\mathbb{P}\{Y_{n+1}\in \overline{E}\}$, the probability of no error (or an upper bound on the error probability). See Figure \ref{fig:thm4} for an illustration of this example.

More generally, for a prediction $\widehat{\mu}(X_{n+1})$, a user must specify error criteria by a test score function $\widehat{S}: \mathbb{R}^d\times \mathbb{R}\rightarrow \mathbb{R}$, as well as minimum and maximum acceptable score values $\tau^-$ and $\tau^+$, assuming $\tau^- < \tau^+$ without loss of generality; in other words, $\widehat{\mu}(X_{n+1})$ is considered erroneous if $\widehat{S}(X_{n+1}, Y_{n+1}) < \tau^-$ or if $\tau^+ < \widehat{S}(X_{n+1}, Y_{n+1})$. (Note that if $\widehat{S}$ is nonnegative, we might let $\tau^-=0$.)
Then, the values of $y$ for which observing $Y_{n+1}=y$ would \textit{not} imply $\widehat{\mu}(X_{n+1})$ is erroneous are:
\begin{align}
    \overline{E} = \big\{y \in \mathbb{R} : \tau^- \leq \widehat{S}(X_{n+1}, y) \leq \tau^+\big\}.
    \label{eq:error_event}
\end{align}
Now, assume a predictive inference method with predictive sets that can be written in the form
\begin{align}
    \widehat{C}_{n, \alpha}^{\text{w-audit}}(X_{n+1}) =  \big\{y \in \mathbb{R} : \ & \widehat{Q}_{\alpha}^-\{p_i^w(X_{n+1})\delta_{V_i^L}\} \leq \widehat{S}(X_{n+1}, y) \leq \widehat{Q}_{1-\alpha}^+\{p_i^w(X_{n+1})\delta_{V_i^U}\}\big\}
    \label{eq:error_method_interval}
\end{align}
with valid coverage guaranteed under covariate shift.
(Note, \eqref{eq:error_method_interval} gives the JAW interval \eqref{JAW} by setting $\widehat{S}(x, y) = y - \widehat{\mu}(x)$, $V_i^L = \widehat{\mu}_{-i}(X_{n+1}) - \widehat{\mu}(X_{n+1}) - R_i^{LOO}$, and $V_i^U = \widehat{\mu}_{-i}(X_{n+1}) - \widehat{\mu}(X_{n+1}) + R_i^{LOO}$; see Appendix \ref{app:JAW_R_guarantee}. Similarly, \eqref{eq:error_method_interval} gives the prediction interval for weighted split conformal prediction \citep{tibshirani2019conformal} for absolute value residual scores when $\widehat{S}(x, y) = |y - \widehat{\mu}(x)|$, and for all calibration data $i$ we let $V_i^U = |Y_i - \widehat{\mu}(X_i)|$ and $V_i^L = 0$.) Then, defining
\begin{align}
    \alpha_{E}^{\text{w-audit}} = \min\Big(\Big\{\alpha' \ : \ \tau^- \leq \ \widehat{Q}_{\alpha'}^-\{p_i^w(X_{n+1})\delta_{V_i^L}\} \ ,\ \widehat{Q}_{1-\alpha'}^+\{p_i^w(X_{n+1})\delta_{V_i^U}\} \leq \tau^+  \Big\}\Big),
\label{eq:alpha_I-CovShift}
\end{align}
we can estimate the probability of $\widehat{\mu}(X_{n+1})$ \textit{not} resulting in an error as in \eqref{eq:error_event} as:
\begin{align}
    \widehat{p}\{Y_{n+1} \in \overline{E}\} = 
    \begin{cases} 1 - \alpha_{E}^{\text{w-audit}} & \text{ if } \alpha_{E}^{\text{w-audit}} \text{ exists} \\
    0 & \text{otherwise}.
    \end{cases}
\label{eq:thm4}
\end{align}
While the target coverage for $\widehat{C}_{n, \alpha_E}^{\text{w-audit}}(X_{n+1})$ is used in \eqref{eq:thm4}, the following theorem gives the worst-case error assessment guarantee for covariate shift (proof in Appendix \ref{prf:4}). Corollary \ref{cor1} in Appendix \ref{app:JAW_R_guarantee} and Corollary \ref{cor2} in Appendix \ref{app:JAWA_R_guarantee} give the error assessment guarantees for JAW-E and JAWA-E respectively. Theorem \ref{thm3} in Appendix \ref{app:error_assess} gives the analogous guarantee for exchangeable data.


\begin{theorem}\textit{Assume a predictive inference method of the form \eqref{eq:error_method_interval} has coverage guarantee $\mathbb{P}\{Y_{n+1} \in \widehat{C}_{n, \alpha}^{\text{w-audit}}(X_{n+1})\} \geq 1 - c_1\alpha - c_2 \ $, with $c_1, c_2 \in \mathbb{R}$, under covariate shift \eqref{eq:cs} where $\tilde{P}_X$ is absolutely continuous with respect to $P_X$. Define $\overline{E}$ as in \eqref{eq:error_event} and $\alpha_{E}^{\text{w-audit}}$ as in \eqref{eq:alpha_I-CovShift}. Then, 
}
\begin{align}
    \mathbb{P}\{Y_{n+1} \in \overline{E}\} \geq 
    \begin{cases} 1 - c_1\alpha_{E}^{\text{w-audit}} - c_2 & \text{ if } \alpha_{E}^{\text{w-audit}} \text{ exists and }\ \alpha_{E}^{\text{w-audit}} < \frac{1 - c_2}{c_1} \\
    0 & \text{otherwise}
    \end{cases}.
\end{align}
\label{thm4}
\end{theorem}
\section[Experiments]{Experiments\footnote{Additional experimental analysis and a link to code for all experiments and included in Appendices \ref{app:analysis} and \ref{app:details}}}
\label{sec:exps}

\subsection{Datasets and creation of covariate shift }

We conduct experiments on five UCI datasets \cite{Dua:2019} with various dimensionality (Table \ref{tab:stats}): airfoil self-noise, red wine quality prediction \citep{cortez2009modeling}, wave energy converters, superconductivity \citep{hamidieh2018data}, and communities and crime \citep{redmond2002data}.

\begin{table}[ht]
\caption{Statistics for the UCI datasets. Only the first 2000 samples were used for the wave and superconductivity datasets (for wave, the first 2000 samples of Adelaide data).}
\label{tab:stats}
\centering
\begin{tabular}{|c | c| c | c | } 
\hline
\textbf{Dataset} & \textbf{\# of samples} & \textbf{\# of features} & \textbf{label range} \\ 
\hline
Airfoil self-noise (airfoil) & 1503 &  5 & [103.38, 140.987] \\ 
\hline
Red wine quality (wine) & 1599 &  11 & [3, 8] \\ 
\hline
Wave energy converters (wave) & 2000 & 48 & [1226969, 1449349] \\
\hline
Superconductivity (superconduct) & 2000 & 81 & [0.2, 136.0] \\
\hline
Communities and crime (communities) & 1994 & 99 & [0, 1] \\
\hline
\end{tabular}
\end{table}

We use exponential tilting to induce covariate shift on the test data, based on the approach used in \cite{tibshirani2019conformal}. We first randomly sample 200 points for the training data, and then sample the biased test data from the remaining datapoints that are not used for training with probabilities proportional to exponential tilting weights. See Appendix \ref{app:cov_shift_details} for additional details.

\subsection{Baselines}

\textbf{Baselines for comparison to JAW} We compared JAW to the following baselines:

\begin{enumerate}[leftmargin=13pt,topsep=0pt,itemsep=0mm]
    \item \textbf{Naive} estimates are based on training data residuals $|Y_{i} -\widehat{\mu}(X_{i})|$, which suffers from overfitting.
    \item \textbf{Jackknife} uses the classic Jackknife resampling as in \eqref{eq:jackknife}.
    \item \textbf{Jackknife+} follows \eqref{eq:jackknife+}, which replaces the prediction $\widehat{\mu}(X_{n+1})$ in jackknife with $\widehat{\mu}_{-i}(X_{n+1})$.
    \item \textbf{Jackknife-mm} is a more conservative alternative to the jackknife+ method that guarantees coverage at the $1-\alpha$ level with exchangeable data, but usually with overly-wide intervals.
   {\small \begin{align}
    \widehat{C}_{n, \alpha}^{\text{jackknife-mm}}(X_{n+1}) &  = \Big[\min_{i = 1,...,n} \widehat{\mu}_{-i}(X_{n+1})- Q^+_{1-\alpha}\{R_i^{LOO}\}, \max_{i = 1,...,n}\widehat{\mu}_{-i}(X_{n+1}) + Q^+_{1-\alpha}\{R_i^{LOO}\}\Big] \nonumber
\end{align}}
    \item \textbf{Cross validation+} (CV+) is similar to jackknife+ except splits data into $k$ portions and replaces the $\widehat{\mu}_{-i}(X_{n+1})$ with $\widehat{\mu}_{-k}(X_{n+1})$, the model trained with the $k$th subset removed. 
    \item \textbf{Split} method follows split conformal prediction, which uses half the data for training and the other half for generating the nonconformity scores.
    \item \textbf{Weighted split} is a version of split conformal with likelihood ratio weights to maintain coverage under covariate shift, as in \cite{tibshirani2019conformal}.
\end{enumerate}

\textbf{Baselines for comparison to JAWA} For influence function orders $K \in \{1, 2, 3\}$, we compared the proposed JAWA-$K$ method with $K$-th order influence function approximations of the jackknife-based baselines that we used as comparisons to JAW---we thus refer to these approximations as IF-$K$ jackknife, IF-$K$ jackknife+, and IF-$K$ jackknife-mm. Each baseline compared to JAWA-$K$ is thus also approximated with the same $K$-th order leave-one-out influence function models.

\subsection{Experimental results} 

We report experimental results on the predictive interval-generation task for both JAW and JAWA and on the error assessment task for JAW, compared to baselines. Additional experimental details and supplementary experiments can be found in Appendix \ref{app:analysis}, including for estimated likelihood ratio weights in \ref{app:jaw_e}, ablation study with shift magnitudes in \ref{app:ablation}, and coverage histograms in \ref{app:histogram_coverage_comparison}.

\subsubsection{Interval generation results for JAW: Coverage and interval width} 

\begin{figure}[ht]
    \setlength{\tabcolsep}{0pt}
    \begin{tabular}{ccccc}
    \multicolumn{5}{c}{\includegraphics[width=0.9\textwidth]{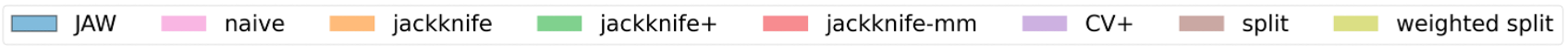}} \\
    \includegraphics[width=0.21\textwidth]{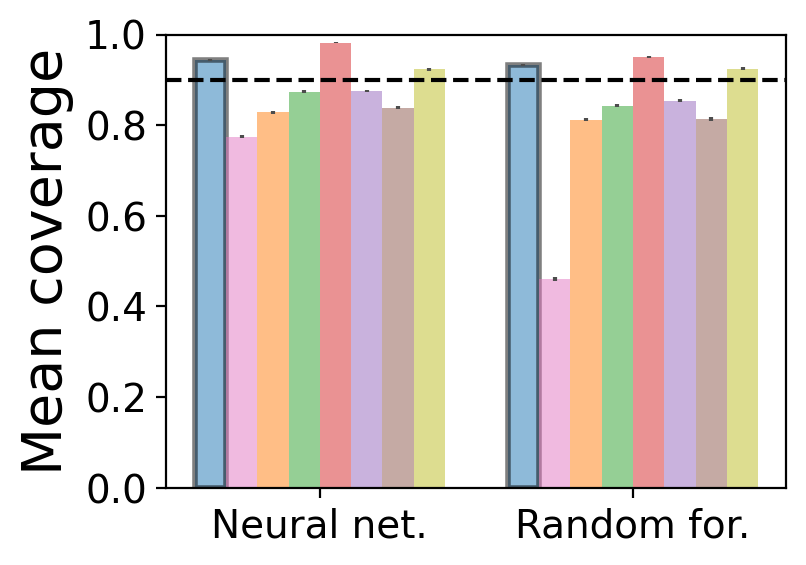}&  \includegraphics[width=0.195\textwidth]{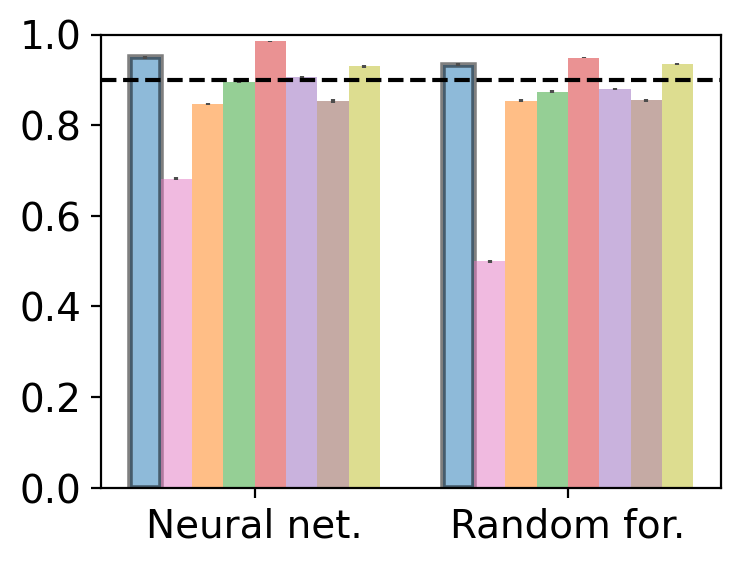} & \includegraphics[width=0.195\textwidth]{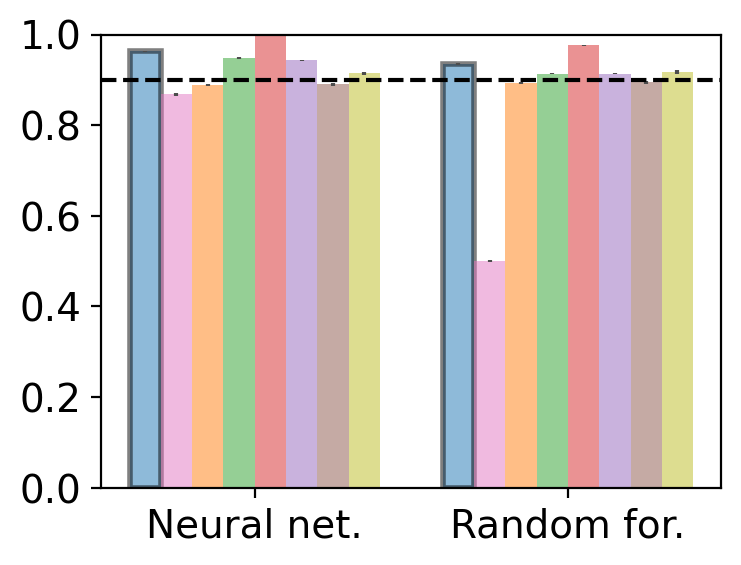} & \includegraphics[width=0.195\textwidth]{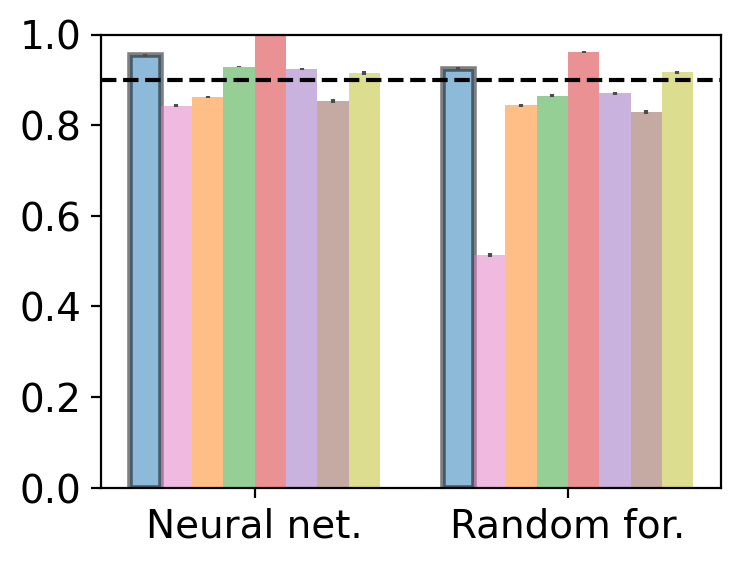} &  \includegraphics[width=0.195\textwidth]{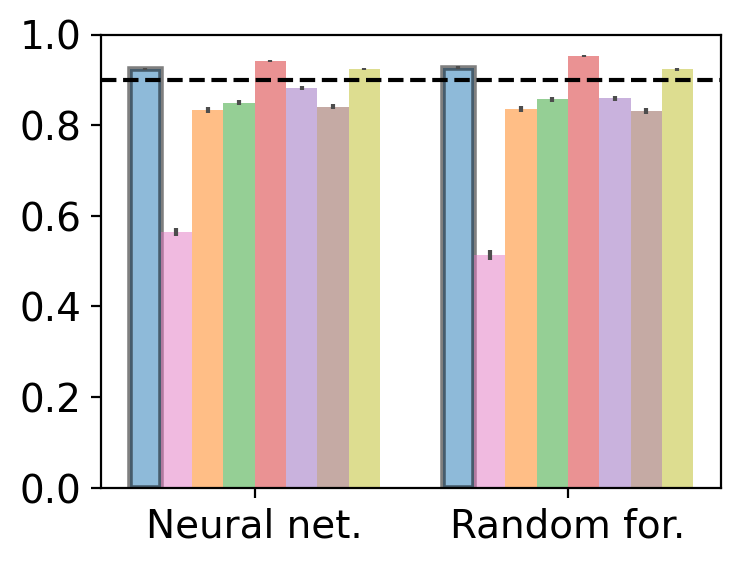} \\
    \includegraphics[width=0.22\textwidth]{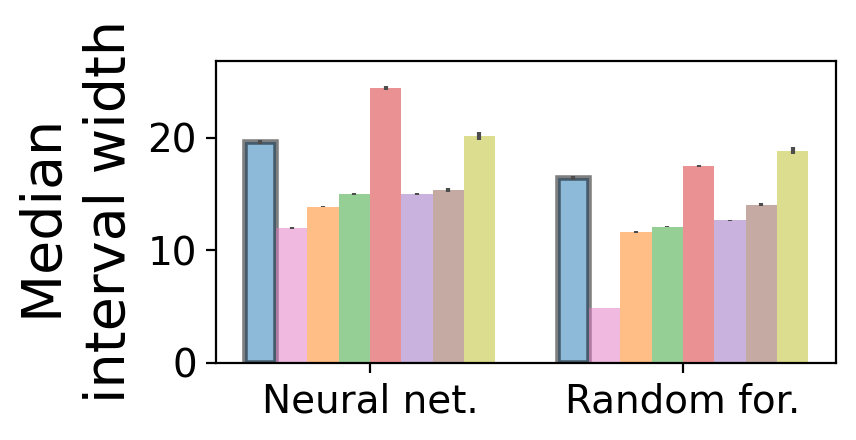} & \includegraphics[width=0.185\textwidth]{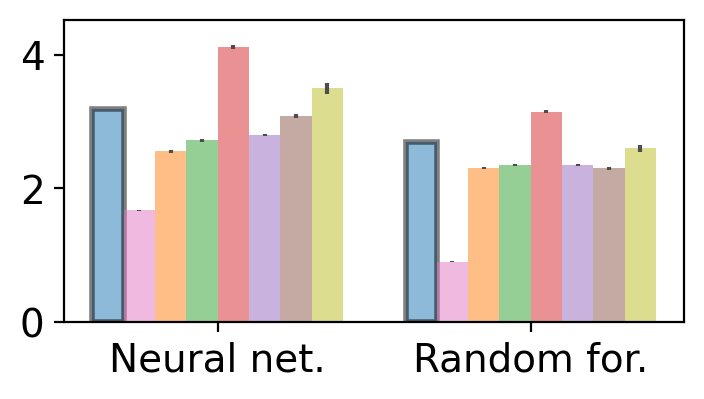} & \includegraphics[width=0.212\textwidth]{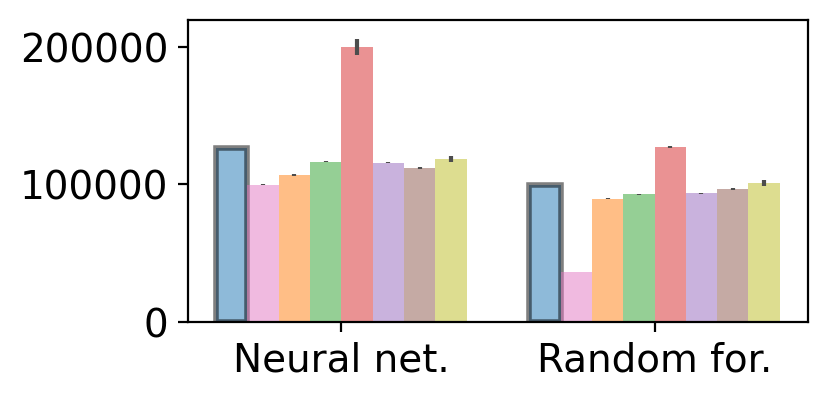} & \includegraphics[width=0.188\textwidth]{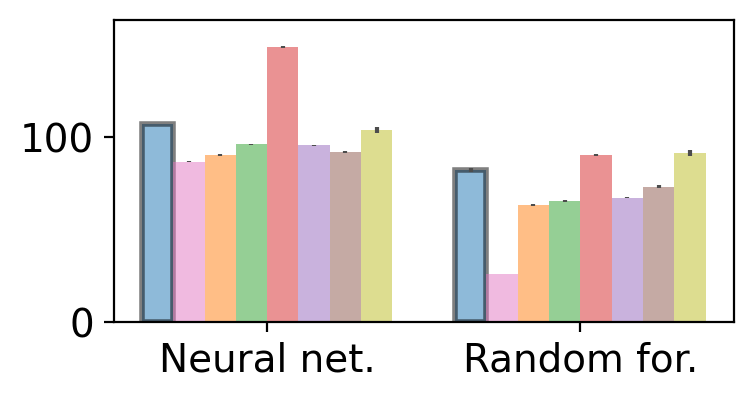} & \includegraphics[width=0.188\textwidth]{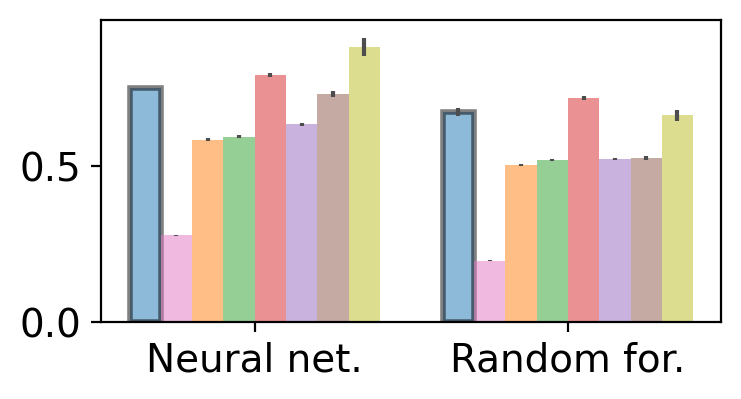} \\
    \footnotesize (a) Airfoil & (b) Wine & (c) Wave & (d) Superconduct & (e) Communities
    \end{tabular}
    \caption{Mean coverage (first row) and median interval width (second row) for neural network and random forest predictors on UCI datasets. Dashed line is the target coverage level ($1-\alpha$ = 0.9). Error bars show the standard error of 1000 repeated experiments. JAW maintains target coverage under covariate shift for all predictor and dataset conditions along with jackknife-mm and weighted split---however, JAW's intervals are generally smaller and thus more informative than jackknife-mm's, and JAW's coverage variance is smaller and thus more reliable than weighted split's (Appendix \ref{app:coverage_variance_comparison}).}
    \label{fig:coverage_width}
\end{figure}

Figure \ref{fig:coverage_width} compares JAW and its baselines, firstly regarding mean coverage and secondarily regarding median interval width, on all five UCI datasets for both neural network and random forest predictors, averaged over 200 experimental replicates. See Appendix \ref{app:models} for predictor function details. Meeting the target coverage level of $1-\alpha$ is the primary goal of the interval-generation audit task, but for methods that meet or nearly meet the target coverage level, smaller interval widths are more informative. Additionally, smaller variance in coverage indicates a more reliable or consistent method.



As seen in Figure \ref{fig:coverage_width}, the JAW predictive interval coverage is above the target level of 0.9 across all datasets, for both random forest and neural network $\widehat{\mu}$ functions, along with the jackknife-mm and weighted split methods. However, JAW's interval widths are generally smaller and thus more informative than those of jackknife-mm (which are often overly large, as noted in \cite{barber2021predictive}). Weighted split and JAW perform similarly on mean coverage and median interval width (both methods have coverage guarantees under covariate shift), but JAW avoids sample splitting and as a result has lower coverage variance than weighted split for all dataset and predictor conditions (see Appendix 
\ref{app:coverage_variance_comparison}), which suggests that JAW's predictive intervals are more reliable. 


\subsubsection{Interval generation results for JAWA: Coverage and interval width}
\begin{figure}[ht]
    \setlength{\tabcolsep}{0pt}
    \begin{tabular}{ccccc}
    \multicolumn{5}{c}{\includegraphics[width=0.45\textwidth]{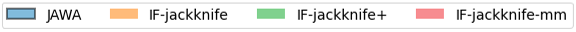}} \\
    \includegraphics[width=0.21\textwidth]{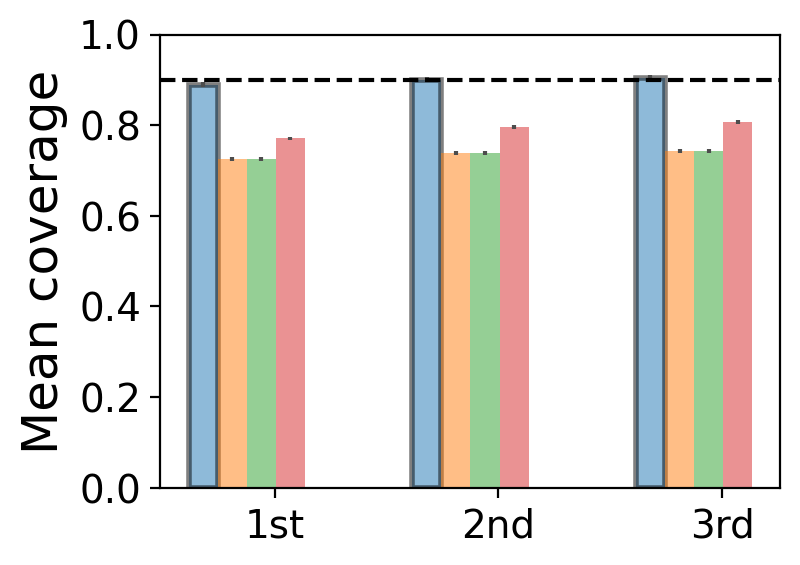}&  \includegraphics[width=0.195\textwidth]{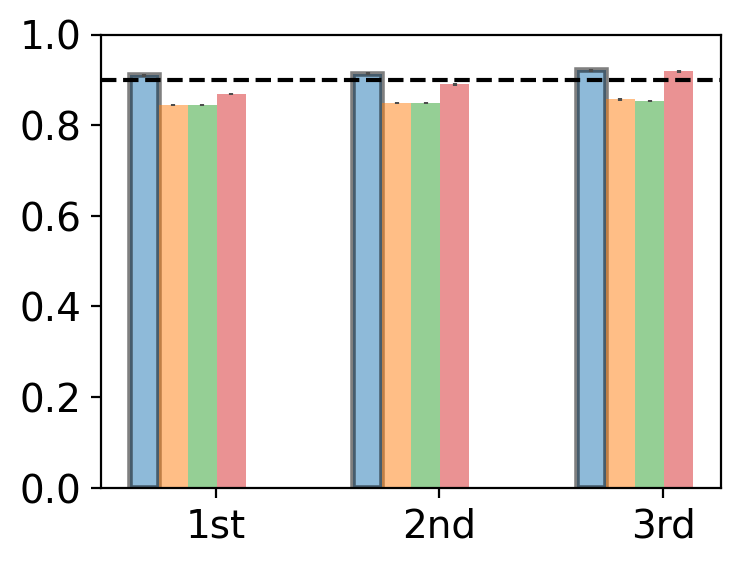} & \includegraphics[width=0.195\textwidth]{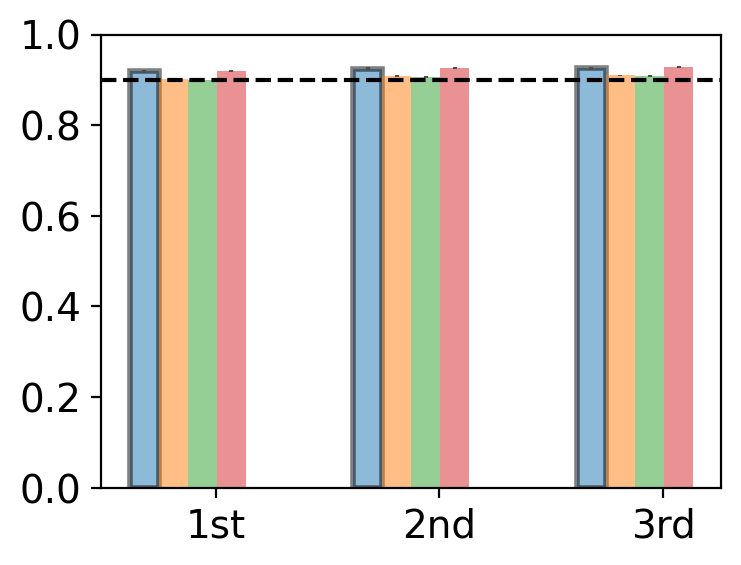} & \includegraphics[width=0.195\textwidth]{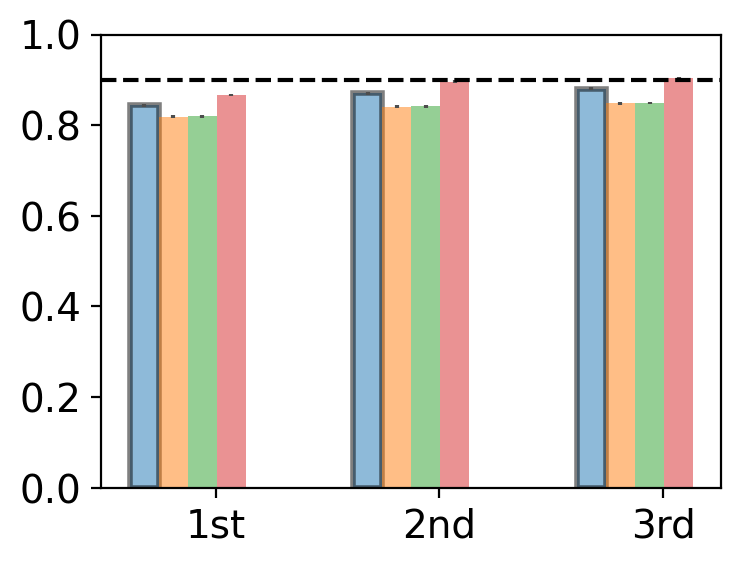} &  \includegraphics[width=0.195\textwidth]{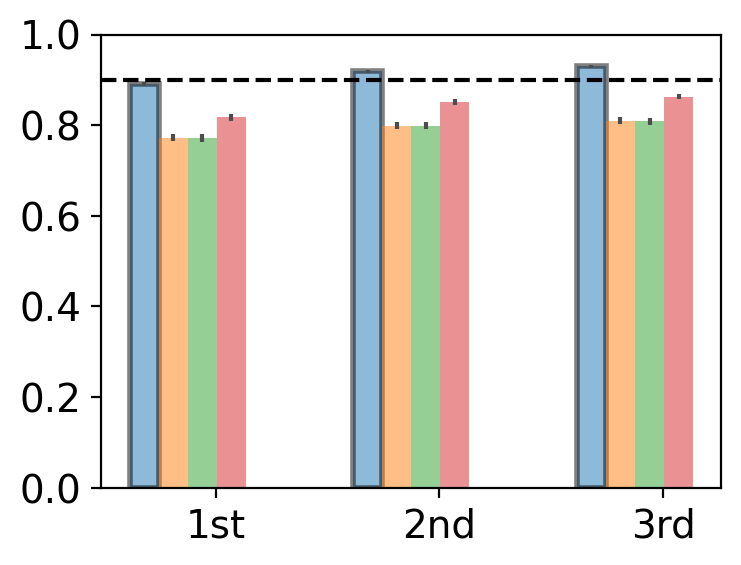} \\
    \includegraphics[width=0.215\textwidth]{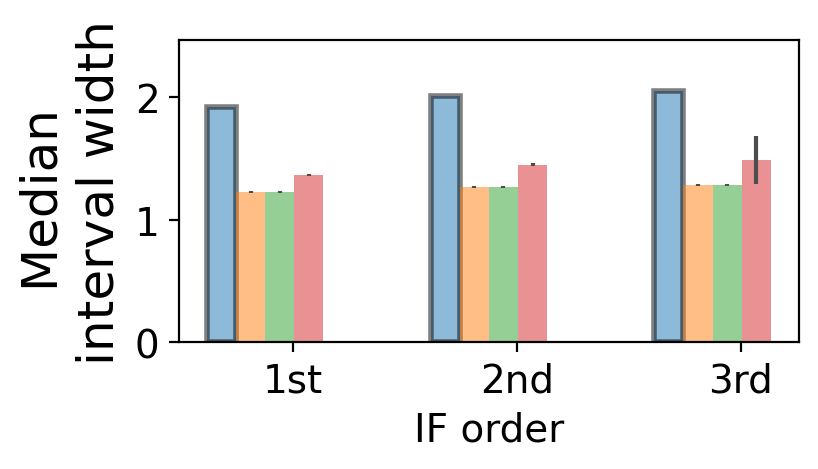}&  \includegraphics[width=0.19\textwidth]{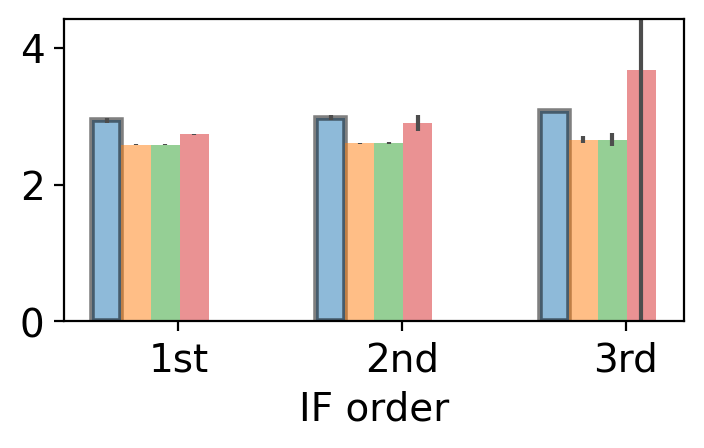} & \includegraphics[width=0.2\textwidth]{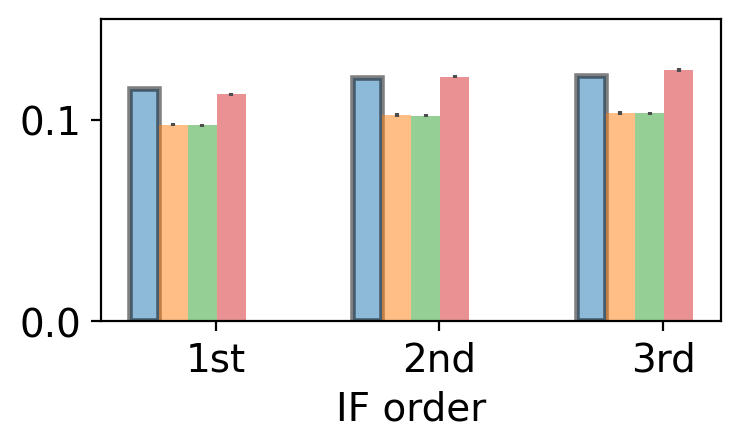} & \includegraphics[width=0.19\textwidth]{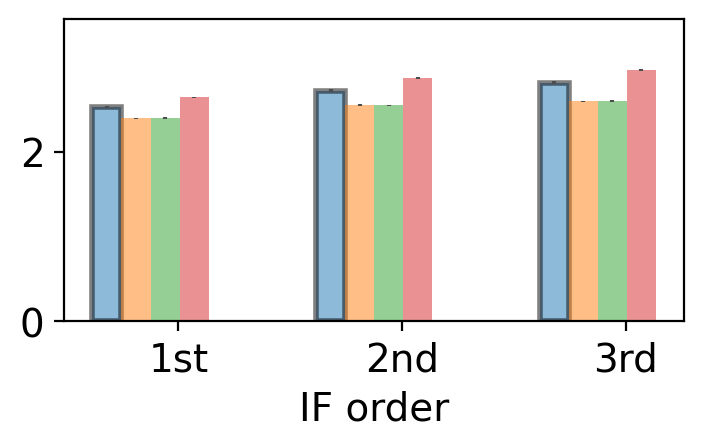} &  \includegraphics[width=0.19\textwidth]{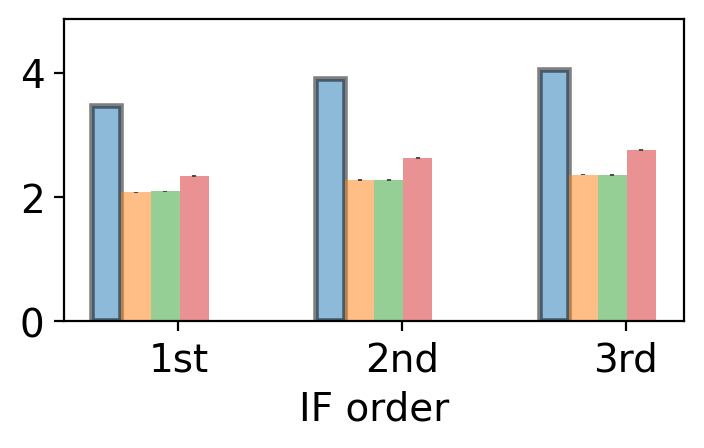} \\ \footnotesize
    (a) Airfoil & (b) Wine & (c) Wave & (d) Superconduct & (e) Communities
    \end{tabular}
    \caption{Mean coverage (first row) and median interval width (second row) for JAWA and baselines for influence function orders $K \in \{1, 2, 3\}$. Dashed line is the target coverage level ($1-\alpha$ = 0.9). Error bar shows the standard error of 200 repeated experiments. JAWA is more consistent than baselines in reaching or nearly reaching the target coverage level across datasets and influence function orders, and it is more computationally efficient than JAW (Appendix \ref{app:runtime}).
    }
    \label{fig:coverage_width_IFs}
\end{figure}

Figure \ref{fig:coverage_width_IFs} evaluates JAWA coverage and interval width compared to baselines for IF orders $K \in \{1, 2, 3\}$ with a neural network predictor (see Appendix \ref{app:models} for predictor details). As with the JAW experiments, coverage at the target level of $1-\alpha = 0.9$ is the primary goal, while secondarily, smaller intervals are more informative for methods that meet or nearly meet target coverage. For three of the five datasets (airfoil, wine, and communities), JAWA is the only method that consistently reaches or nearly reaches the target coverage level. JAWA and all the baselines perform well on the wave datasets, and in the superconduct dataset JAWA still outperforms approximations of jackknife and jackknife+ for all IF orders. Appendix \ref{app:runtime} provides an an example empirical comparison of JAWA and JAW runtimes, which demonstrates that JAWA can be orders of magnitude faster to compute. 

\subsubsection{Error assessment results for JAW-E: AUC}
\begin{figure}[ht]
 \setlength{\tabcolsep}{0pt}
 \begin{center}
\begin{tabular}{ccccc}
    & \multicolumn{3}{c}{\includegraphics[width=0.45\textwidth]{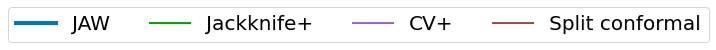}} \\
    \includegraphics[width=0.2\textwidth]{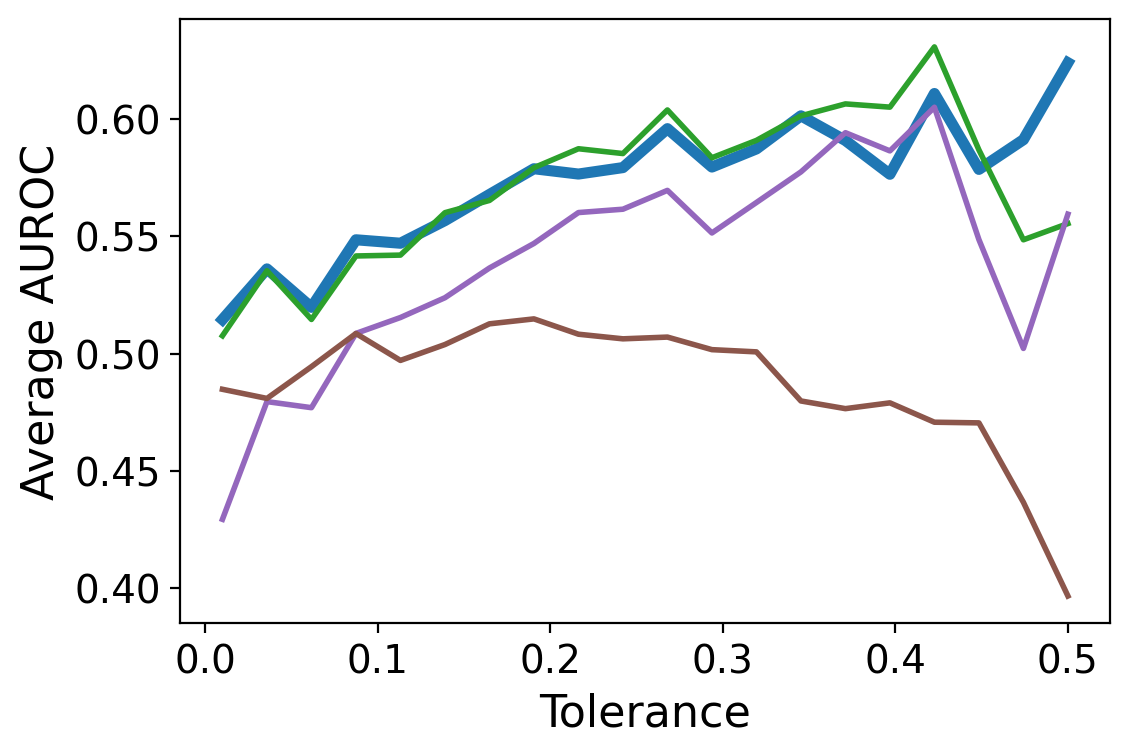} & 
    \includegraphics[width=0.2\textwidth]{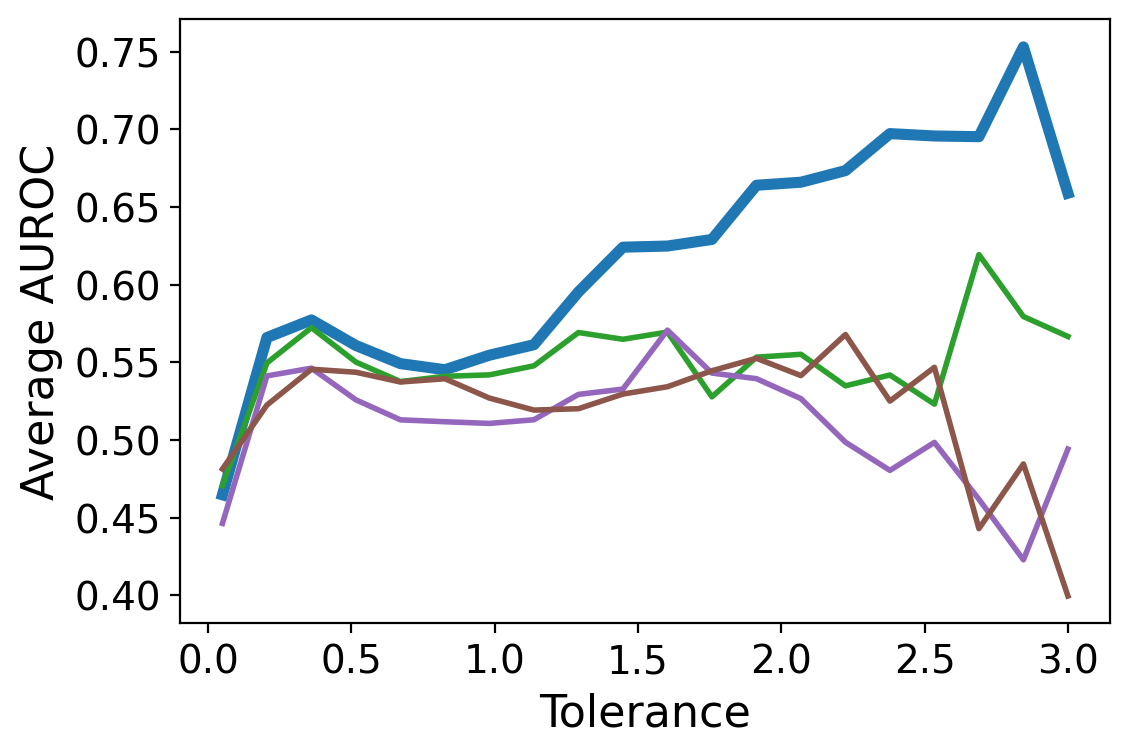} & 
    \includegraphics[width=0.2\textwidth]{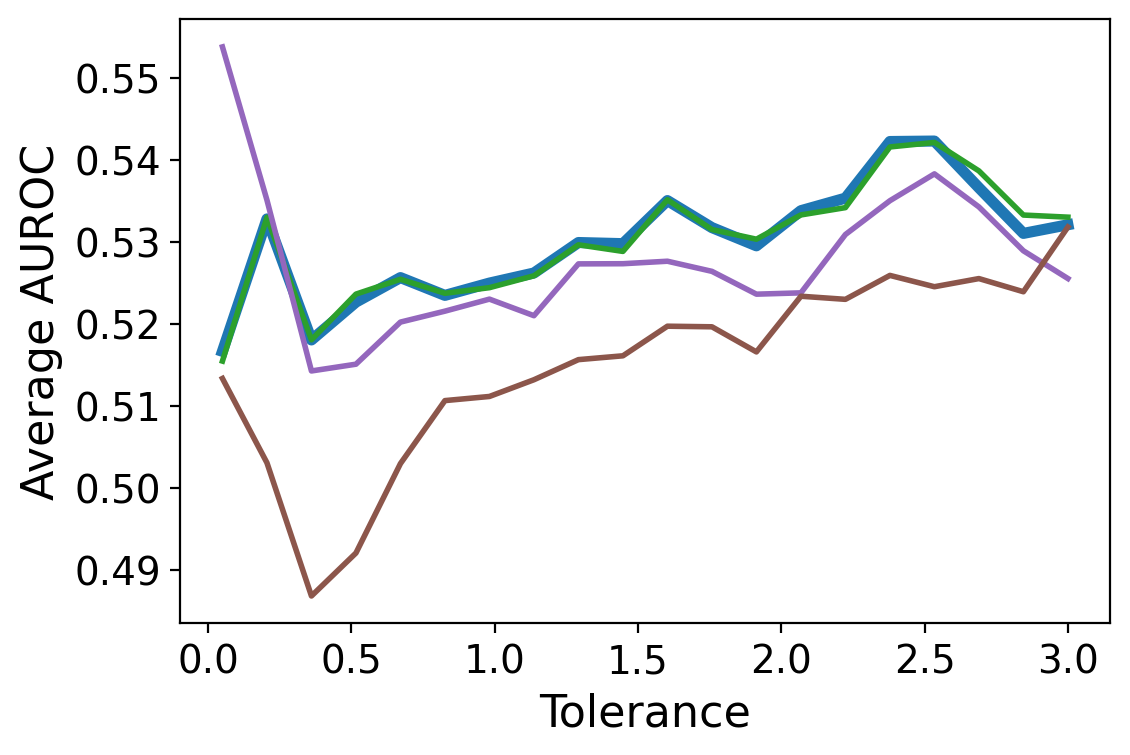} & 
    \includegraphics[width=0.2\textwidth]{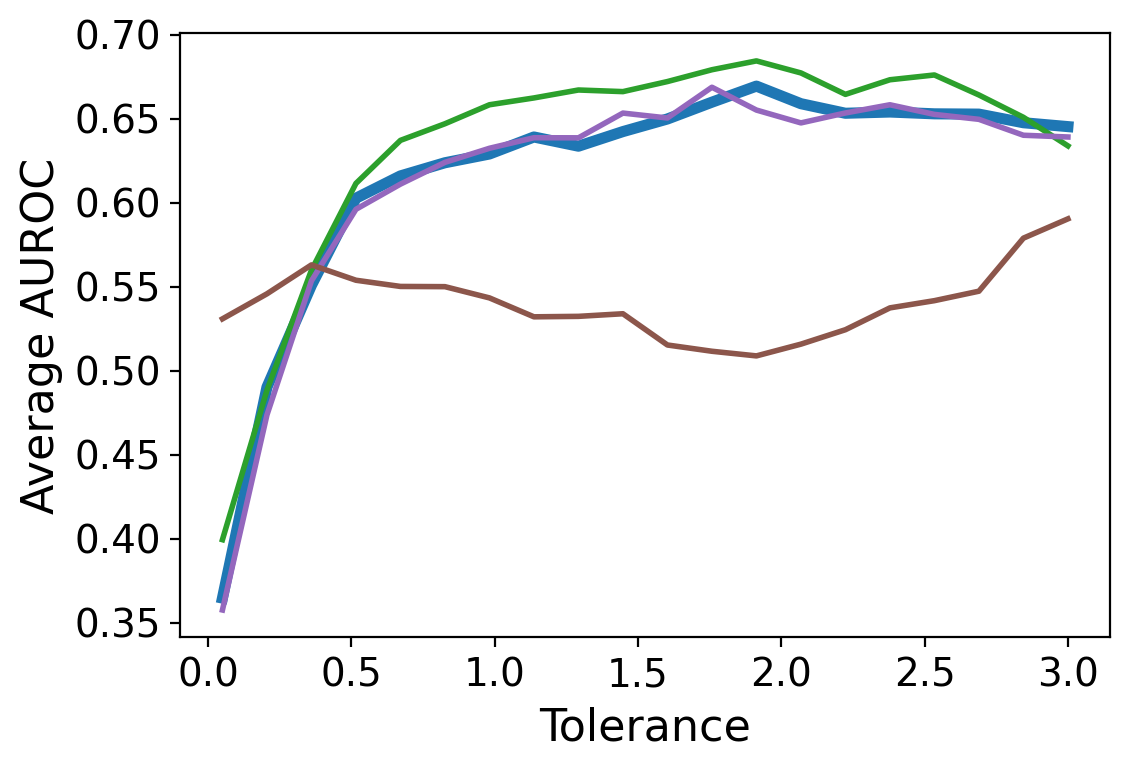} & 
    \includegraphics[width=0.2\textwidth]{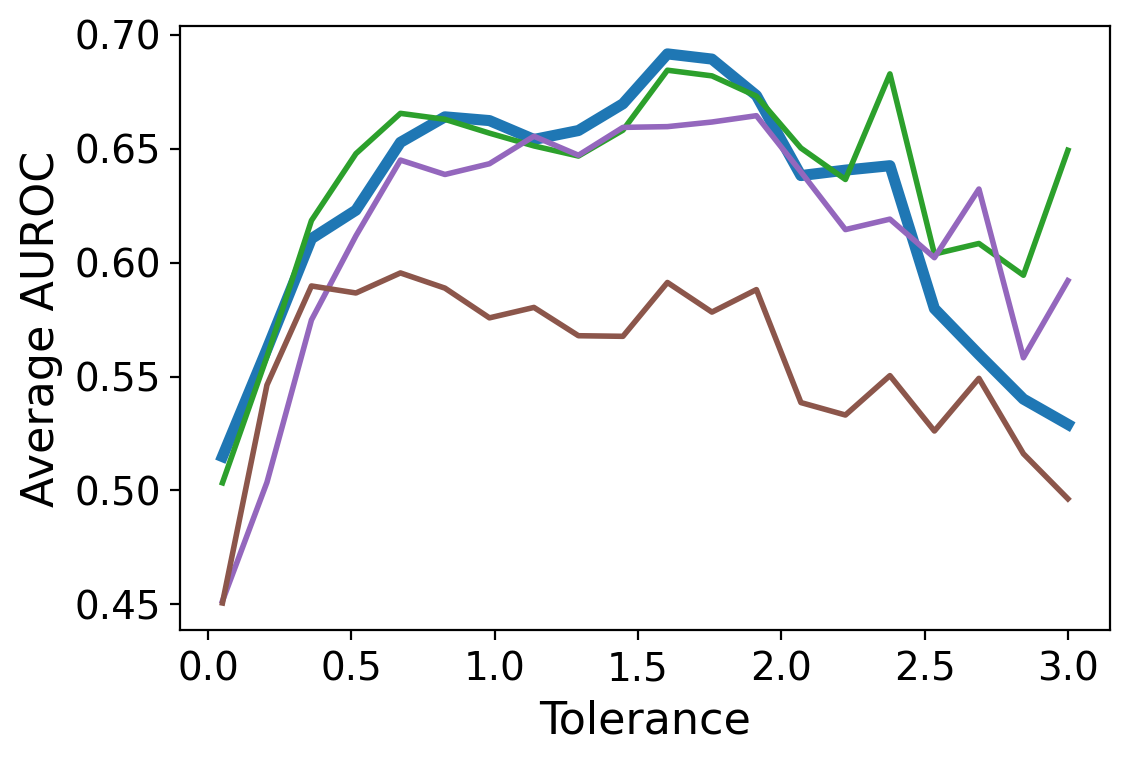}\\
     (a) Airfoil & (b) Wine & (c) Wave & (d) Superconduct & (e) Communities
    \end{tabular}
    \end{center}
    \caption{AUROC values for tolerance levels $\tau$ across the three datasets for the random forest predictor, averaged across 5 experiment replicates. Results for neural net predictor in Appendix \ref{app:auc_additional}.} 
    \label{fig:auc}
\end{figure}

We now turn to an error-assessment audit task where the goal is to evaluate a method's ability to estimate the probability that a given prediction is erroneous or not, based on the criterion $|Y_{n+1}-\widehat{\mu}(X_{n+1})|>\tau$. Let $\overline{E} = [\widehat{\mu}(X_{n+1}) - \tau, \ \widehat{\mu}(X_{n+1}) + \tau]$. Then, the goal is to estimate the probability that $\widehat{\mu}(X_{n+1})$ is correct, i.e., $Y_{n+1} \in  \overline{E}$; or an error, i.e., $Y_{n+1} \not\in  \overline{E}$. 
For four predictive interval-generation methods repurposed to the error assessment task (JAW-E, jackknife+E, cross validation+E, and split conformal-E), Figure \ref{fig:auc} reports the area under the receiver operating characteristic curve (AUROC) for values of $\tau$ selected with uniform spacing in the interval $[q_{n, 0.05}^-\{R_i^{LOO}\}, \ q_{n, 0.05}^+\{R_i^{LOO}\}]$ (between the 0.05 and 0.95 quantiles of the leave-one-out residuals for a dataset) for five repeated experiments with the random forest predictor. Better performing methods have higher overall AUROC values for all values $\tau$. 
JAW achieves AUROC values comparable to jackknife+ and CV+ for most tolerance levels and datasets. JAW moreover achieves higher AUROC values than its baselines in the wine dataset with the random forest predictor. These results suggest that, for error assessment, JAW likely performs at least as well as its baselines and may offer a marginal benefit depending on the dataset and predictor.

\section{Conclusion}
\label{sec:conclusion}

In this paper, we develop JAWS, a series of wrapper methods for distribution-free predictive uncertainty auditing tasks when the data exchangeability assumption is violated due to covariate shift. We also propose a general approach to repurposing any distribution-free predictive inference method to the error assessment task. We provide rigorous finite-sample guarantees for JAW and JAW-E on the interval generation and error assessment tasks respectively, and analogous asymptotic guarantees for the computationally efficient JAWA and JAWA-E. We moreover demonstrate superior performance of the JAWS series on a variety of datasets.
In supplementary experiments we investigate a number of JAWS' limitations: weight estimation can address the assumed access to oracle weights with similar empirical performance (Appendix \ref{app:jaw_e}), and JAW's increased coverage variance with covariate shift can be explained by reduced effective sample size due to importance weighting (Appendix \ref{app:ablation}). Additionally, we note that JAW and JAWA share a limitation with weighted conformal prediction \citep{tibshirani2019conformal} of potentially producing overly large intervals in extreme covariate shift cases where a test point's normalized likelihood ratio approaches or exceeds $\alpha$. In the future, we aim to address the problems of reducing coverage variance and improving predictive interval sharpness.

\acksection
This work was supported by the National Science Foundation grant IIS-1840088. We thank Yoav Wald for helpful discussions and advice, as well as Peter Schulam for sharing code that facilitated our influence function approximation implementation and AUC experiments.

\bibliography{references}

\newpage 


\appendix

\input{appendix}

\end{document}

%% file: appendix.tex
\section{Supplementary background details}

\subsection{Error assessment motivation: Concrete example}
\label{app:dose_example}

The error-assessment approach to predictor auditing may be more actionable than the interval-generation approach in error-critical or high-stakes decision-making situations where there is a clearly defined margin of error that is considered safe or acceptable. One example is chemical or radiation therapy dose prediction for cancer treatment, where administering the correct dosage within $5\%-10\%$ is error-critical. Machine learning is being increasingly employed in cancer chemotherapy and radiotherapy for purposes including dose optimization \citep{feng2018machine, huynh2020artificial}. Dose errors are one of the most common types of errors in chemotherapy and radiotherapy, occurring when a patient is given a substantially higher or lower than optimal amount of chemical or radiation treatment \citep{weingart2018chemotherapy, van2004errors}. An overdose of either chemotherapeutics or radiation can be harmful or even lethal to a patient, whereas underdose can result in a reduced anticancer effect \citep{gurney2002calculate}. A dose error is generally defined as a percentage deviation, between an administered dose and the truly optimal dose that should have been administered, beyond some error tolerance: $5\%$ or $10\%$ are commonly used deviation thresholds for defining errors \citep{cohen1996preventing, van2004errors}. Accordingly, the probability that the optimal dosage level lies within say $10\%$ of the predicted dosage level may be of greater interest to a provider and their patient than identifying a predictive interval with some predetermined coverage probability (which would provide no error assurance whenever the predictive interval extends beyond the safe threshold, say of $\pm10\%$). 

\subsection{Supplementary background on covariate shift}
\label{app:cov_shift_background}

Covariate shift is a type of dataset shift where the $Y | X$ distribution is the same between training and test data but the marginal $X$ distributions are allowed to change \citep{sugiyama2007covariate,shimodaira2000improving}. This is a strong but common assumption for many dataset shift problems. Covariate shift is also closed related to data missingness and sample selection bias \citep{bickel2009discriminative}. The most prevalent method for correcting the shift is by applying likelihood ratio or ``importance'' weights \citep{sugiyama2007covariate,shimodaira2000improving}. Density ratio estimation is then a key subproblem of covariate shift correction \citep{sugiyama2012density}. Other methods dealing with covariate shift include matching the (kernel) representation between the two distributions \citep{gretton2009covariate,yu2012analysis,zhang2013covariate,zhao2021reducing} and robust optimization \citep{liu2014robust,chen2016robust,duchi2019distributionally,rezaei2021robust}. 

\subsection{Supplementary comparison to \cite{barber2022conformal}}
\label{app:comparison_barber_etal_2022}

In the Section \ref{sec:weighted_conformal} of the main paper we contrast our JAW method to the results in \cite{barber2022conformal} regarding the nonexchangeable jackknife+. We emphasize that \cite{barber2022conformal} uses fixed weights and compensates for unknown violations of exchangeability at the expense of a coverage gap, whereas JAW uses data-dependent likelihood ratio weights and assumes covariate shift but does not suffer from a coverage gap. Additionally, it is also important to take note of and contrast our work with an extension of the framework in \cite{barber2022conformal} to data-dependent weights that the authors briefly discuss in their Section 5.3, subsection titled ``Fixed versus data-dependent weights'' (though this extension is not a primary focus of their work). In short, this extension from \cite{barber2022conformal} does not generalize to JAW beyond giving a trivial coverage guarantee. 

In particular, \cite{barber2022conformal} do not propose a likelihood-ratio weighting of the jackknife+, but if one were to define the weights in their nonexchangeable jackknife+ as data-dependent, likelihood ratio weights like in our JAW method, then the extension discussed in Section 5.3 of \cite{barber2022conformal} would in general suffer a coverage gap that could approach 1 under covariate shift. That is, under covariate shift assumptions and with $w_i = w(X_i)$ representing the likelihood ratio for datapoint $i$, the conditional total variation distance between the original ordered data $Z = (Z_1, ..., Z_{n+1})$ and the swapped data $Z^i = (Z_1, ..., Z_{i-1}, Z_{n+1}, Z_{i+1}, ..., Z_n)$ can generally approach 1 for nontrivial covariate shift, i.e., $\text{d}_{TV}(Z, Z^i | w_1, ..., w_n, t_1, ..., t_{n+1}) \rightarrow 1$. This is because under the covariate shift the training data $\{Z_1, ..., Z_n\}$ and test point $Z_{n+1}$ are not exchangeable (they are weighted exchangeable, i.e., $wZ \stackrel{d}{=} w^iZ^i$), meaning that the unweighted data distributions $Z$ and $Z^i$ may have arbitrarily large total variation distance. The result would then be the trivial coverage guarantee (i.e., only guaranteeing coverage probability $\geq 0$).

\subsection{Supplemnetary background on influence functions}
\label{app:IFs_background}

In this work we implement the algorithm proposed by \cite{giordano2019higher} to compute higher-order influence functions (IFs), so we refer \cite{giordano2019higher} for more comprehensive details and theory. However, in this supplementary section we provide additional details on basic IFs theory and our use of IFs for the convenience of the interested reader.


For a weight vector variable $\omega \in \mathbb{R}^n$ and a fixed instance of the variable $\omega = \tilde{\omega}$ representing a specific reweighting of the data, let us denote $\widehat{\mu}_{\tilde{\omega}}$ as the refitted model and $\hat{\theta}({\tilde{\omega}})$ as the refitted model parameters that would be obtained by retraining the model with data weights $\tilde{\omega}$. With our notation in this section we maintain some similarity to the notation in \cite{giordano2019higher}, but we use the Greek character $\omega$ rather than $w$ to disambiguate the IF data weights $\omega$ from the likelihood-ratio weights $w$ introduced in Section \ref{sec:weighted_conformal}. For the leave-one-out weight vectors that are of primary interest for approximating the jackknife+ and related methods with influence functions, for ease of notation we say that $\tilde{\omega} = -i$ denotes the all ones vector except with zero in the $i$-th component so that $\widehat{\mu}_{-i}$ still denotes the leave-one-out retrained model, and we denote the corresponding leave-one-out parameters as $\hat{\theta}_{-i} = \hat{\theta}(-i)$. 


For any specific weights $\tilde{\omega}$, influence functions assume that $\hat{\theta}(\tilde{\omega})$ is a local minimum of the objective function, and thus that $\hat{\theta}(\tilde{\omega})$ is the solution to the following system of equations, where $G$ is the gradient of the objective function with respect to the model parameters:
\begin{align}
    \hat{\theta}(\tilde{\omega}) := \theta \text{ such that } G(\theta, \tilde{\omega}) := \frac{1}{n}\Big(g_0(\theta) + \sum_{i=1}^n\tilde{\omega}_ig_i(\theta)\Big) = 0,
    \label{eq:theta_est_eq}
\end{align}
where $g_i(\theta)$ is the gradient of the objective function for datapoint $i$ and $g_0(\theta)$ is a prior or regularization term. For the predictor $\widehat{\mu} = \widehat{\mu}_{1_n}$ trained on the full, original dataset, we have $\tilde{\omega} = 1_n$ and can thus denote the model parameters for $\widehat{\mu}$ as $\hat{\theta} = \hat{\theta}(1_n)$. For a resampling-based uncertainty quantification method like the jackknife+ (or bootstrap, cross validation, or other jackknife methods), retraining the model for each new reweighting of the training data can sometimes be computationally burdensome or prohibitive. In these cases, we can instead estimate $\hat{\theta}(\omega)$ using influence functions to compute a Taylor series expansion in $\omega$ centered at $1_n$ (or more specifically a Von Mises expansion, see \cite{fernholz2012mises}). A first-order influence function---which we will denote as $\delta^1_\omega \hat{\theta}(1_n)$ for consistency with notation in \cite{giordano2019higher}---refers to the first-order directional derivative of the parameters $\hat{\theta}(\omega)$ with respect to the weights $\omega$: 
\begin{align}
    \delta^1_\omega \hat{\theta}(1_n) = \sum_{i=1}^n \frac{\partial\hat{\theta}(\omega)}{\partial \omega_i}\bigg|_{\omega=1_n}\Delta \omega_i,
    \label{eq:IF_1st_order}
\end{align}
where $\Delta \omega = \omega - 1_n$ is the direction of change in weights relative to the original weights $1_n$. The first-order influence function $\delta^1_\omega \hat{\theta}(1_n)$ thus enables a first-order Taylor series approximatinon of $\hat{\theta}(\omega)$, given by 
\begin{align}\hat{\theta}^{\text{IF-}1}(\omega) := \hat{\theta}(1_n) + \delta^1_\omega \hat{\theta}(1_n).
\label{eq:IF_TS_1st_order}
\end{align}
Computing the influence function $\delta^1_\omega \hat{\theta}(1_n)$ requires differentiation through the chain rule because $\hat{\theta}(\omega)$ is only implicitly a function of $\omega$ through estimating equation \eqref{eq:theta_est_eq}. The first-order Taylor series approximation of $\hat{\theta}(\omega)$ given in \eqref{eq:IF_TS_1st_order} can then be rewritten as 
\begin{align}\hat{\theta}^{\text{IF-}1}(\omega) := \hat{\theta}(1_n) - \hat{H}(\hat{\theta})^{-1}G(\hat{\theta})(w - 1_n).
\label{eq:IF_HG}
\end{align}
where $\hat{H}(\hat{\theta}) = \hat{H}(\hat{\theta}(1_n), 1_n)$ and $G(\hat{\theta}) = G(\hat{\theta}(1_n), 1_n)$ are the Hessian and the gradient of the objective function. 

Similarly, higher-order Taylor series approximations of $\hat{\theta}(\omega)$ can be obtained using higher order influence functions $\delta^k_\omega \hat{\theta}(1_n)$, where the $K$-th order Taylor series is given by 
\begin{align}\hat{\theta}^{\text{IF-}K}(\omega) := \hat{\theta}(1_n) + \sum_{k=1}^{K}\frac{1}{k!}\delta^k_\omega \hat{\theta}(1_n).
\label{eq:kth_order_app}
\end{align}
Computing $\hat{\theta}^{\text{IF-K}}(\omega)$ requires several assumptions. See \cite{giordano2019higher} for a formal list, but informally we assume that $\hat{\theta}(1_n)$ is the solution to $G(\hat{\theta}(1_n), 1_n) = 0$, that $G(\theta, 1_n)$ is $K+1$ times continuously differentiable, that the hessian $H(\hat{\theta})$ is strongly positive definite (meaning that the objective function is strongly convex in the neighborhood of the local solution), and the norm of the derivative $d_\theta^kG(\theta, 1_n)$ has a finite upper bound for $1 \leq k \leq K + 1$. In this work, we implement the recursive procedure based on forward-mode automatic differentiation to achieve memory-efficient computation of higher-order directional derivatives \cite {maclaurin2015autograd} as described in \cite{giordano2019higher}.

While \cite{alaa2020discriminative} propose a higher-order IF approximation of the jackknife+, their method assumes exchangeable (e.g., IID) train and test data and offer experiments with only first and second order IF approximations to the jackknife+. Our proposed JAWA sequence extends the IF approximation of the jackknife+ proposed by \cite{alaa2020discriminative} to the setting of covariate shift, and we demonstrate the benefits of this extension on a variety of datasets and orders of influence function approximation.

\section{Supplementary theoretical results}

\subsection{Error assessment assuming exchangeable data}
\label{app:error_assess}

While in Section \ref{sec:error_shift} of the main paper we present a general approach to repurposing predictive interval-generating methods with validity under covariate shift to the error assessment task, here we present the analogous results under the exchangeable data assumption. The results in this section directly apply to common predictive interval-generating methods including split conformal, jackknife+, and cross-validation+.

\begin{figure}[ht]
\centering
\includegraphics[width=0.9\textwidth]{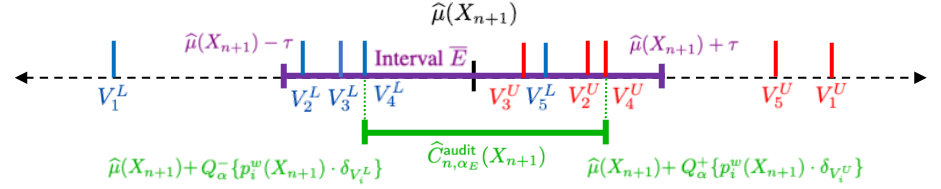}
\caption{Illustration of terms involved in computing $\alpha_E^{\text{audit}}$ when errors are defined by the event $|Y_{n+1}-\widehat{\mu}(X_{n+1})|>\tau$. The interval $\overline{E} = [\widehat{\mu}(X_{n+1}) -\tau, \widehat{\mu}(X_{n+1}) +\tau]$ is shown in violet, the values $\{V_i^L\}$ in blue, the values $\{V_i^U\}$ in red, and the interval $\widehat{C}_{n, \alpha}^{\text{audit}}(X_{n+1})$ in green. Each vertical line at a location $V_i$ on the real line represents a point mass $\delta_{V_i}$ with height $\frac{1}{n+1}$.}
\label{fig:thm3}
\end{figure}

The setup is the same as in the main paper Section \ref{sec:error_shift}, where we first define
\begin{align}
    \overline{E} = \big\{y \in \mathbb{R} : \tau^- \leq \widehat{S}(X_{n+1}, y) \leq \tau^+\big\}.
    \label{eq:error_event_app1}
\end{align}
However, unlike the covariate shift setting, we instead assume exchangeable data and access to a predictive inference method with valid predictive intervals of the form 
\begin{align}
    \widehat{C}_{n, \alpha}^{\text{audit}}(X_{n+1}) =  \big\{y  : \ & \widehat{Q}_{\alpha}^-\{\tfrac{1}{n+1}\delta_{V_i^L}\} \leq \widehat{S}(X_{n+1}, y) \leq \widehat{Q}_{1-\alpha}^+\{\tfrac{1}{n+1}\delta_{V_i^U}\}\big\}
    \label{eq:error_method_interval_iid}
\end{align}
Recall that we use $Q^-_{\alpha}\{\tfrac{1}{n+1}\delta_{V_i^L}\}$ to denote the level $\alpha$ quantile of the empirical distribution $\sum_i^n[\tfrac{1}{n+1}\delta_{V_i^L}] + \delta_{-\infty}$ and $Q^+_{1-\alpha}\{\tfrac{1}{n+1}\delta_{V_i^U}\}$ to denote the level $1-\alpha$ quantile of the empirical distribution $\sum_i^n[\tfrac{1}{n+1}\delta_{V_i^U}] + \delta_{\infty}$.
(Analogous to as stated in Section \ref{sec:error_shift}, \eqref{eq:error_method_interval_iid} gives the jackknife+ interval \eqref{eq:jackknife+} by setting $\widehat{S}(x, y) = y - \widehat{\mu}(x)$, $V_i^L = \widehat{\mu}_{-i}(X_{n+1}) - \widehat{\mu}(X_{n+1}) - R_i^{LOO}$, and $V_i^U = \widehat{\mu}_{-i}(X_{n+1}) - \widehat{\mu}(X_{n+1}) + R_i^{LOO}$. And, \eqref{eq:error_method_interval_iid} gives the prediction interval for split conformal prediction for absolute value residual scores when $\widehat{S}(x, y) = |y - \widehat{\mu}(x)|$, and for all calibration data $i$ we let $V_i^U = |Y_i - \widehat{\mu}(X_i)|$ and $V_i^L = 0$.)
Then, define $\alpha_E^{\text{audit}}$ as
\begin{align}
    \alpha_E^{\text{audit}} = \min\Big(\Big\{\alpha' \ : \ \tau^- \leq \ \widehat{Q}_{\alpha'}^-\{\tfrac{1}{n+1}\delta_{V_i^L}\} \ ,\ \widehat{Q}_{1-\alpha'}^+\{\tfrac{1}{n+1}\delta_{V_i^U}\} \leq \tau^+  \Big\}\Big).
\label{eq:alpha_I-IID}
\end{align}
We can then estimate the probability of $\widehat{\mu}(X_{n+1})$ \textit{not} resulting in an error as in \eqref{eq:error_event} as:
\begin{align}
    \widehat{p}\{Y_{n+1}\in \overline{E}\} = 
    \begin{cases} 1 - \alpha_E^{\text{audit}} & \text{ if } \alpha_E^{\text{audit}} \text{ exists} \\
    0 & \text{otherwise}.
    \end{cases}
\label{eq:thm3}
\end{align}
While the target coverage for $\widehat{C}_{n, \alpha_E}^{\text{audit}}(X_{n+1})$ is used in \eqref{eq:thm4}, the following theorem gives the worst-case error assessment guarantee for exchangeable data (proof in Appendix \ref{prf:3}). 
\begin{theorem}\textit{If a predictive inference method that generates predictive sets of the form \eqref{eq:error_method_interval_iid} has coverage guarantee $\mathbb{P}\{Y_{n+1} \in \widehat{C}_{n, \alpha}^{\text{audit}}(X_{n+1})\} \geq 1 - c_1\alpha - c_2 \ $ assuming exchangeable data, where $c_1, c_2 \in \mathbb{R}$, define $\overline{E}$ as in \eqref{eq:error_event_app1} and $\alpha_{E}^{\text{w-audit}}$ as in \eqref{eq:alpha_I-IID}. Then, 
}
\begin{align}
    \mathbb{P}\{Y_{n+1}\in \overline{E}\} \geq 
    \begin{cases} 1 - c_1\alpha_E^{\text{audit}} - c_2 & \text{ if } \alpha_E^{\text{audit}} \text{ exists and }\ \alpha_E^{\text{audit}} < \frac{1 - c_2}{c_1} \\
    0 & \text{otherwise}
    \end{cases}.
\end{align}
\label{thm3}
\end{theorem}

\subsection{JAW-E error assessment guarantee}
\label{app:JAW_R_guarantee}

We now state the error assessment guarantee for JAW-E as Corollary \ref{cor1}, which follows directly from Theorem \ref{thm4}. First, recall that we assume a predictive inference that has valid coverage under covariate shift and can be written in the form of \eqref{eq:error_method_interval}, which we restate here:
\begin{align}
    \widehat{C}_{n, \alpha}^{\text{w-audit}}(X_{n+1}) =  \big\{y \in \mathbb{R} : \ & \widehat{Q}_{\alpha}^-\{p_i^w(X_{n+1})\delta_{V_i^L}\} \leq \widehat{S}(X_{n+1}, y) \leq \widehat{Q}_{1-\alpha}^+\{p_i^w(X_{n+1})\delta_{V_i^U}\}\big\}
    \label{eq:error_method_interval_app_JAW_E}
\end{align}
To obtain the JAW predictive interval from \eqref{eq:error_method_interval_app_JAW_E}, we define the test point score function\footnote{Note that the test point score function $\widehat{S}(x, y) = y - \widehat{\mu}(x)$ that we use to obtain an alternative definition of the JAW interval in \eqref{eq:error_method_interval_app_JAW_E2} (and could analogously be used to define the jackknife+) has nuanced differences from the score functions used in weighted standard conformal prediction methods (as well as in their unweighted variants). As mentioned in the main paper,  \eqref{eq:error_method_interval_app_JAW_E} yields the weighted split conformal prediction interval for absolute value residual scores when $\widehat{S}(x, y) = |y - \widehat{\mu}(x)|$, and for all holdout calibration data $i$ we let $V_i^U = |Y_i - \widehat{\mu}(X_i)|$ and $V_i^L = 0$---so, we observe that for weighted split conformal prediction $V_i^U = \widehat{S}(X_i, Y_i)$ for all calibration data $i$, and thus $\widehat{S}$ can be understood as a ``nonconformity score'' as in standard conformal prediction. However, for JAW (and the jackknife+) there is a less clear correspondence between $\widehat{S}(x, y) = y - \widehat{\mu}(x)$ and $\{V_i^U\}$ (or $\{V_i^L\}$). We thus choose to define $\widehat{S}$ as a \textit{test point} score function in an effort to simultaneously maintain greater clarity on its meaning from a user's perspective, maintain intuitive connections to standard conformal prediction methods, and also avoid suggesting that $\{V_i^U\}$ and $\{V_i^L\}$ are directly defined from $\widehat{S}$ in the case of JAW and the jackknife+. It is also worth noting that there may be some score functions for which the jackknife+ and JAW are not defined, in which case the corresponding error assessment methods would not be defined.
} as $\widehat{S}(x, y) = y - \widehat{\mu}(x)$, and for all $i \in \{1, ..., n\}$ we let $V_i^L = \widehat{\mu}_{-i}(X_{n+1}) - \widehat{\mu}(X_{n+1}) - R_i^{LOO}$ and $V_i^U = \widehat{\mu}_{-i}(X_{n+1}) - \widehat{\mu}(X_{n+1}) + R_i^{LOO}$:
\begin{align}
    \widehat{C}_{n, \alpha}^{\text{w-audit}}(X_{n+1}) =  \big\{y \in \mathbb{R} : \ & \widehat{Q}_{\alpha}^-\{p_i^w(X_{n+1})\delta_{\widehat{\mu}_{-i}(X_{n+1}) - \widehat{\mu}(X_{n+1}) - R_i^{LOO}}\} \leq y - \widehat{\mu}(X_{n+1}) \nonumber \\
    \leq & \widehat{Q}_{1-\alpha}^+\{p_i^w(X_{n+1})\delta_{\widehat{\mu}_{-i}(X_{n+1}) - \widehat{\mu}(X_{n+1}) + R_i^{LOO}}\}\big\} \nonumber \\
    =  \big\{y \in \mathbb{R} : \ & \widehat{Q}_{\alpha}^-\{p_i^w(X_{n+1})\delta_{\widehat{\mu}_{-i}(X_{n+1}) - R_i^{LOO}}\}  \leq y \nonumber \\
    \leq & \widehat{Q}_{1-\alpha}^+\{p_i^w(X_{n+1})\delta_{\widehat{\mu}_{-i}(X_{n+1}) + R_i^{LOO}}\}\big\} \nonumber \\
    = \widehat{C}_{n, \alpha}^{\text{JAW}}(X&_{n+1})
    \label{eq:error_method_interval_app_JAW_E2}
\end{align}
Then, let us define $\alpha_E^{JAW}$ as:
\begin{align}
    \alpha_E^{JAW} = \min\Bigg(\Big\{\alpha' \ : \ \tau^- \leq \ & Q_{\alpha'}^-\{p_i^w(X_{n+1})\delta_{\widehat{\mu}_{-i}(X_{n+1}) - R_i^{LOO}}\} \nonumber ,\\
    \ & Q_{1-\alpha'}^+\{p_i^w(X_{n+1})\delta_{\widehat{\mu}_{-i}(X_{n+1}) + R_i^{LOO}}\} \leq \tau^+ \Big\}\Bigg).
    \label{app:alpha_E_JAW}
\end{align}


\begin{corollary}
\label{cor1}
\textit{Assume data under covariate shift from \eqref{eq:cs} where $\tilde{P}_X$ is absolutely continuous with respect to $P_X$. Define $\overline{E}$ as in \eqref{eq:error_event} and $\alpha_{E}^{\text{JAW}}$ as in \eqref{app:alpha_E_JAW}. Then, }
\begin{align}
    \mathbb{P}\{Y_{n+1}\in \overline{E}\} \geq 
    \begin{cases} 1 - 2\alpha_E^{JAW} &  \text{ if } \alpha_E^{\text{JAW}} \text{ exists and }\ \alpha_E^{JAW} < \frac{1}{2} \\
    0 & \text{otherwise}
    \end{cases}
\end{align}
\end{corollary}


\subsection{JAWA-E error assessment guarantee}
\label{app:JAWA_R_guarantee}

Lastly for our theoretical results, we state the error assessment guarantee for JAWA-E as Corollary \ref{cor2}. Whereas Corollary \ref{cor1} holds for finite samples, Corollary \ref{cor2} holds in the limit either of the number of samples or in the order of the influence function approximation.





First, define
\begin{align}
    \alpha_E^{\text{JAWA-}K} = \min\Bigg(\Big\{\alpha' \ : \ \tau^- \leq \ Q_{\alpha'}^-\{p_i^w(X_{n+1})\delta_{\widehat{\mu}_{-i}^{\text{IF-}K}(X_{n+1})- R_i^{\text{IF-}K, LOO}}\} \ ,\nonumber \\
    Q_{1-\alpha'}^+\{p_i^w(X_{n+1})\delta_{\widehat{\mu}_{-i}^{\text{IF-}K}(X_{n+1})+ R_i^{\text{IF-}K, LOO}}\} \leq \tau^+  \Big\}\Bigg).
    \label{app:alpha_E_JAWA}
\end{align}

\begin{corollary}
\label{cor2}
\textit{Let Assumptions 1 - 4 and either Condition 2 or Condition 4 from \cite{giordano2019higher} hold uniformly for all $n$. Assume data under covariate shift from \eqref{eq:cs} where $\tilde{P}_X$ is absolutely continuous with respect to $P_X$. Define $\overline{E}$ as in \eqref{eq:error_event} and $\alpha_{E}^{\text{JAWA-}K}$ as in \eqref{app:alpha_E_JAWA}. Then, 
}

Then, as either $n\rightarrow \infty$ or as $K \rightarrow \infty$, we have
\begin{align}
    \mathbb{P}\{Y_{n+1}\in \overline{E}\} \geq 
    \begin{cases} 1 - 2\alpha_E^{\text{JAWA-}K} &  \text{ if }  \alpha_E^{\text{JAWA-}K} \text{ exists and }\ \alpha_E^{\text{JAWA-}K} < \frac{1}{2} \\
    0 & \text{otherwise}
    \end{cases}
\end{align}
\end{corollary}

\section{Proofs for theoretical results}
\label{app:proofs}

\subsection{Proof of Theorem \ref{thm:1}}

\label{prf:1}
\begin{proof}

    
    We use (a) - (d) to denote four setup steps, and we use 1-3 to denote the main steps in the proof. Our first two initial setup steps (a) and (b) are identical to the corresponding setup steps in the proof for Theorem 1 in \cite{barber2021predictive}:
    \begin{itemize}
        \item[(a)] First, we suppose the hypothetical case where in addition to the training data $\{(X_1, Y_1), ..., (X_n, Y_n)\}$, we also have access to the test point $(X_{n+1}, Y_{n+1})$. For each pair of indices $i, j \in \{1, ..., n+1\}$ with $i \neq j$, we define $\tilde{\mu}_{-(i, j)}$ as the regression function fitted on the training and test data except with the points $i$ and $j$ removed. (We follow the notation in \cite{barber2021predictive} where $\tilde{\mu}$ rather than $\widehat{\mu}$ reminds us that the former is fit on a subset of data $1, ..., n+1$ that may contain the test point $n+1$.) We note that $\tilde{\mu}_{-(i, j)} = \tilde{\mu}_{-(j, i)}$ for any $i \neq j$, and $\tilde{\mu}_{-(i, n+1)} = \widehat{\mu}_{-i}$ for any $i = 1, ..., n$. 
        \item[(b)] We also define the same matrix of residuals in \cite{barber2021predictive}, $R \in \mathbb{R}^{(n+1)\times (n+1)}$, with entries 
        $$R_{ij} = \begin{cases}
        + \infty & i = j, \\
        |Y_i - \tilde{\mu}_{-(i, j)}(X_i)| & i \neq j
        \end{cases}$$
        such that the off-diagonal entries $R_{ij}$ represent the residual for the $i$th datapoint where both $i$ and $j$ are not seen by the regression fitting. 
    \end{itemize}
    
    At this point we begin to introduce some changes to the proof in \cite{barber2021predictive}: 
    
    \begin{itemize}
        \item[(c)] We define a weighted version of the comparison matrix in \cite{barber2021predictive}, which we call $A^w \in \mathbb{R}^{(n+1) \times (n+1)}$, with entries $A^w_{ij} = w(X_i)w(X_j)\cdot\mathbbm{1}\{R_{ij} > R_{ji}\}$. That is, $\mathbbm{1}\{R_{ij} > R_{ji}\}$ is the indicator for the event that, when $i$ and $j$ are excluded from the regression fitting, $i$ has larger residual than $j$; and $A^w_{ij}$ is this indicator multiplied by $w(X_i)w(X_j)$, the product of the likelihood ratios for points $i$ and $j$. For any $i, j \in \{1, ..., n+1\}$, note that $A^w_{ij} > 0$ implies $A^w_{ji} = 0$ for any $i, j \in \{1, ..., n+1\}$. Moreover, note that in the absence of covariate shift, $w(X_i) = w(X_j) = 1$ for all $i, j \in \{1, ..., n+1\}$ and the weighted comparison matrix $A^w$ is equivalent to the unweighted comparison matrix $A$ described in \cite{barber2021predictive}. 
        
        
        \item[(d)] Next, as in \cite{barber2021predictive} we are interested in identifying points that have unusually large residuals and are thus hard to predict. \cite{barber2021predictive} defined such points with unusually large residuals as points $i$ where $\mathbbm{1}\{R_{ij} > R_{ji}\}$ for a sufficiently large fraction of other points $j$. However, in the covariate shift setting, we need to account for the fact that the informativeness of the comparison $\mathbbm{1}\{R_{ij} > R_{ji}\}$ depends on the likelihood of $j$ in the test distribution relative to the training distribution: If $w(X_j) > w(X_{j'})$ for some points $j, j' \in \{1, ..., n+1\} \backslash i, \ j \neq j'$, then the comparison $\mathbbm{1}\{R_{ij} > R_{ji}\}$ should contain more information about how difficult $i$ is to predict than the comparison $\mathbbm{1}\{R_{ij'} > R_{j'i}\}$. In particular, we are interested in identifying points $i$ where $\mathbbm{1}\{R_{ij} > R_{ji}\}$ for a sufficiently large \textit{total normalized weight} of other points $j$. With this motivation, we here define the set of ``strange'' points $\mathcal{S}(A^w) \subseteq \{1, ..., n+1\}$ in the following two equivalent ways that each serve a different illustrative purpose:
        \begin{align*}
            \mathcal{S}(A^w) & = \Big\{i \in \{1, ..., n+1\} \ : \ w(X_i)>0, \quad \sum_{j=1}^{n+1}\Big(p_j^w(X_{n+1})\cdot\mathbbm{1}\{R_{ij} > R_{ji}\}\Big) \geq 1 - \alpha\Big\} \\
            & = \Big\{i \in \{1, ..., n+1\} \ : \ w(X_i)>0, \quad \frac{\sum_{j=1}^{n+1}A^w_{ij}}{{w(X_i)\sum_{k=1}^{n+1}w(X_k)}} \geq 1 - \alpha\Big\}
        \end{align*}
        
        The first definition represents our intuition of $\mathcal{S}(A^w)$ as a set of ``strange'' points, which we have described (where $\mathbbm{1}\{R_{ij} > R_{ji}\}$ for a sufficiently large total normalized weight of other points $j$). That is, in the first definition it is relatively straightforward to see how $\mathcal{S}(A^w) \subseteq \{1, ..., n+1\}$ is the set of points $i \in \{1, ..., n+1\}$ such that for all the points $j \in \{1, ..., n+1\}, j \neq i$ where $R_{ij} > R_{ji}$, that the sum of the normalized weights $p_j^w(X_{n+1})$ of all such points $j$ is sufficiently large (at least $1 - \alpha$). On the other hand, the second definition represents how the set of strange points can be computed from the weighted comparison matrix $A^w$ and the likelihood ratio weights $w$. Note that in the second definition, when $w(X_k) = 1$ for all $k = 1, ..., n+1$ it is straightforward to see that $\mathcal{S}(A^w)$ is equivalent to the set of strange points $\mathcal{S}(A)$ in the \cite{barber2021predictive} proof (to see this, observe that in this case $\sum_{k=1}^{n+1}w(X_k) = n+1$). 
     
    \end{itemize}
    
    The following main steps in our proof take the following structure, similar to as in \cite{barber2021predictive}:
    \begin{itemize}
        \item Step 1: Establish deterministically that $\sum_{i \in \mathcal{S}(A^w)}p_i^w(X_{n+1}) \leq 2\alpha$. That is, for any comparison matrix $A^w$, it is impossible to have the total normalized weight of all the strange points exceed $2\alpha$. 
        \item Step 2: Using the fact that the datapoints are weighted exchangeable, show that the probability that the test point $n+1$ is strange (i.e., $n+1 \in \mathcal{S}(A^w))$ is thus bounded by $2\alpha$. 
        \item Step 3: Lastly, verify that the JAW interval can only fail to cover the test label value $Y_{n+1}$ if $n+1$ is a strange point. 
    \end{itemize}
    
    \textit{Step 1: Bounding the total normalized weight of the strange points.} This proof step follows and generalizes the corresponding proof step for Theorem 1 in \cite{barber2021predictive}, which relies on Landau's theorem for tournaments \citep{landau1953dominance}. For each pair of points $i$ and $j$ where $i \neq j$, let us say that $i$ ``wins'' its game against point $j$ if $A^w_{ij}>0$, that is if both $i$ and $j$ have nonzero density in the test distribution and if there is a higher residual on point $i$ than on point $j$ for the regression model $\tilde{\mu}_{-(i, j)}$. We say that $i$ loses its game with $j$ otherwise. 

    However, whereas \cite{barber2021predictive} derive a bound on the \textit{number} of strange points from a bound on the \textit{number of pairs} of strange points, we instead derive a bound on the \textit{total normalized weight} of the strange points from a bound on the sum of the \textit{product of normalized weights} for two strange points in a pair. As we will see, this idea generalizes the idea of counting pairs of points to account for continuous weights on the points: If all points have uniform unnormalized weight of 1, then, after adjusting for a normalizing constant in our construction, the product of unnormalized weights of points in a pair is 1 for all pairs and our construction reduces to bounding the number of distinct pairs of strange points.
    
    Observe that, by the definition of a strange point, the points that each strange point $i \in \mathcal{S}(A^w)$ wins against must have total normalized weight greater than or equal to $(1-\alpha)$, and thus the points that each strange point $i \in \mathcal{S}(A^w)$ loses to can only have total normalized weight at most $\alpha - p_i^w(X_{n+1})$ (our definition does not allow $i$ to lose to itself). That is:
    
    \begin{align*}
        \begin{matrix}
        \text{Total normalized weight} \\
        \text{of points that } i \text{ loses to}
        \end{matrix} = \sum_{j=1}^{n+1}\Big(p_j^w(X_{n+1})\cdot\mathbbm{1}\{R_{ij}\leq R_{ji}\}\Big) \leq \alpha - p_i^w(X_{n+1})
    \end{align*}

    This inequality will help us obtain an upper bound on the sum of the product of normalized weights between strange points in a pair. To aid with intuition, it may be helpful to think about a correspondance between a product of two weights and the area of a rectangle with side lengths equal to each weight value. Suppose that for each strange point $i \in \mathcal{S}(A^w)$ we construct a rectangle $L_i$ with width equal point $i$'s normalized weight, $p_i^w(X_{n+1})$, and length equal to the largest total normalized weight that the points that $i$ loses to could have, $\alpha - p_i^w(X_{n+1})$. In addition, suppose that we also construct a second rectangle $L_i'$ for each strange point $i \in \mathcal{S}(A^w)$ with width equal to $p_i^w(X_{n+1})$---note that $L_i'$ has the same width as $L_i$---but with length equal to half the total normalized weight of all of the strange points other than $i$, that is, $\frac{1}{2}\sum_{j \in \mathcal{S}(A^w)\backslash i}\ p_j^w(X_{n+1})$.
    
    We now take a moment to describe the meaning of the total area of the set of rectangles $\{L_i\}$ in a way that we will soon make use of: The total area of $\{L_i\}$ is an upper bound on the sum of products of normalized weights for all points in a pair where one point is a strange point and the other point is a point that the strange point loses to. To see this, note that by construction the area of any rectangle $L_i$ is the product of point $i$'s normalized weight (i.e., $p_i^w(X_{n+1})$) with an upper bound on the total normalized weight that the points $i$ loses to could have (i.e., $\alpha - p_i^w(X_{n+1})$). Thus, the area of $L_i$ is by construction an upper bound on the product of point $i$'s normalized weight (i.e., $p_i^w(X_{n+1})$) with the total normalized weight of the points that $i$ \textit{actually} loses to. 
    To state with more precise notation that we will use again later, for each point $j$ that $i$ \textit{actually} loses to, let us construct a rectangle $L_{ij}$ with width $p_i^w(X_{n+1})$ and length $p_j^w(X_{n+1})$. Then, for all these points $j$, we can arrange the rectangles $\{L_{ij}\}$ so that they are contained within $L_{i}$ and so that $L_{ij}$ and $L_{ij}'$ have zero overlapping area for all $j \neq j'$: that is, by this construction $\sum_{j: i \text{ loses to } j}\text{Area}(L_{ij}) \leq \text{Area}(L_{i})$. So, it is equivalent to describe the area of $L_i$ as an upper bound on the sum, over all points $j$ that $i$ loses to, of the product of $i$'s normalized weight with $j$'s normalized weight; and thus by extension, the total area of $\{L_i\}$ is as we described earlier.
    

    On the other hand, the total area of the set of rectangles $\{L_i'\}$ is the sum of the product of the normalized weights of two strange points in a pair over all pairs of strange points, where the factor of $\frac{1}{2}$ avoids double counting the pairs of strange points. To see this, note that for every pair of strange points $\{i, j\}$ there is a distinct subrectangle---call it $L_{ij}'$---that is contained in $L_i'$, such that $L_{ij}'$ has width $p_i^w(X_{n+1})$ and length $\frac{1}{2}p_j^w(X_{n+1})$ (where we also assume that for any $j \neq j'$, $L_{ij}$ and $L_{ij}'$ overlapping area of zero). Moreover, for this pair of strange points $\{i, j\}$ there is also an analogous subrectangle $L_{ji}'$ with width $p_j^w(X_{n+1})$ and length $\frac{1}{2}p_i^w(X_{n+1})$ contained in $L_j'$. Thus, the combined area of $L_{ji}'$ and $L_{ij}'$ is $\text{Area}(L_{ij}') + \text{Area}(L_{ji}') = p_i^w(X_{n+1})\cdot p_j^w(X_{n+1})$, and the total area of the set of rectangles $\{L_i'\}$ is as described. (Furthermore, note that when the unnormalized weights are all equal to 1 as in \cite{barber2021predictive}, the area of $\{L_i'\}$---adjusted by a normalization constant---is equivalent to the total number of pairs of strange points $s(s-1)/2$, where $s = |S(A^w)|$ is the number of strange points.)
    
    Now, observe that any pair of two strange points is also a pair of points where one point is strange and the other is a point that the strange point loses to, so the set of pairs of points included in the construction of $\{L_i'\}$ is a subset of the set of pairs of points for which the area of $\{L_i\}$ is the upper bound previously described. To be more precise, let $\{i, j\}$ be a pair of strange points, where (without loss of generality) let us say $i$ loses to $j$. Then, for the $L_{ij}'$ and $L_{ji}'$ as described before, there exists a distinct $L_{ij}$ such that $\text{Area}(L_{ij}') + \text{Area}(L_{ji}') = \text{Area}(L_{ij})$. More generally, we see that the total area of all the subrectangles $\{L_{ij}'\}$ is bounded by the total area of the subrectangles $\{L_{ij}\}$, that is $\sum_{i, j \in \mathcal{S}(A^w), \ i \neq j}\text{Area}(L_{ij}') = \sum_{i, j \in \mathcal{S}(A^w), \ i \neq j}\text{Area}(L_{ij}) \leq \sum_{i \in \mathcal{S}(A^w), \ i \text{ loses to } j}\text{Area}(L_{ij})$. Moreover, by construction $\sum_{i, j \in \mathcal{S}(A^w), \ i \neq j}\text{Area}(L_{ij}') = \sum_{i \in \mathcal{S}(A^w)}\text{Area}(L_{i}')$ and $\sum_{i \in \mathcal{S}(A^w), \ i \text{ loses to } j}\text{Area}(L_{ij}) \leq \sum_{i \in \mathcal{S}(A^w)}\text{Area}(L_{i})$. Therefore, the area of the set of rectangles $\{L_i'\}$ is less than or equal to the area of rectangles $\{L_i\}$, which we can write as follows:
    \begin{align*}
        \sum_{i \in \mathcal{S}(A^w)}\text{Area}(L_i') & \leq \sum_{i \in \mathcal{S}(A^w)}\text{Area}(L_i)\\
        \sum_{i \in \mathcal{S}(A^w)}\Big(p_i^w(X_{n+1})\cdot\frac{1}{2}\sum_{j \in \mathcal{S}(A^w)\backslash i}\ p_j^w(X_{n+1})\Big) & \leq \sum_{i \in \mathcal{S}(A^w)}\Big(p_i^w(X_{n+1})\cdot\big(\alpha - p_i^w(X_{n+1})\big)\Big)
        \tag{C.1.1}
        \label{C.1.1}
    \end{align*}
    Recall that we defined $p_i^w(X_{n+1}) = w(X_i)/\sum_{k=1}^{n+1}w(X_k) \ \forall \ i \in \{1, ..., n+1\}$, so in the uniform weighted case where $w(X_i) = 1  \ \forall \ i \in \{1, ..., n+1\}$ then $\sum_{k=1}^{n+1}w(X_k) = n+1$, and multiplying both sides of the inequality above by $(n+1)^2$ yields the analogous inequality in \cite{barber2021predictive} that bounds the number of pairs of points. 
    
    We now proceed to solve for an upper bound on $\sum_{i \in \mathcal{S}(A^w)}p_i^w(X_{n+1})$, the total normalized weight of strange points:
    {\small
    \begin{align*}
       \frac{1}{2}\sum_{i \in \mathcal{S}(A^w)}\Big(p_i^w(X_{n+1})\cdot\sum_{j \in \mathcal{S}(A^w)\backslash i}\ p_j^w(X_{n+1})\Big) & \leq \sum_{i \in \mathcal{S}(A^w)}\Big(p_i^w(X_{n+1})\cdot\big(\alpha - p_i^w(X_{n+1})\big)\Big) \\
       \frac{1}{2}\sum_{i, j \in \mathcal{S}(A^w), \ i\neq j}p_i^w(X_{n+1}) p_j^w(X_{n+1}) & \leq \quad \alpha\sum_{i \in \mathcal{S}(A^w)}p_i^w(X_{n+1}) \ - \sum_{i \in \mathcal{S}(A^w)}p_i^w(X_{n+1})^2 \\
       \frac{1}{2}\Bigg(\sum_{i \in \mathcal{S}(A^w)}p_i^w(X_{n+1})^2 \ + \sum_{i, j \in \mathcal{S}(A^w)}p_i^w(X_{n+1}) p_j^w(X_{n+1})\Bigg) & \leq \quad \alpha  \sum_{i \in \mathcal{S}(A^w)}p_i^w(X_{n+1}) \\
        \frac{1}{2}\Bigg(\sum_{i \in \mathcal{S}(A^w)}p_i^w(X_{n+1})^2 \ + \sum_{i \in \mathcal{S}(A^w)}\Big(p_i^w(X_{n+1})\cdot \sum_{j \in \mathcal{S}(A^w)}p_j^w(X_{n+1})\Big)\Bigg) & \leq \quad \alpha \sum_{i \in \mathcal{S}(A^w)}p_i^w(X_{n+1}) \\
        \frac{1}{2}\Bigg(\sum_{i \in \mathcal{S}(A^w)}p_i^w(X_{n+1})^2 \ + \Big(\sum_{i \in \mathcal{S}(A^w)}p_i^w(X_{n+1})\Big)^2\Bigg) & \leq \quad \alpha \sum_{i \in \mathcal{S}(A^w)}p_i^w(X_{n+1}) \\
        \frac{\sum_{i \in \mathcal{S}(A^w)}p_i^w(X_{n+1})^2}{\sum_{i \in \mathcal{S}(A^w)}p_i^w(X_{n+1})} \ + \Big(\sum_{i \in \mathcal{S}(A^w)}p_i^w(X_{n+1})\Big) & \leq 2\alpha \\
        \sum_{i \in \mathcal{S}(A^w)}p_i^w(X_{n+1}) & \leq 2\alpha - \frac{\sum_{i \in \mathcal{S}(A^w)}p_i^w(X_{n+1})^2}{\sum_{i \in \mathcal{S}(A^w)}p_i^w(X_{n+1})}
    \end{align*}}
    where because $0 \leq p_i^w(X_{n+1}) \leq 1 \ \forall \ i = 1, ..., n+1$ and $p_i^w(X_{n+1}) > 0$ for some $i \in \{1, ..., n+1\}$, we have $0 \leq p_i^w(X_{n+1})^2 \leq p_i^w(X_{n+1}) \ \forall \ i = 1, ..., n+1$ and thus $0 \leq \frac{\sum_{i \in \mathcal{S}(A^w)}p_i^w(X_{n+1})^2}{\sum_{i \in \mathcal{S}(A^w)}p_i^w(X_{n+1})} \leq 1$, and we have
    \begin{align}
        \sum_{i \in \mathcal{S}(A^w)}p_i^w(X_{n+1}) \leq 2\alpha
    \end{align}
    as desired. 
    
    \textit{Step 2: Weighted exchangeability of the datapoints.} We now leverage the weighted exchangeability of the data to show that, since the total weight of the strange points is at most $2\alpha$, that a test point has at most $2\alpha$ probability of being strange. 
    
    Note that the datapoints $(X_1, Y_1), ..., (X_{n+1}, Y_{n+1})$ are weighted exchangeable with a regression fitting algorithm $\mathcal{A}$ that is invariant to the ordering of the data. As a result, $\mathbbm{1}\{R_{ij} > R_{ji}\}$ are weighted exchangeable random variables such that $w(X_i)w(X_j)\mathbbm{1}\{R_{ij} > R_{ji}\}$ can be treated exchangeably, and thus it follows that $A^w \stackrel{\text{d}}{=} \Pi A^w \Pi^T$ for any $(n+1)\times (n+1)$ permutation matrix $\Pi$, where $\stackrel{\text{d}}{=}$ denotes equality in disribution. 
    
    Specifically, for any index $j \in \{1, ..., n+1\}$, suppose take $\Pi$ to be the permutation matrix with $\Pi_{j, n+1} = 1$ (a permutation mapping $n+1$ to $j$). Then, deterministically
    
    $$n+1 \in \mathcal{S}(A^w) \iff j \in \mathcal{S}(\Pi A^w\Pi^T)$$
    
    so thus 
    
    $$\mathbb{P}\{n+1 \in \mathcal{S}(A^w)\} = \mathbb{P}\{j \in \mathcal{S}(\Pi A^w\Pi^T)\} = \mathbb{P}\{j \in \mathcal{S}(A^w)\}$$
    
    for all $j = 1, ..., n+1$. That is, an arbitrary training point $j$ is equally likely to be strange as the test point $n+1$. Multiplying by $\frac{w(X_j)}{\sum_{k=1}^{n+1}w(X_k)}$, we obtain
    
    $$\frac{w(X_j)}{\sum_{k=1}^{n+1}w(X_k)}\mathbb{P}\{n+1 \in \mathcal{S}(A^w)\} = \frac{w(X_j)}{\sum_{k=1}^{n+1}w(X_k)}\mathbb{P}\{j \in \mathcal{S}(A^w)\}$$
    
    And summing over $j$, we have
    
    \begin{align*}
        \sum_{j=1}^{n+1}\frac{w(X_j)}{\sum_{k=1}^{n+1}w(X_k)}\mathbb{P}\{n+1 \in \mathcal{S}(A^w)\} & = \sum_{j=1}^{n+1}\frac{w(X_j)}{\sum_{k=1}^{n+1}w(X_k)}\mathbb{P}\{j \in \mathcal{S}(A^w)\} \\
        \mathbb{P}\{n+1 \in \mathcal{S}(A^w)\}\sum_{j=1}^{n+1}\frac{w(X_j)}{\sum_{k=1}^{n+1}w(X_k)} & = \sum_{j=1}^{n+1}p_j^w(X_{n+1})\mathbb{P}\{j \in \mathcal{S}(A^w)\} \\
        \mathbb{P}\{n+1 \in \mathcal{S}(A^w)\} & = \sum_{j=1}^{n+1}p_j^w(X_{n+1})\mathbb{P}\{j \in \mathcal{S}(A^w)\} \\
        & = \mathbb{E}\bigg[\sum_{j\in \mathcal{S}(A^w)}p_j^w(X_{n+1})\bigg] \\
        & \leq 2\alpha
    \end{align*}
    where the last line follows from Step 1.
    
    

    \textit{Step 3: Connection to JAW:} 
    We would now like to connect our strange point result from step 2 to coverage of the JAW prediction interval. Following the approach of \cite{barber2021predictive}, suppose that $Y_{n+1} \not\in \widehat{C}_{n, \alpha}^{\text{JAW}}(X_{n+1})$. Then, either
    \begin{align*}
        Y_{n+1} & > Q_{1-\alpha}^+\big\{p_i^w(X_{n+1})\delta_{\widehat{\mu}_{-i}(X_{n+1}) + R_i^{LOO}}\big\} \\
        & \implies \sum_{i = 1}^{n}p_i^w(X_{n+1})\cdot \mathbbm{1}\big\{Y_{n+1} > \widehat{\mu}_{-i}(X_{n+1}) + R_i^{LOO}\big\}\geq 1 - \alpha
    \end{align*}
    or otherwise
    \begin{align*}
        Y_{n+1} & < Q_{\alpha}^-\big\{p_i^w(X_{n+1})\delta_{\widehat{\mu}_{-i}(X_{n+1}) + R_i^{LOO}}\big\} \\
        & \implies \sum_{i = 1}^{n}p_i^w(X_{n+1})\cdot \mathbbm{1}\big\{Y_{n+1} < \widehat{\mu}_{-i}(X_{n+1}) - R_i^{LOO}\big\}\geq 1 - \alpha
    \end{align*}
    And we can write the union of these two events as 
    \begin{align*}
        1-\alpha \leq & \sum_{i=1}^n p_i^w(X_{n+1})\cdot \mathbbm{1}\big\{Y_{n+1} \not\in \widehat{\mu}_{-i}(X_{n+1}) \pm R_i^{LOO}\big\} \\
        & = \sum_{i=1}^n p_i^w(X_{n+1})\cdot \mathbbm{1}\big\{\big|Y_i - \widehat{\mu}_{-i}(X_i)\big| < \big|Y_{n+1} - \widehat{\mu}_{-i}(X_{n+1})\big|\big\} \\
        & = \sum_{i=1}^{n+1} p_i^w(X_{n+1})\cdot \mathbbm{1}\big\{R_{i, n+1} < R_{n+1, i}\big\} 
    \end{align*}
    from which we see that $n+1\in\mathcal{S}(A^w)$---that is, $n+1$ is a strange point. This result together with the result from Step 2 gives us
    \begin{align*}
        \mathbb{P}\big\{Y_{n+1} \not\in \widehat{C}_{n, \alpha}^{\text{JAW}}(X_{n+1})\big\} & \leq \mathbb{P}\big\{n+1 \in \mathcal{S}(A^w)\big\} \leq 2\alpha \\
        \therefore \ \mathbb{P}\big\{Y_{n+1} \in \widehat{C}_{n, \alpha}^{\text{JAW}}(X_{n+1})\big\} & \geq 1 - 2\alpha
    \end{align*}

\end{proof}

\subsection{Proof of Theorem \ref{thm2}}
\label{prf:2}

\begin{proof}
    First, assume that Assumptions 1 - 4 and Condition 2 from \cite{giordano2019higher} hold uniformly for all $n$ (where $n$ is the number of training points). Then, Proposition 1 from \cite{giordano2019higher} establishes that
    
    \begin{align}
        \max_{i \in [n]} \Big|\Big|\hat{\theta}^{\text{IF-}K}_{-i} - \hat{\theta}_{-i}\Big|\Big|_2 = O_p(N^{-\frac{1}{2}(K + 1)})
    \end{align}
    
    So, for fixed $K$:
    
    \begin{align}
        \lim_{N\rightarrow \infty}\max_{i \in [n]} \Big|\Big|\hat{\theta}^{\text{IF-}K}_{-i} - \hat{\theta}_{-i}\Big|\Big|_2 = O_p(N^{-\frac{1}{2}(K + 1)}) = 0
    \end{align}
    
    Or, for fixed $N$:
    
    \begin{align}
        \lim_{K\rightarrow \infty}\max_{i \in [n]} \Big|\Big|\hat{\theta}^{\text{IF-}K}_{-i} - \hat{\theta}_{-i}\Big|\Big|_2 = O_p(N^{-\frac{1}{2}(K + 1)}) = 0
    \end{align}
    
    Thus, $\hat{\theta}^{\text{IF-}K}_{-i} \rightarrow \hat{\theta}_{-i}$ as either $N \rightarrow \infty$ or $K \rightarrow \infty$. This implies that $\widehat{\mu}^{\text{IF-}K}_{-i} \rightarrow \widehat{\mu}_{-i}$ as either $N \rightarrow \infty$ or $K \rightarrow \infty$ because the model $\widehat{\mu}_{-i}$ is fully determined by its parameters $\hat{\theta}_{-i}$. Therefore, $\widehat{C}_{n, \alpha}^{\text{JAWA-}K}(X_{n+1}) \rightarrow \widehat{C}_{n, \alpha}^{\text{JAW}}(X_{n+1})$ in the limit of $N$ or $K$, and thus by Theorem \ref{thm:1}, $\mathbb{P}\{Y_{n+1} \in \widehat{C}_{n, \alpha}^{\text{JAWA-}K}(X_{n+1})\} \geq 1 - 2\alpha$ as $N \rightarrow \infty$ or $K \rightarrow \infty$. 
    
    Now, separately assume that Assumptions 1 - 4 and Condition 4 from \cite{giordano2019higher} hold uniformly for all $n$. Then, Proposition 3 from \cite{giordano2019higher} gives that 
    
    \begin{align}
        \max_{i \in [n]} \Big|\Big|\hat{\theta}^{\text{IF-}K}_{-i} - \hat{\theta}_{-i}\Big|\Big|_2 = O(N^{-(K + 1)})
    \end{align}
    
    The rest follows from a similar argument as when we assumed Condition 2. 
\end{proof}

\subsection{Proof of Theorem \ref{thm4}}

\label{prf:4}
\begin{proof}
    Recall that we define $\overline{E}$ as
    \begin{align}
        \overline{E} = \big\{y \in \mathbb{R} : \tau^- \leq \widehat{S}(X_{n+1}, y) \leq \tau^+\big\},
        \label{eq:error_event_app}
    \end{align}
    that we assume access to a predictive inference method with prediction sets given by
    \begin{align}
        \widehat{C}_{n, \alpha}^{\text{w-audit}}(X_{n+1}) =  \big\{y \in \mathbb{R} : \ & \widehat{Q}_{\alpha}^-\{p_i^w(X_{n+1})\delta_{V_i^L}\} \leq \widehat{S}(X_{n+1}, y) \leq \widehat{Q}_{1-\alpha}^+\{p_i^w(X_{n+1})\delta_{V_i^U}\}\big\}
        \label{eq:error_method_interval_app}
    \end{align}
    and moreover that we define $\alpha_E^{\text{w-audit}}$ as 
    \begin{align}
        \alpha_E^{\text{w-audit}} = \min\Big(\Big\{\alpha' \ : \ \tau^- \leq \ \widehat{Q}_{\alpha'}^-\{p_i^w(X_{n+1})\delta_{V_i^L}\} \ ,\ \widehat{Q}_{1-\alpha'}^+\{p_i^w(X_{n+1})\delta_{V_i^U}\} \leq \tau^+  \Big\}\Big).
    \label{eq:alpha_I-CovShift_app}
    \end{align}
    Then, if $\alpha_E^{\text{w-audit}}$ exists and $\alpha_E^{\text{w-audit}} < \frac{1 - c_2}{c_1}$, then by construction we can combine \eqref{eq:alpha_I-CovShift_app} with the definition of the prediction set $\widehat{C}_{n, \alpha_E}^{\text{w-audit}}(X_{n+1})$ to obtain
    \begin{align}
        \widehat{C}_{n, \alpha_E}^{\text{w-audit}}(X_{n+1}) =  \Big\{y  : \  \tau^- \leq \widehat{Q}_{\alpha_E}^-\{p_i^w(X_{n+1})\delta_{V_i^L}\} & \leq \widehat{S}(X_{n+1}, y) \nonumber \\
        & \leq \widehat{Q}_{1-\alpha_E}^+\{p_i^w(X_{n+1})\delta_{V_i^L}\}\leq \tau^+\Big\},
        \label{eq:alpha_E_subset}
    \end{align}
    which shows that $\widehat{C}_{n, \alpha_E}^{\text{w-audit}}(X_{n+1})\subseteq \overline{E}$. Thus, $\mathbb{P}\{Y_{n+1} \in \overline{E}\} \geq \mathbb{P}\{Y_{n+1} \in \widehat{C}_{n, \alpha_E}^{\text{w-audit}}(X_{n+1})\}$, and by the coverage guarantee for $\widehat{C}_{n, \alpha_E}^{\text{audit}}(X_{n+1})$ it follows that
    \begin{align}
        \mathbb{P}\{Y_{n+1} \in \overline{E}\} & \geq \mathbb{P}\{Y_{n+1} \in \widehat{C}_{n, \alpha_E}^{\text{w-audit}}(X_{n+1})\} \geq 1 - c_1\alpha_E^{\text{w-audit}} - c_2.
    \end{align}
    Otherwise, $\alpha_E^{\text{w-audit}}$ does not exist or $\alpha_E^{\text{w-audit}} \geq \frac{1 - c_2}{c_1} \implies 1 - c_1 \alpha_E^{\text{audit}} - c_2 \leq 0$. Neither of these cases yield a nontrivial (positive) lower bound for $\mathbb{P}\{Y_{n+1} \in \overline{E}\}$, so for these cases
    \begin{align}
        \mathbb{P}\{Y_{n+1} \in \overline{E}\} & \geq \mathbb{P}\{Y_{n+1} \in \widehat{C}_{n, \alpha_E}^{\text{audit}}(X_{n+1})\} \geq 0.
    \end{align}

\end{proof}

\subsection{Proof of Theorem \ref{thm3}}

\label{prf:3}
\begin{proof}
    The proof proceeds similarly as the proof for Theorem \ref{thm4} in Appendix \ref{prf:4}, except replacing the data-dependent weights $p_i^w(X_{n+1})$ with the uniform weights $\frac{1}{n+1}$. 

\end{proof}

\newpage

\section{Additional experimental details and analysis}
\label{app:analysis}

\subsection{Creation of covariate shift}
\label{app:cov_shift_details}

To induce covariate shift, test points were sampled from the set of points not used for training with exponential tilting weights such that the total number of test points was equal to half the number of points not used for training. For the relatively lower dimensional airfoil and wine datasets, the weights took the form $w(x) = \exp(x^{T}\beta)$, while for the relatively higher dimensional datasets the weights took the form $w(x) = \exp(x_{\text{PCA}}^{T}\beta)$ where $x_{\text{PCA}}$ is some PCA-based representation of the covariates data $x$. 

\begin{figure}[hbt]
    \includegraphics[width=\textwidth]{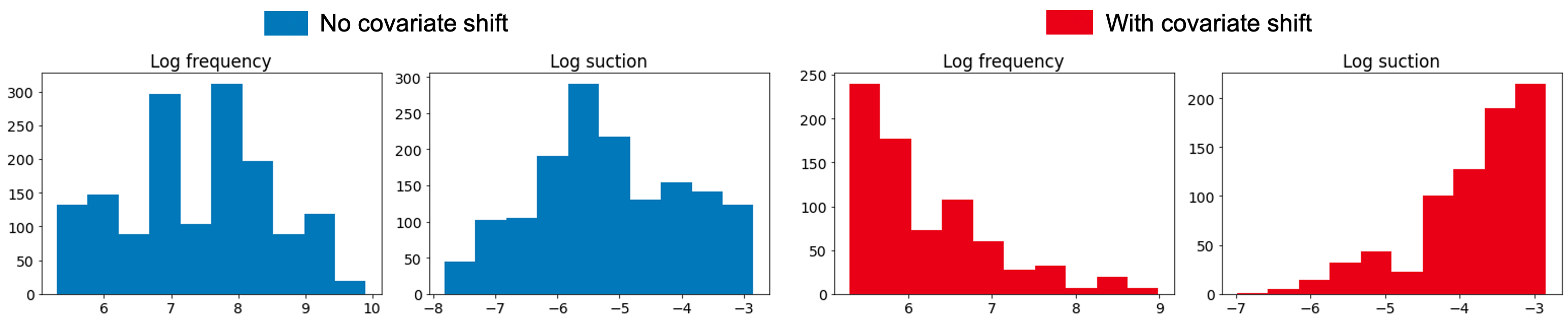}
\caption{Distribution log frequency and log suction features of airfoil dataset before and after exponential tilting.}
\label{fig:shift}
\end{figure}

Figure \ref{fig:shift} shows the distribution first and last  features in the airfoil dataset before and after the exponential tilting is applied to induce covariate shift with parameter $\beta = (-1, 0, 0, 0, 1)$. In our main experiments, exponential tilting parameters were selected for each dataset so that the associated covariate shift would result in a similar loss in how informative the training set is regarding the biased test set, as assess by reduced effective sample size. 

Specifically, for a training set size of 200 points for each dataset, the exponential tilting parameters were selected and tuned so that the estimated effective sample size of the training data was reduced to approximately 50, averaged across 1000 random train-test splits. For training data $X_1, ..., X_n$ and likelihood ratio $w$, the effective sample size was estimated using the following commonly-used heuristic $\widehat{n} = [\sum_{i=1}^n|w(X_i)|]^2 / \sum_{i=1}^n|w(X_i)|^2$  \citep{gretton2009covariate, reddi2015doubly, tibshirani2019conformal}.

The specific selections of $\beta$ that resulted in approximately $\widehat{n} = 50$ for each dataset are as follows. For the airfoil dataset, unless otherwise specified the tilting parameter was $\beta_{\text{airfoil}} = (-0.85, 0, 0, 0, 0.85)$, which induced covariate shift such that points with low values of the first feature and high values of the last feature were more likely to appear in the test distribution (see Figure \ref{fig:shift}). The wine dataset was similarly tilted using the first and last components, with a tilting parameter of $\beta_{\text{wine}} = (-0.53, 0, ..., 0, 0.53)$. The wave dataset is composed of 48 total features, of which the first 32 features are latitude and longitude values, and where the remaining 16 features are absorbed power values. Accordingly, the first principal component of only the 32 location features was used for tilting along with the first principal component of only the 16 absorbed power values, with a tilting parameter of $\beta_{\text{wave}} = (-0.0000925, 0.0000925)$ unless otherwise specified. For the superconductivity dataset, only the first principal component of all of the data was used for tilting, with tilting parameter $\beta_{\text{superconduct}} = 0.00062$. Lastly, for the communities and crime dataset, the first two principal components of the whole dataset were used with tilting parameter $\beta_{\text{communities}} = (-0.825, 0.825)$.

\subsection{Models}
\label{app:models}

For all experiments with JAW and its baselines we used two different regression predictors $\widehat{\mu}$: random forests (scikit-learn RandomForestRegressor), and neural networks (scikit-learn MLPRegressor with LBFGS optimizer, logistic activation, and default parameters otherwise).

For all experiments comparing the coverage and interval width of JAWA to other influence function approximated baselines, we used a neural network predictor with one hidden layer consisting of 25 hidden units. Covariate and label data were centered and scaled. The neural network was trained for 2000 epochs with batch sizes of 50 and a learning rate of 0.0001, which generally resulted in convergence. The objective function for the neural network in JAWA is the negative log likelihood with a Gaussian prior or L2 regularization term. The L2 regularization was added to satisfy assumptions for computing IFs described in \cite{giordano2019higher} and due to empirical findings of first-order IFs for neural networks requiring regularization for reliable results \citep{basu2020influence}. The L2 regularization $\lambda$ parameter was tuned using a grid search prior to all experiments using a ``tuning'' validation set of 200 samples that were excluded from both the training and test sets in the experiments (see Appendix \ref{app:L2} for more details regarding the L2 regularization tuning).

\subsection{Comparison of coverage variance for JAW and weighted split}
\label{app:coverage_variance_comparison}

\begin{table}[ht]
\caption{Coverage variance for JAW and weighted split conformal prediction, averaged across 1000 experimental replicates (i.e., statistics are the variance of all of the 1000 mean coverage statistics, one for each experiment). Lower coverage variances indicate more reliable coverage. The coverage variance for JAW is lower than that of weighted split conformal prediction in all datasets and predictor conditions due to JAW avoiding the sample splitting required by weighted split.}
\label{tab:coverage_variance}
\footnotesize
\centering
\begin{tabular}{|c | c | c | c | c| c | c | c | c | c| c|} 
\hline
\textbf{Dataset} & \multicolumn{2}{c|}{\textbf{Airfoil}} & \multicolumn{2}{c|}{\textbf{Wine}} & \multicolumn{2}{c|}{\textbf{Wave}} & \multicolumn{2}{c|}{\textbf{Superconduct}} & \multicolumn{2}{c|}{\textbf{Communities}} \\
\cline{1-11}
\textbf{Method} & \textbf{NN} & \textbf{RF} & \textbf{NN} & \textbf{RF} & \textbf{NN} & \textbf{RF} & \textbf{NN} & \textbf{RF} & \textbf{NN} & \textbf{RF}  \\ 
\hline
Weighted split & 0.0022 & 0.0023 & 0.0019 & 0.0017 & 0.0030 & 0.0029 & 0.0040 & 0.0035 & 0.00194 & 0.0021 \\
\hline
JAW & 0.0010 & 0.0019 & 0.0013 & 0.0015 & 0.0005 & 0.0014 & 0.0021 & 0.0030 & 0.00189 & 0.0014 \\ 
\hline

\end{tabular}
\end{table}



\subsection{Additional AUC results}
\label{app:auc_additional}

Due to space constraints, in the main paper we only report error assessment AUC results for the random forest predictor condition. For completeness, here we now present error assessment results for the neural network predictor, which are similar:

\begin{figure}[ht]
 \setlength{\tabcolsep}{1pt}
\begin{tabular}{ccccc}
    \multicolumn{3}{c}{\includegraphics[width=0.5\textwidth]{figures/AUC_key.png}} \\
    \includegraphics[width=0.2\textwidth]{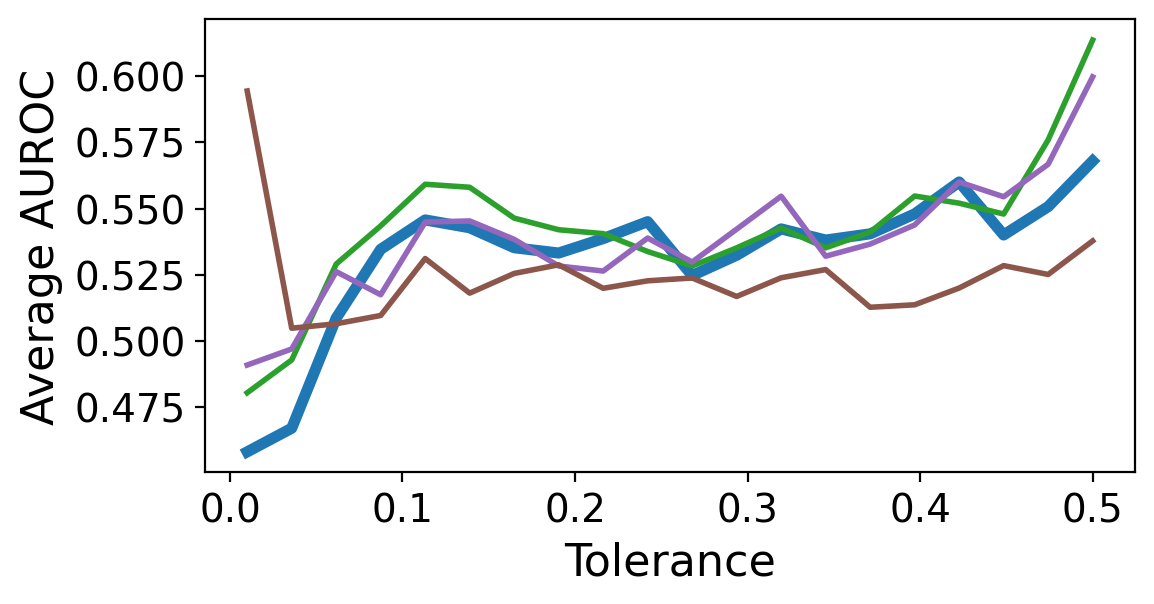} & 
    \includegraphics[width=0.2\textwidth]{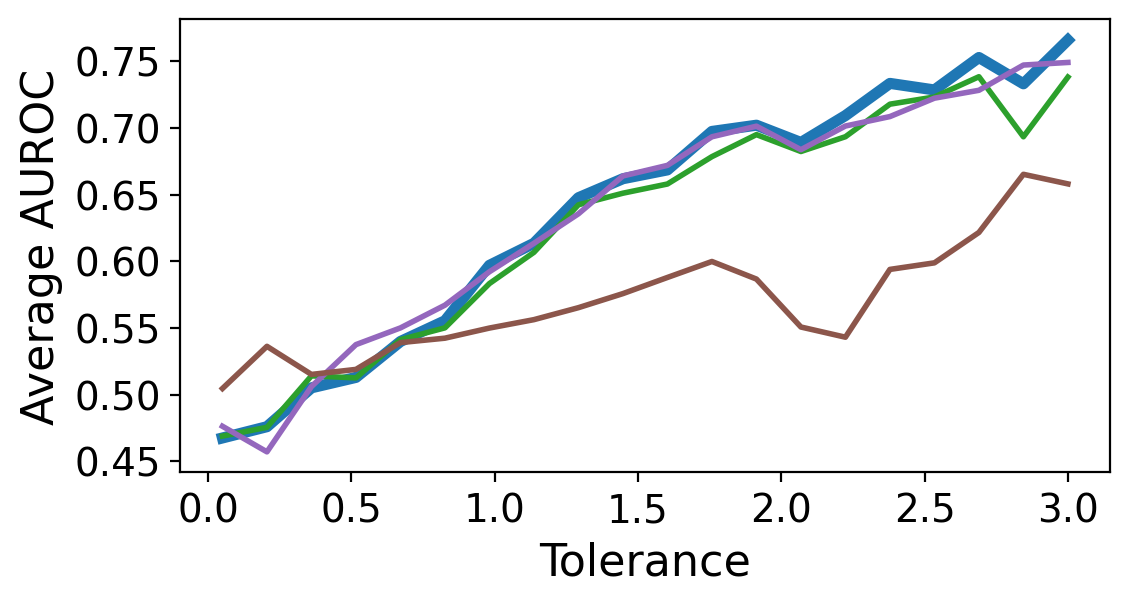} & 
    \includegraphics[width=0.2\textwidth]{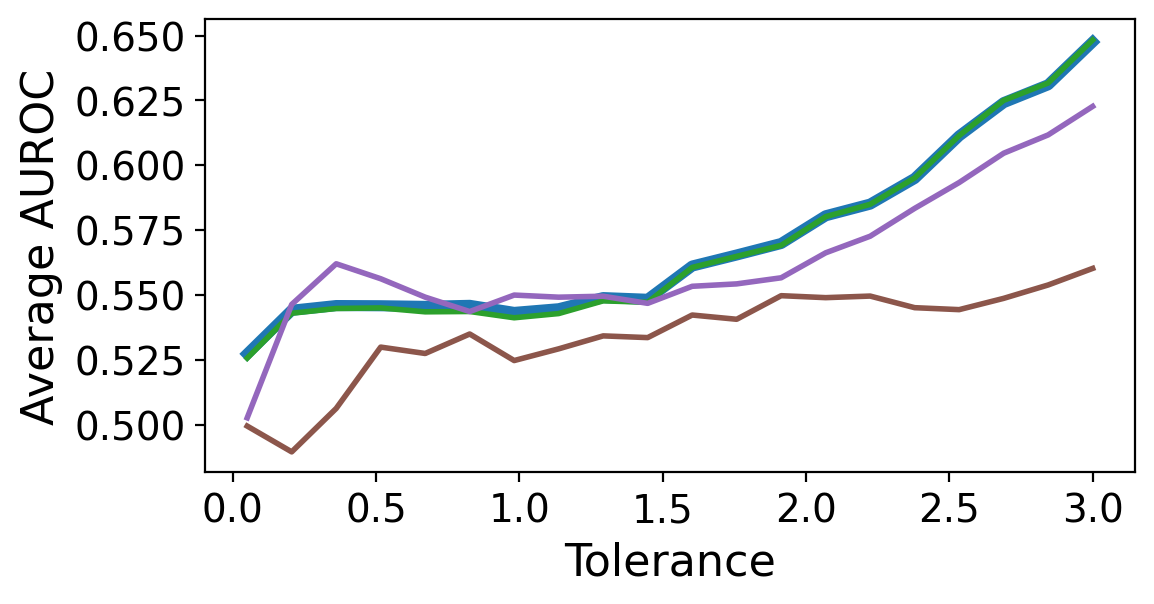} & 
    \includegraphics[width=0.2\textwidth]{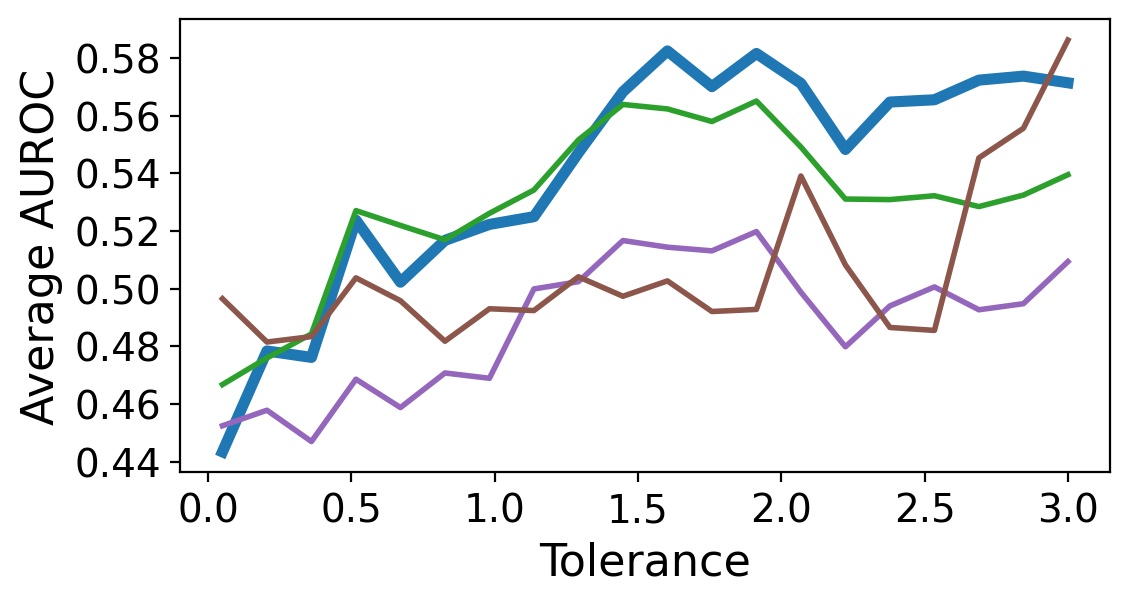} & 
    \includegraphics[width=0.2\textwidth]{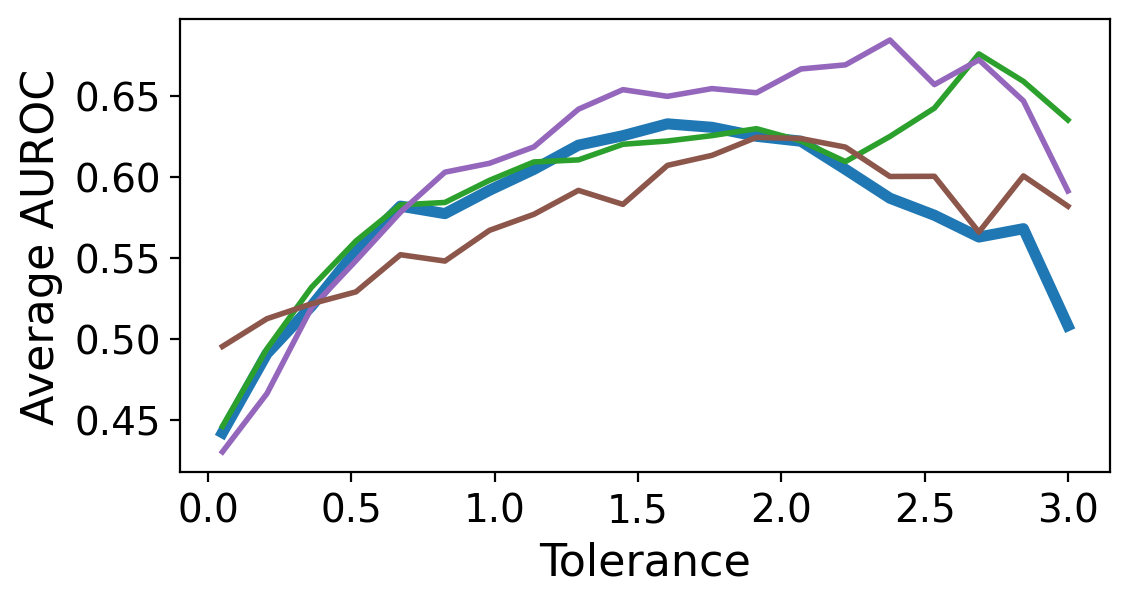}\\
     (a) Airfoil & (b) Wine & (c) Wave & (d) Superconduct & (e) Communities
    \end{tabular}
    \caption{AUROC values for different tolerance levels $\tau$ across the three datasets for the random forest predictor, averaged across 5 experiment replicates, with random forest $\widehat{\mu}$. See Appendix \ref{app:auc_additional} for additional experiments with neural network predictor.} 
    \label{fig:auc_app}
\end{figure}

\subsection{JAW-E: JAW with estimated weights}
\label{app:jaw_e}

JAW assumes access to oracle likelihood ratio weights, but that in practice this information is often not available. In such cases, the likelihood ratios can be estimated through an approach such as probabilistic classification, moment matching, or minimization of $\phi$-divergences (for a review of likelihood ratio estimation approaches see \cite{sugiyama2012density}). We refer to JAW with \textbf{E}stimated likelihood ratio weights as JAW-E to acknowledge that the method's coverage performance will depend on the quality of the likelihood ratio estimates.

The following experiments compare coverage histograms of JAW with oracle likelihood-ratio weights those of JAW-E with weights estimated from probabilstic classification. We follow the approach used in \cite{tibshirani2019conformal} for estimating the likelihood-ratio weights using logistic regression and random forest classifiers. Specifically, for training covariate data $X_1, ..., X_n$ and test covariate data $X_{n+1}, ..., X_{n+m}$ where $C_i = 0$ for $i = 1, ..., n$ and $C_i = 1$ for $i = n+1, ..., n+m$, the conditional odds ratio $\mathbb{P}(C=1 | X = x) / \mathbb{P}(C=0 | X = x)$ can be used as an equivalent substitute to the likelihood ratio weight function $w(x)$ due to the normalization of the weights for use in JAW. Thus, for an estimate $\widehat{p}(x) \approx \mathbb{P}(C=1 | X = x)$ obtained from a classifier such as logistic regression or random forest, then we can use the following estimated weight function in place of likelihood-ratio weights:

\begin{align}
    \widehat{w}(x) = \frac{\widehat{p}(x)}{1-\widehat{p}(x)}.
\end{align}

\begin{figure}[ht]
    \setlength{\tabcolsep}{1pt}
    \begin{tabular}{cccccc}
    \multicolumn{5}{r}{\includegraphics[width=\textwidth]{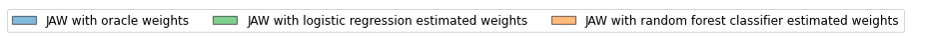}} \\
    \includegraphics[width=0.2\textwidth]{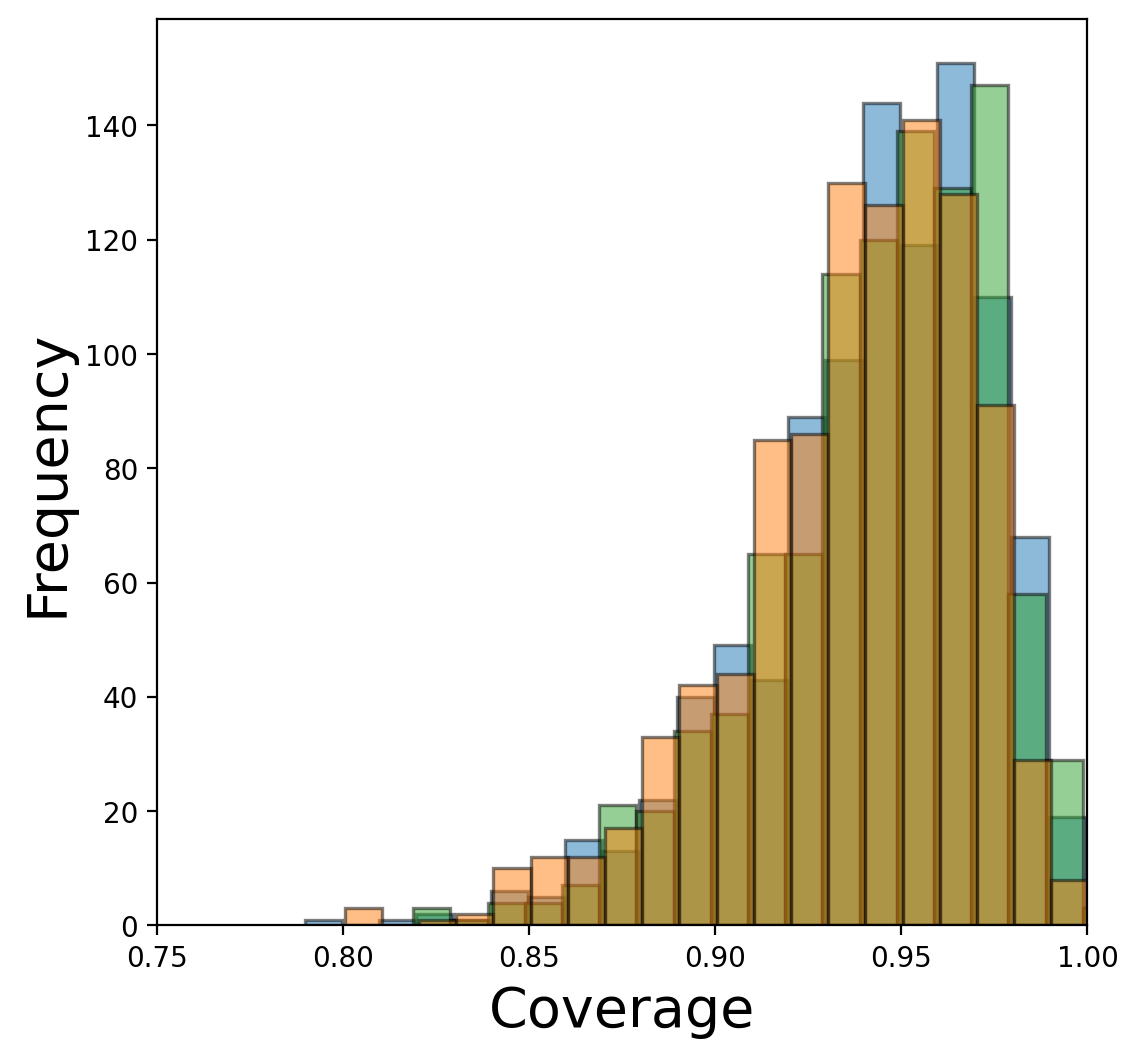}&  \includegraphics[width=0.2\textwidth]{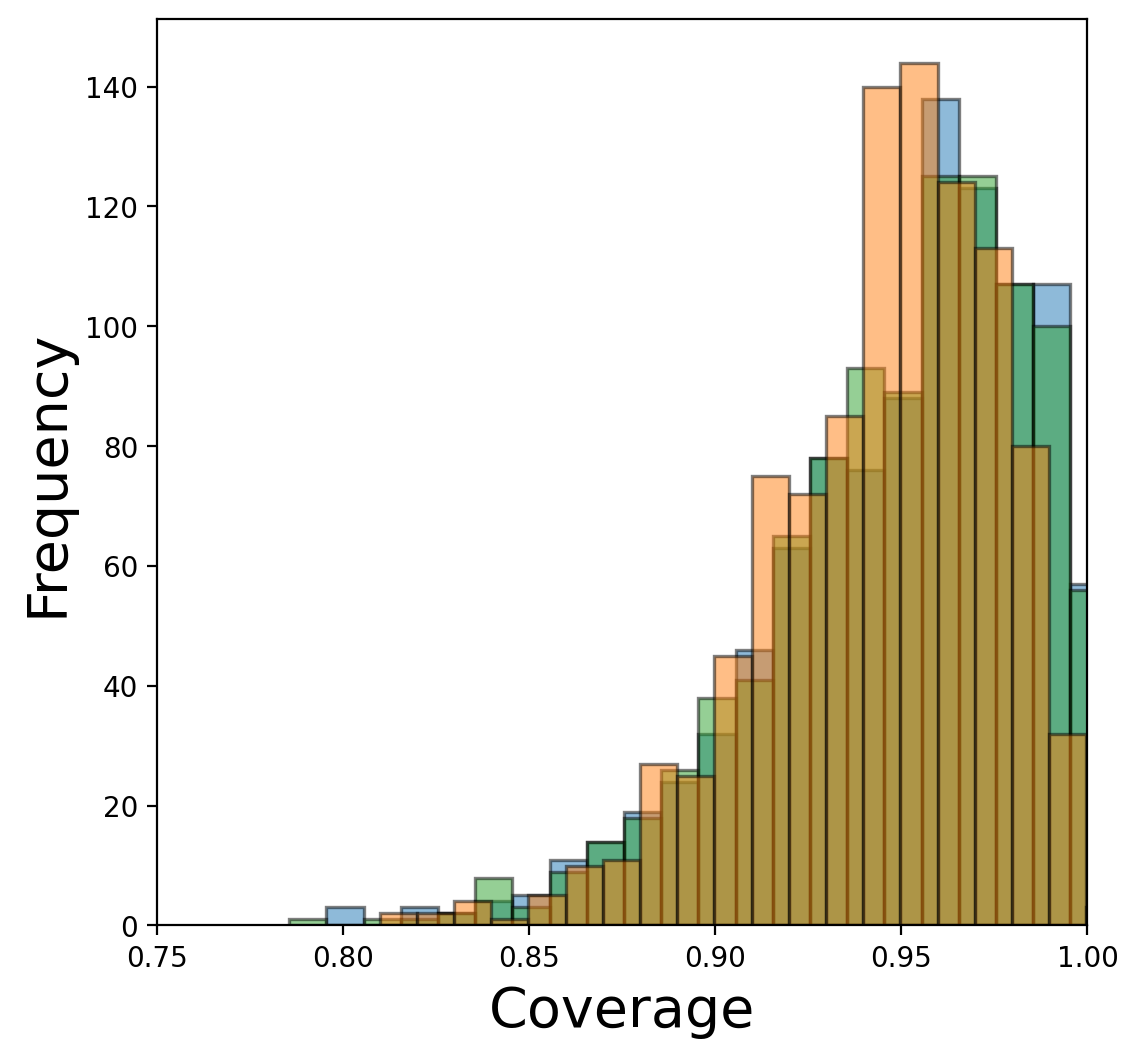} & \includegraphics[width=0.2\textwidth]{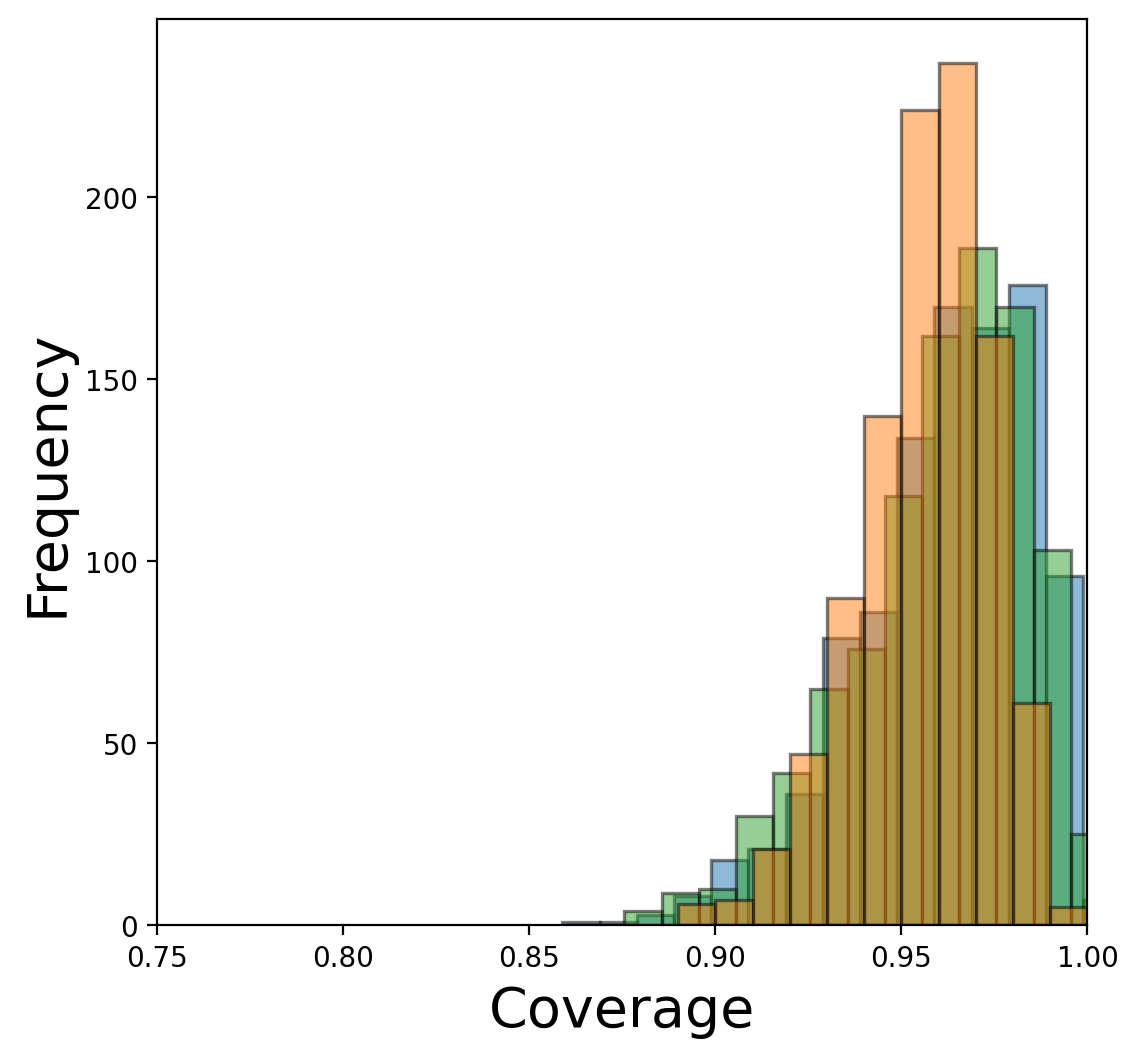} & \includegraphics[width=0.2\textwidth]{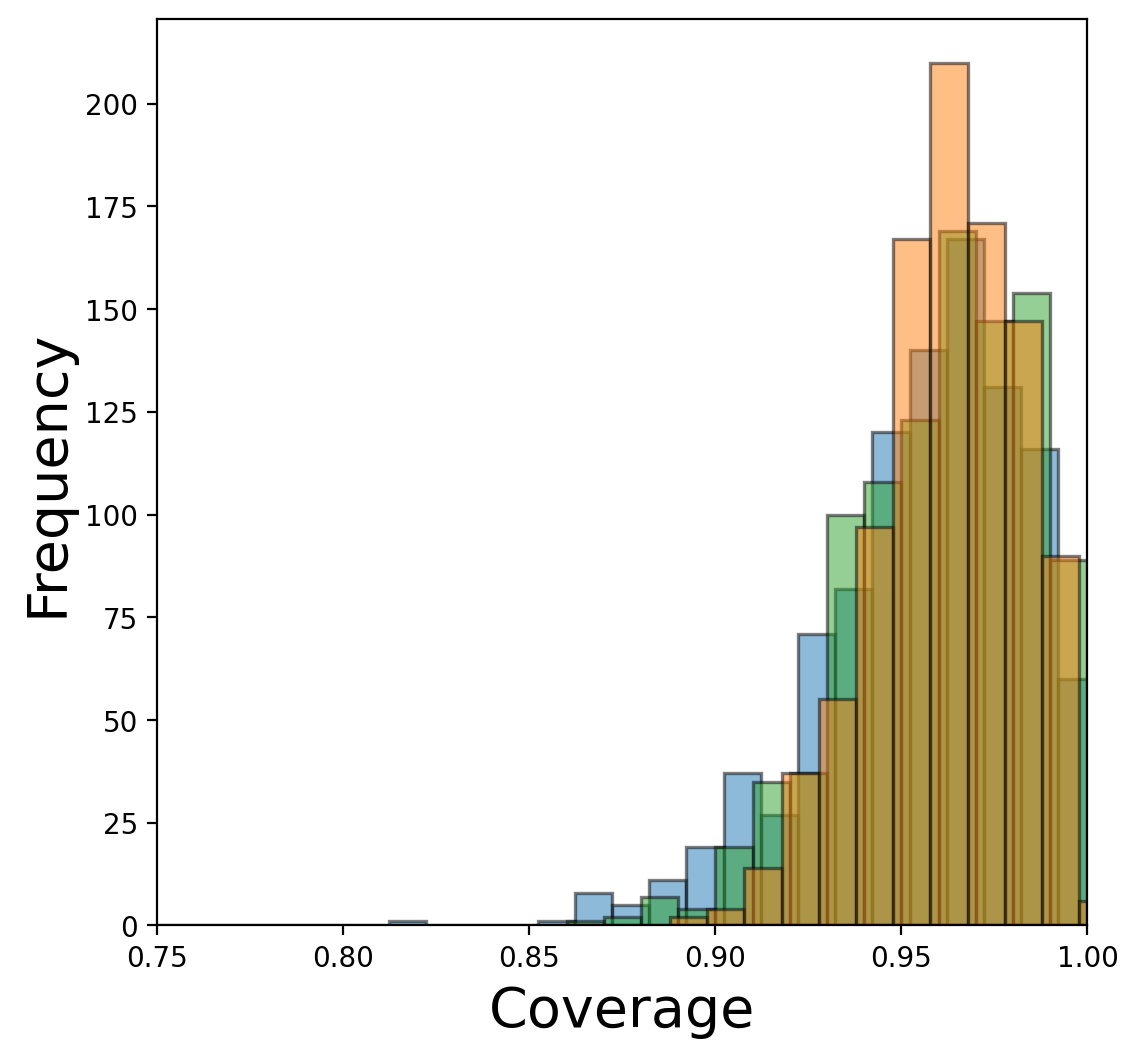} &  \includegraphics[width=0.2\textwidth]{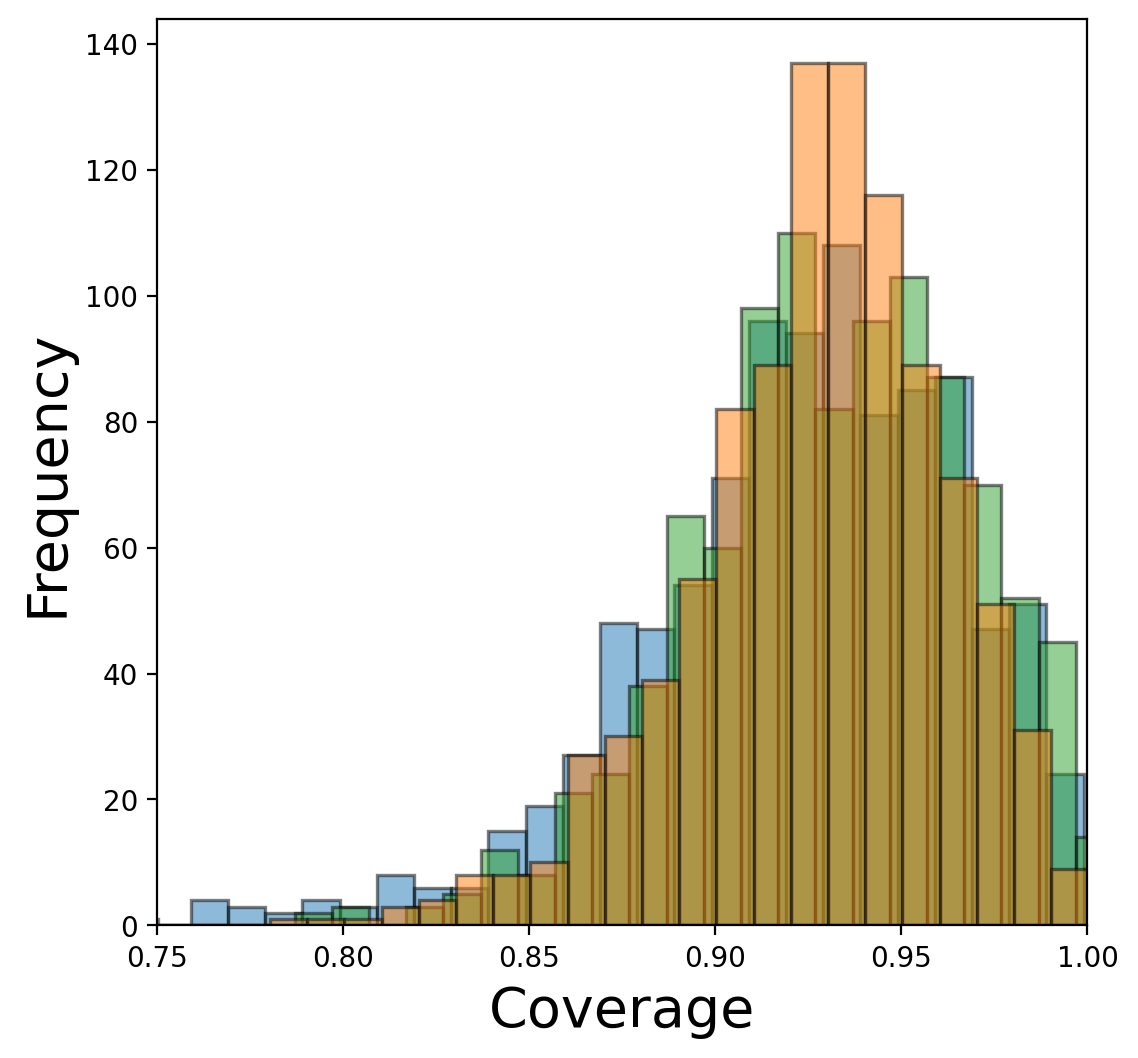} \\
    (a) Airfoil & (b) Wine & (c) Wave & (d) Superconduct & (e) Communities
    \end{tabular}
    \caption{Comparison of JAW coverage under covariate shift with oracle versus JAW-E coverage with estimated likelihood ratio weights for neural network predictor across all datasets. Blue is oracle weights, green is weights estimated with logistic regression, and orange is weights estimated with random forest classifier. Histograms represent 1000 experimental replicates.}
    \label{fig:JAW-E}
\end{figure}

Figure \ref{fig:JAW-E} illustrates the coverage performance of JAW-E, or JAW with estimated likelihood ratio weights, compared to JAW for weights estimated using both logistic regression and random forest classifiers as described in Section \ref{app:models}. Results are for both neural network and random forest regression predictors across all five UCI datasets. We observe that the coverage histograms for JAW-E with both weight estimation methods are largely directly overlapping with the coverage histogram for JAW with oracle weights. These results demonstrate the applicability of JAW-E for predictive inference under covariate shift when the true likelihood ratio is not known but can be estimated from the data. 

\subsection{Ablation studies on shift magnitudes}
\label{app:ablation}

\begin{figure}[ht]
    \setlength{\tabcolsep}{1pt}
    \begin{tabular}{cccc}
    \multicolumn{4}{c}{\includegraphics[width=0.2\textwidth]{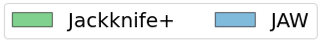}} \\
    \includegraphics[width=0.25\textwidth]{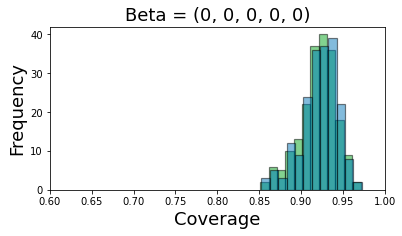}&
    \includegraphics[width=0.25\textwidth]{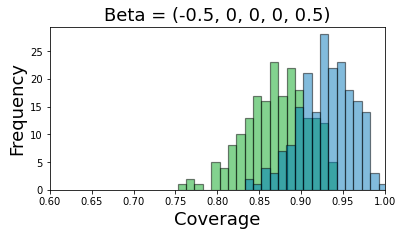}&
    \includegraphics[width=0.25\textwidth]{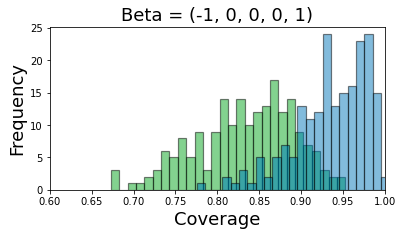} &
    \includegraphics[width=0.25\textwidth]{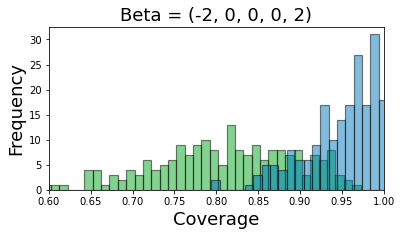}
    \end{tabular}
    \caption{JAW performance compared to jackknife+ on the airfoil dataset with random forest $\widehat{\mu}$ function, under increasing magnitude of covariate shift (different $\beta$ values), with 200 replicates.}
    \label{fig:coverage_shift_magnitudes}
\end{figure}

We demonstrate the effect of different magnitudes of covariate shift by comparing the coverage performance of JAW and the jackknife+ on the airfoil dataset with different magnitudes of the exponential tilting bias parameter $\beta$. Informed by these experiments depicted in Figure \ref{fig:coverage_shift_magnitudes}---where JAW's mean coverage remains consistent but the variance in coverage increases with increased covariate shift magnitude---we performed additional experiments to investigate the potential cause of JAW's increased variance. Specifically, we compare histograms of JAW's coverage at a fixed covariate shift magnitude to that of jackknife+ without covariate shift but with reduced ``effective sample size'', which is known to be reduced by likelihood ratio weighting. \cite{tibshirani2019conformal} made a similar comparison between weighted split conformal prediction under covariate shift and standard split conformal prediction with reduced effective sample size, and we use the same heuristic for effective sample size estimation \citep{gretton2009covariate, reddi2015doubly} (which we also used for selecting exponential tilting parameter values for each dataset in Figure \ref{fig:coverage_width}): 

\begin{align}
    \widehat{n} = \frac{[\sum_{i=1}^n|w(X_i)|]^2}{\sum_{i=1}^n|w(X_i)|^2} = \frac{||w(X_{1:n})||_1^2}{||w(X_{1:n})||_2^2}. \nonumber
\end{align}

\textbf{Effect of different magnitudes of covariate shift} As shown in Figure \ref{fig:coverage_shift_magnitudes}, the extent of covariate shift can be controlled by modifying a parameter in the exponential tilting weights so that weights are are more or less drastic. When the bias parameter is set to 0 this corresponds to no bias or IID train and test data. We can see that JAW is robust to different amounts of covariate shift, generating high coverage even under high level of shift. 

\begin{figure}[ht]
    \begin{center}
    \includegraphics[width=0.5\textwidth]{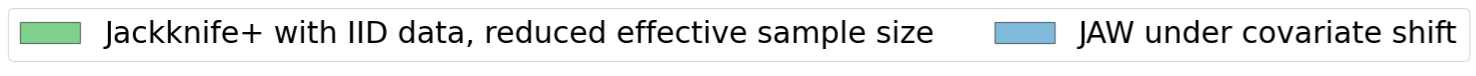} \\
    \begin{tabular}{cc}
    \includegraphics[width=0.2\textwidth]{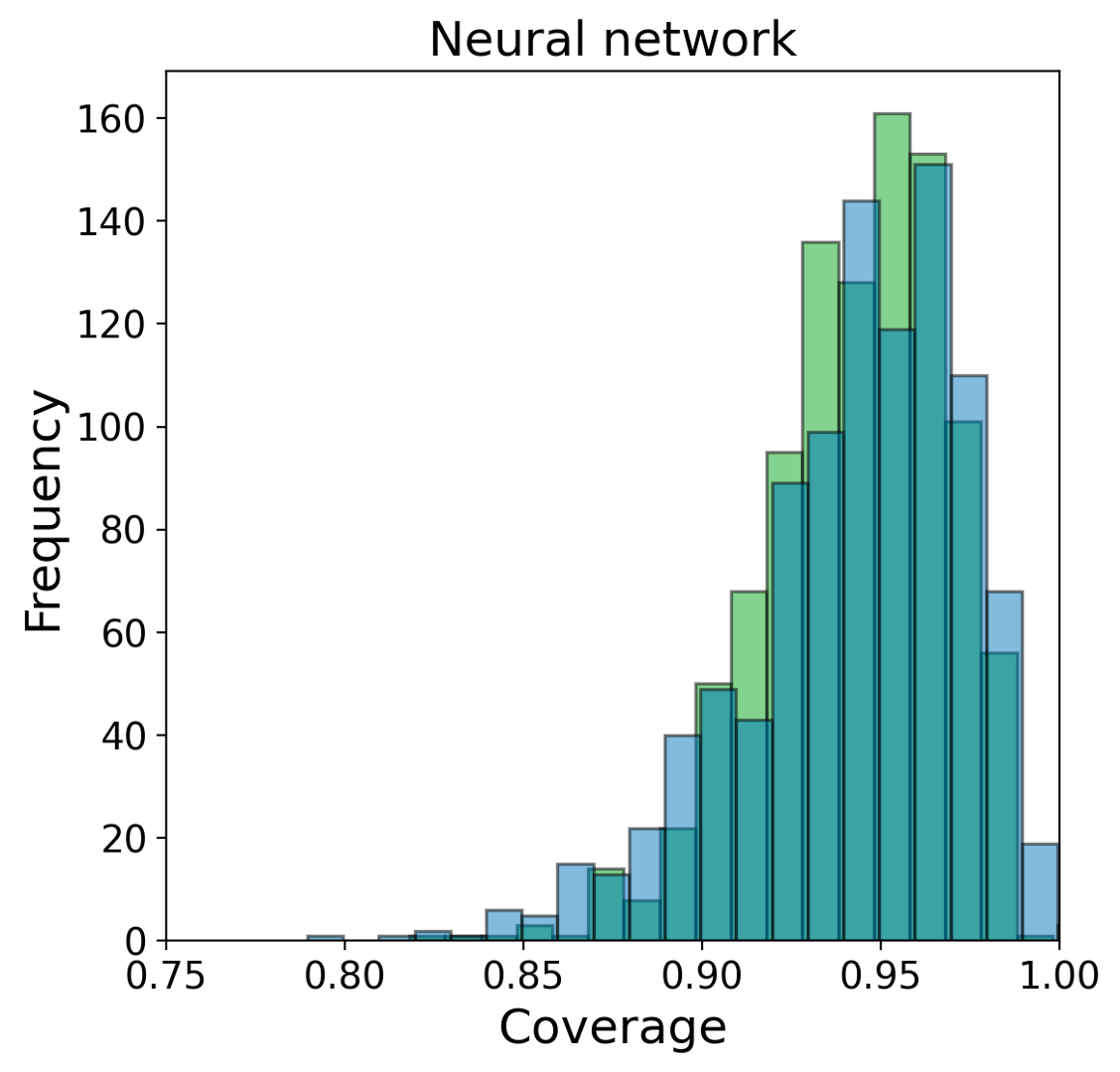}&
    \includegraphics[width=0.2\textwidth]{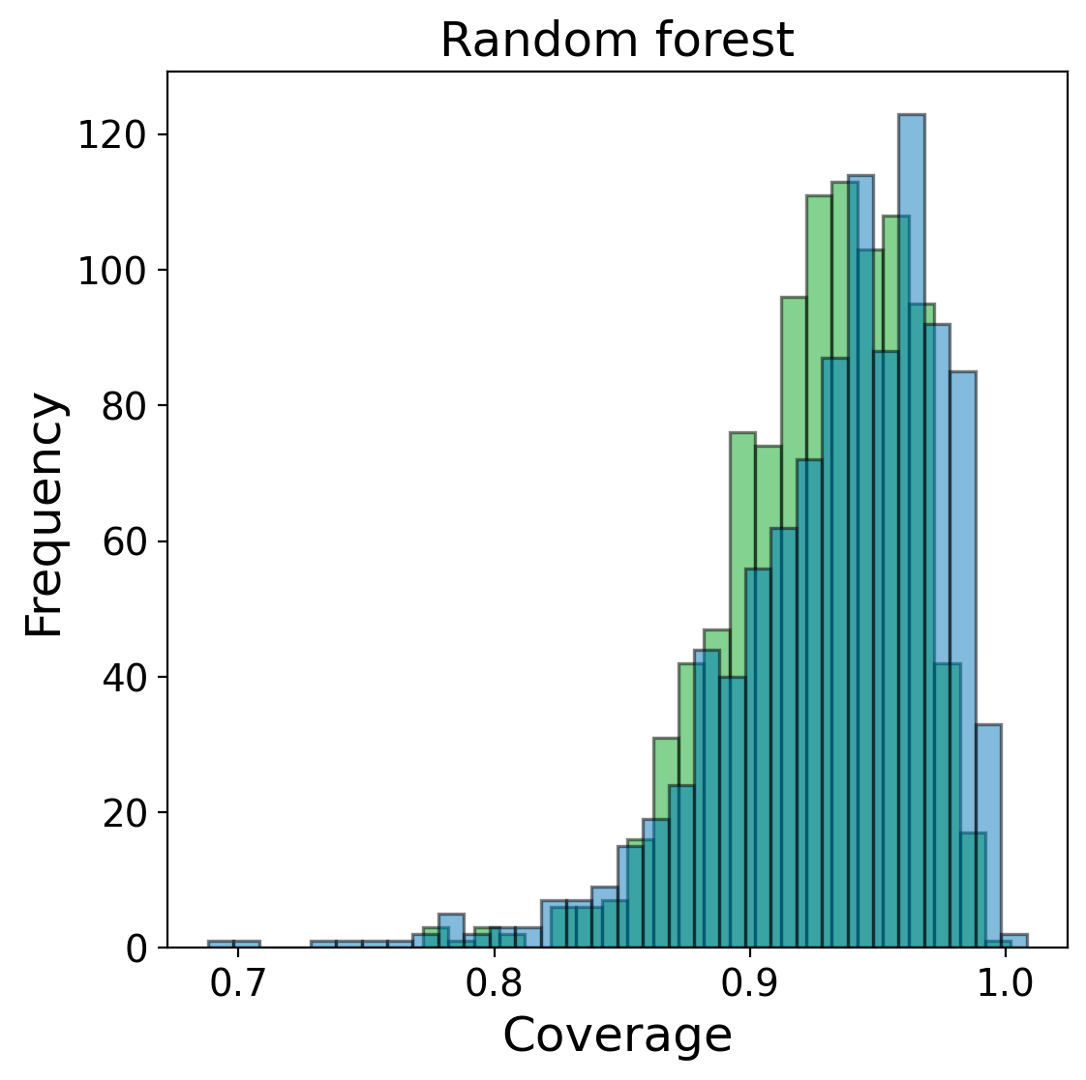}\\
    \end{tabular}
    \end{center}
    \caption{Comparison of JAW coverage histogram under covariate shift (blue) to jackknife+ coverage histogram (green) with no covariate shift but reduced effective sample size corresponding to the magnitude of covariate shift that JAW is evaluated on. Experiments are for both neural network (left) and random forest (right) predictors on the airfoil dataset, with 1000 experimental replicates. The largely overlapping histograms suggests that the increase in JAW coverage variance observed in Figure \ref{fig:coverage_shift_magnitudes} is largely due to the decrease in effective sample size inherent to likelihood ratio weighting.}
    \label{fig:effective_SS}
\end{figure}

\textbf{Reduced effective sample size accounts for JAW increase in coverage variance under shift}
While JAW's mean coverage remains relatively consistent under different magnitudes of covariate shift as seen in Figure  \ref{fig:coverage_shift_magnitudes}, we also observe that the variance in coverage is higher for higher levels of shift. We hypothesized that this increase in variance is due to the high variance issue associated with important weighting methods that is well known \citep{reddi2015doubly,li2020robust} in the literature. We evaluate this hypothesis with effective sample size experiments reported in Figure \ref{fig:effective_SS} that compare a histogram of JAW's coverage under covariate shift with the coverage of jackknife+ with IID data but reduced effective sample size corresponding to the magnitude of covariate shift that JAW is evaluated on (see Appendices \ref{app:cov_shift_details} and \ref{app:ablation} for details). In Figure \ref{fig:effective_SS} we see that the coverage histogram for JAW under covariate shift is nearly directly overlapping with the histogram for jackknife+ coverage with no shift but reduced effective sample size. This result suggests that the reduction of effective sample size due to likelihood ratio weighting is largely if not entirely responsible for the increase in JAW coverage variance for increased shift magnitudes. We leave the variance reduction of our work to the future work.

\subsection{Empirical runtime of JAWA compared to JAW}
\label{app:runtime}

Whereas JAW requires retraining $n$ leave-one-out models, JAWA does not require any retraining, and thus generally enjoys significantly faster runtime than JAW. In Table \ref{tab:runtime} we report the empirical runtime of JAW compared to JAWA for different orders of JAWA's influence function approximation. In these experiments, JAWA is orders of magnitude faster than JAWA regardless of whether the influence function approximation is first, second, or third order (though of course the specific runtime statistics depend on the model architecture, optimization scheme, or dataset). JAWA's runtime does not increase substantially (relative to JAW's runtime) with increased influence function orders for $K \in \{1, 2, 3\}$.

\begin{table}[ht]
\caption{Example empirical comparison between the runtime for JAW and JAWA-$K$ for different influence function approximation orders $K \in \{1, 2, 3\}$ for the neural network predictor used in the JAWA experiments (see Appendix \ref{app:models}), rounded up to the nearest second. This runtime experiment was performed on an 8-core personal computer with 32 GB of memory.}
\label{tab:runtime}
\centering
\begin{tabular}{|c | c | c | c | c | c| } 
\hline
\textbf{Method} & \textbf{Airfoil} & \textbf{Wine} & \textbf{Wave} & \textbf{Superconduct} & \textbf{Communities} \\
\hline
JAW    & 58 min, 39 s& 59 min, 18 s & 1 hr, 24 min, 24 s & 1 hr, 26 min, 53 s & 1 hr, 25 min, 42 s \\
\hline
JAWA-1 & 1 s & 2 s & 4 s & 7 s & 8 s \\
\hline
JAWA-2 & 3 s & 4 s & 6 s & 11 s & 14 s \\
\hline
JAWA-3 & 11 s & 12 s & 16 s & 21 s & 23 s \\
\hline
\end{tabular}
\end{table}

\subsection{L2 regularization for JAWA experiments}
\label{app:L2}

For the experiments involving JAWA and its baselines, the following L2 regularization tuning procedure was used for the neural network described in the second paragraph of \ref{app:models}. The grid search evaluated the coverage of the first-order influence function approximation of the jackknife+ at different values of the regularization tuning parameter $\lambda \in \{0.5, 1, 2, 4, 8, 16, 32, 64, 96, 128\}$ for 10 train-test splits among all data for a dataset aside from the holdout tuning set. The smallest value of $\lambda$ in the grid search for which the coverage of the first-order influence function approximation of the jackknife+ exceeded 0.875 was used. The coverage calibration threshold of 0.875 was selected because the change in coverage due to increased $\lambda$ appeared to plateau just above or below the target coverage rate of 0.9 for each dataset, so setting the threshold slightly below 0.9 can help avoid over-regularizing. See \cite{angelopoulos2020uncertainty} for a discussion of calibrating uncertainty estimation in conformal prediction. This grid search procedure identified a separate $\lambda$ regularization parameter for each dataset: $\lambda_{\text{air}} = 1, \lambda_{\text{win}} = 8, \lambda_{\text{wav}}= 4, \lambda_{\text{sup}} = 96, \lambda_{\text{com}} = 64$. Additionally, we also added a dampening term to the Hessian (for IFs computation) as in \cite{koh2017understanding} so that the smallest eigenvalue of the Hessian was at least 0.5.

\begin{figure}[ht]
    \setlength{\tabcolsep}{1pt}
    \begin{tabular}{ccccc}
    \includegraphics[width=0.2\textwidth]{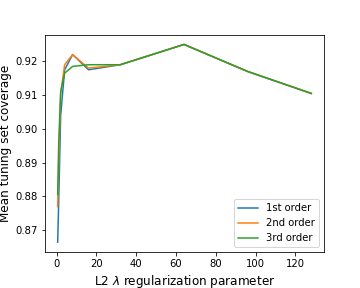}&  \includegraphics[width=0.2\textwidth]{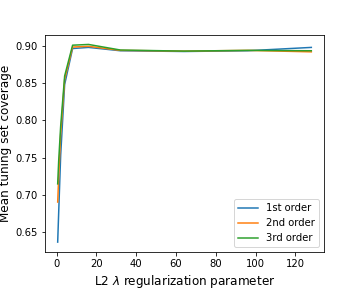} & \includegraphics[width=0.2\textwidth]{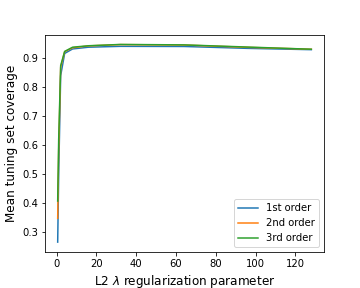} & \includegraphics[width=0.2\textwidth]{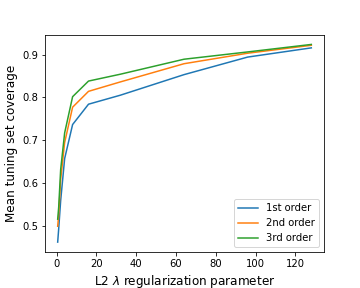} &  \includegraphics[width=0.2\textwidth]{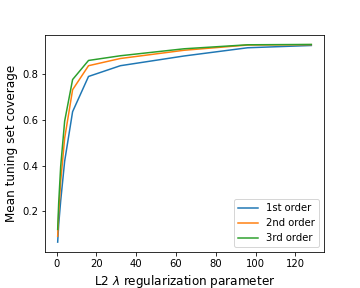} \\
    (a) Airfoil & (b) Wine & (c) Wave & (d) Superconduct & (e) Communities
    \end{tabular}
    \caption{Grid search plots for tuning the $\lambda$ L2 regularization parameter for influence function coverage experiments. All experiments are done with 1st, 2nd, and 3rd order influence function approximations of the jackkinfe+ (denoted in blue, orange, and green lines in the figure). The y-axis for each plot is the average coverage on the tuning dataset for each L2 regularization parameter $\lambda \in \{0.5, 1, 2, 4, 8, 16, 32, 64, 96, 128\}$.}
    \label{fig:grid_search}
\end{figure}

\subsection{Histogram comparison of jackknife+ and JAW coverage under covariate shift}
\label{app:histogram_coverage_comparison}

\begin{figure}[ht]
\centering
    \includegraphics[width=0.4\textwidth]{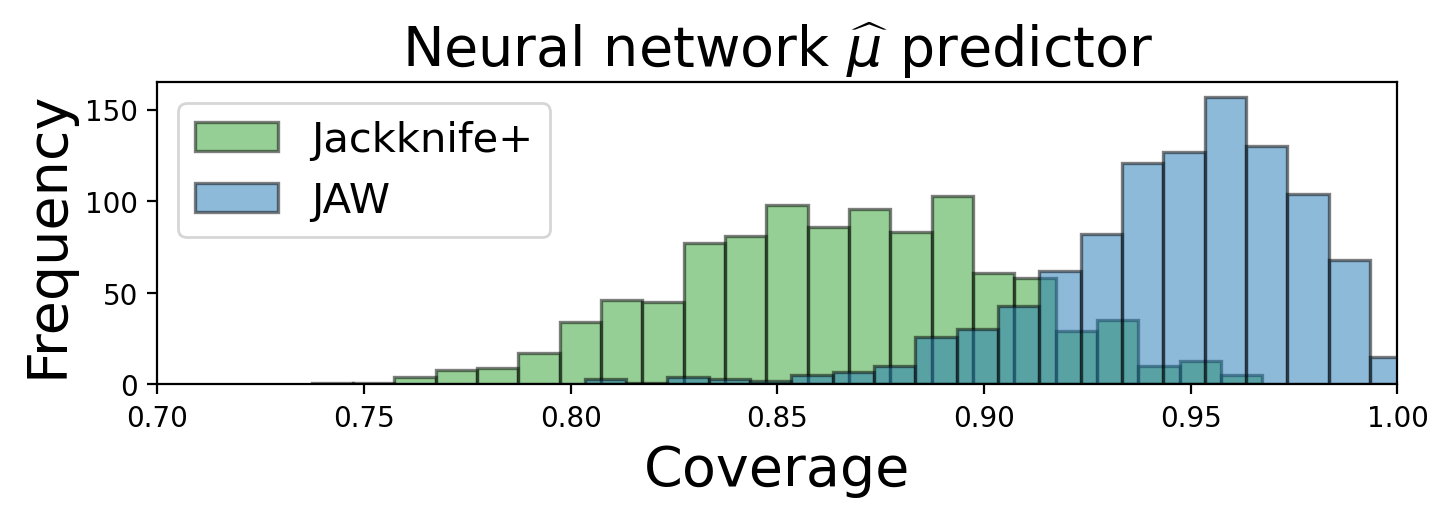}
    \includegraphics[width=0.4\textwidth]{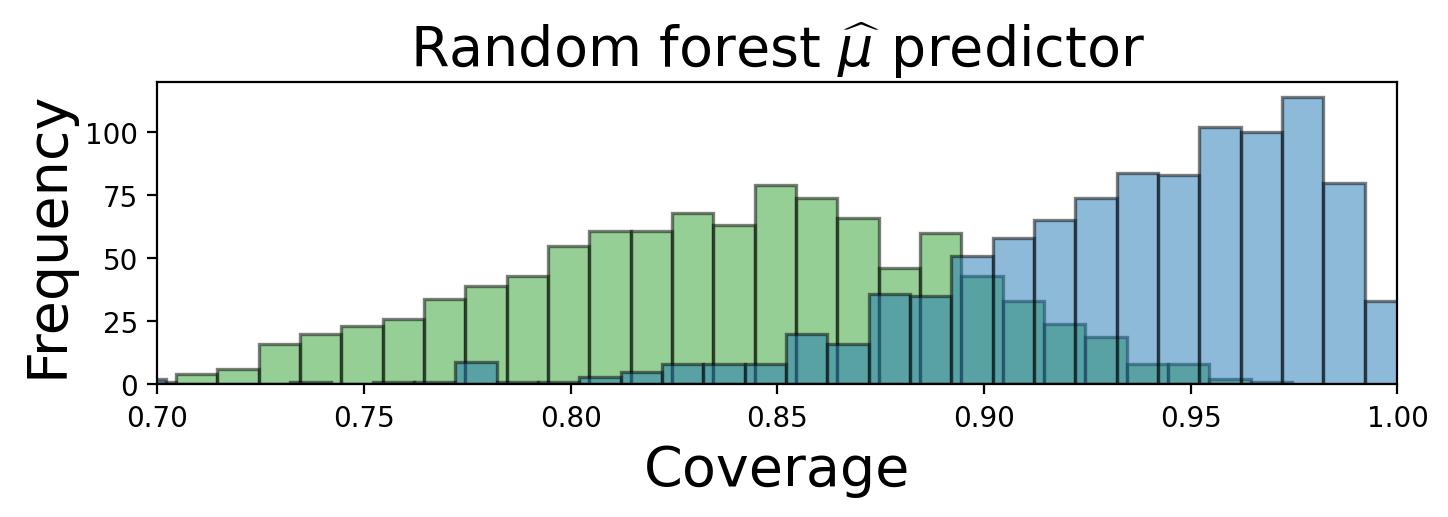}
    \caption{Jackknife+ versus JAW coverage under covariate shift for the airfoil dataset, when $\beta = (-1, 0, 0, 0, 1)$, for 1000 replicates. JAW maintains the high coverage under covariate shift.}
    \label{fig:jaw_hist}
\end{figure}

\subsection{Cases where jackknife+ may not lose coverage} 
\begin{figure}[thbp]
 \setlength{\tabcolsep}{1pt}
 \centering
\begin{tabular}{cc}
    \includegraphics[width=0.45\textwidth]{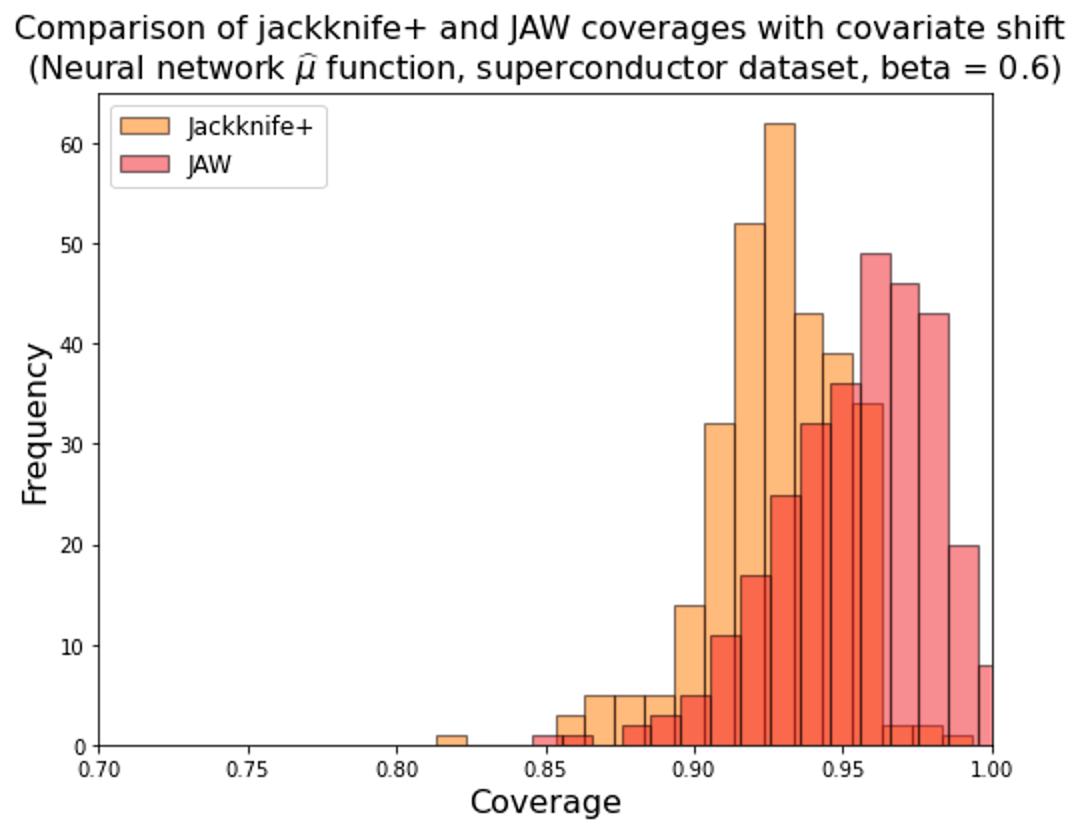} & \includegraphics[width=0.45\textwidth]{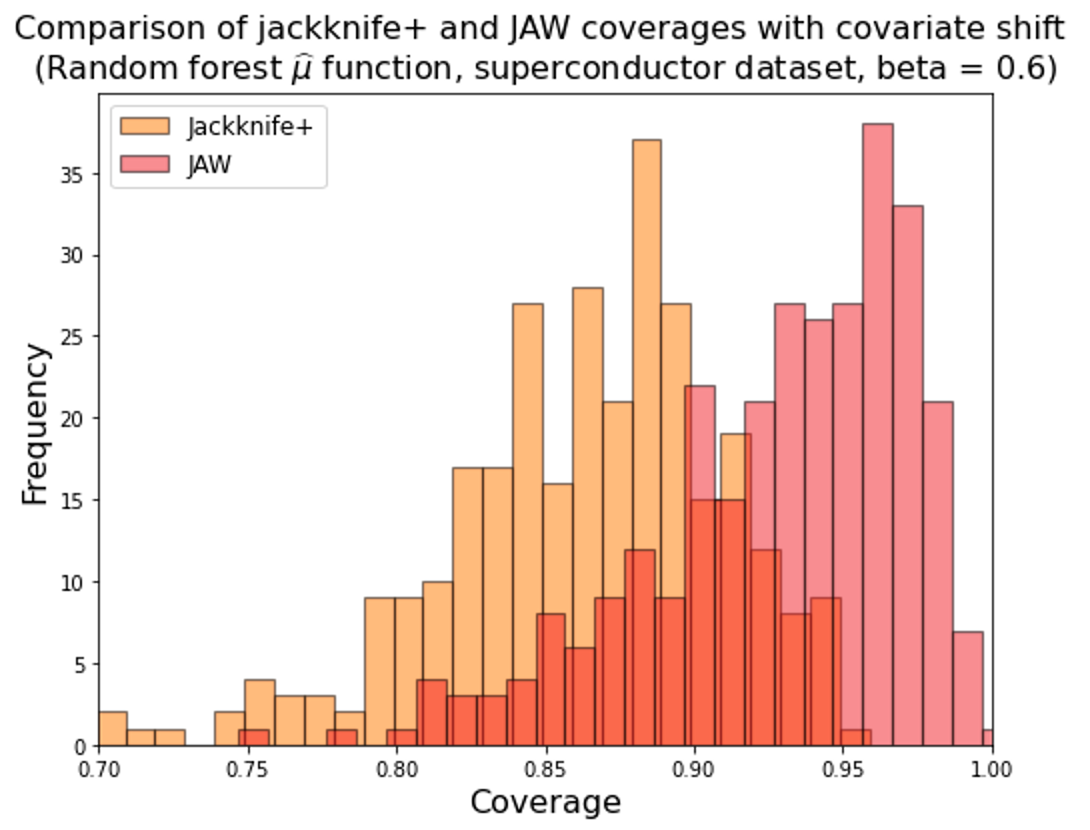}
    \end{tabular}
    \caption{Comparison of the histogram of coverage on Superconductor dataset under covariate shift on the first principal component of the data, with tilting parameter $\beta = 0.6$. JAW still achieves high coverage while jackknife+ loses coverage significantly for the random forest $\widehat{\mu}$ predictor (right). For the neural network $\widehat{\mu}$ predictor (left), jackknife+ does not substantially lose coverage, while JAW has marginally higher coverage, illustrating minimal benefit of JAW over jackknife+ in this case. This is 300 replicates of the experiments.}
    \label{fig:superconduct_hist}
\end{figure}

Although JAW maintains significantly higher coverage than jackknife+ in most conditions, our results suggest that there are some cases when jackknife+ may not lose coverage despite lacking a coverage guarantee for covariate shift. For instance, in Figure \ref{fig:coverage_width} jackknife+ does lose coverage for the random forest $\widehat{\mu}$ predictor, but it does not appear to lose coverage below the target level with the neural network $\widehat{\mu}$ predictor. Figure \ref{fig:superconduct_hist} allows for a closer look at this observation, with the coverage histograms for JAW and jackknife+ on the superconductivity dataset for both random forest and neural network $\widehat{\mu}$ predictors. In Figure \ref{fig:superconduct_hist} there does appear to be a slight loss of coverage for the jackknife+ with neural network $\widehat{\mu}$ predictor, but not as significant of a loss of coverage as with a random forest $\widehat{\mu}$.

\begin{figure}[ht]
 \setlength{\tabcolsep}{1pt}
 \centering
    \includegraphics[width=0.9\textwidth]{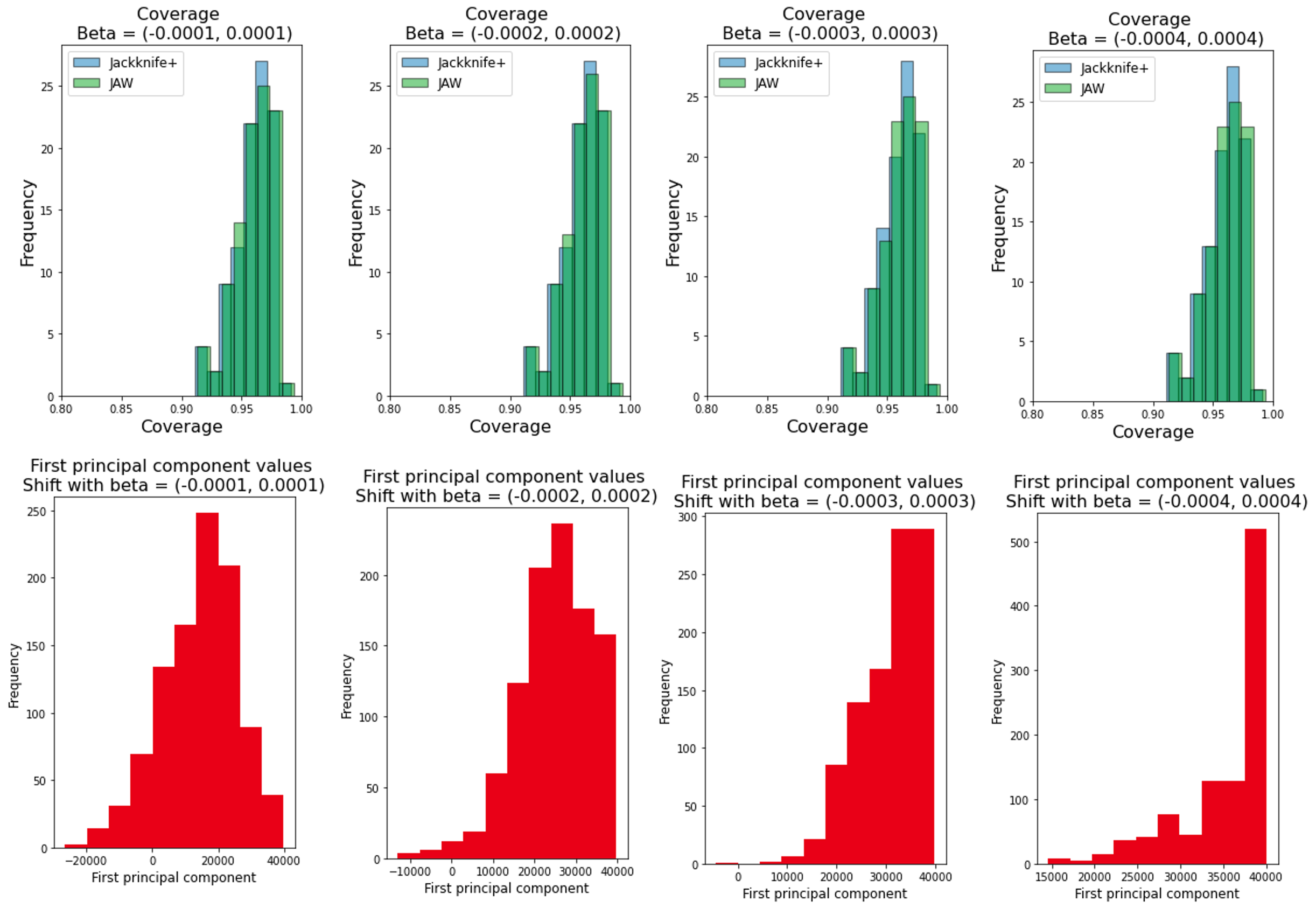}
    \caption{JAW and jackknife+ coverage for different levels of covariate shift levels on the wave energy converters dataset. Each column corresponds to a different level of shift, with increasing shift towards the right. The top row compares JAW (green) and jackknife+ (blue) coverage for a given shift level. The bottom row depicts the first principal component of the data at a given shift level. Neither jackknife+ nor JAW lose coverage at any tested shift level. This is 100 replicates of the experiments.}
    \label{fig:wave_hists}
\end{figure}

A stronger example were jackknife+ appears to not lose coverage under covariate shift is the wave dataset, where JAW and jackknife+ appear to have similar coverage (Figure \ref{fig:coverage_width}). Figure \ref{fig:wave_hists} examines this observation more closely by comparing JAW and jackknife+ coverage histograms corresponding to increasing levels of covariate shift. For the wave dataset, jackknife+ does not seem to lose coverage regardless of the extent of covariate shift.

Though we leave detailed analysis of the conditions that cause jackknife+ to lose coverage or not for future work, we conjecture that jackknife+ loss of coverage may be related covariate shift that makes difficult-to-predict datapoints more likely in the test distribution, and conversely that jackknife+ may not lose coverage when covariate shift does not make difficult-to-predict datapoints more likely in the test distribution. That is, the covariate shift method we use---exponential tilting---causes rare training points to be more common in the test distribution based on the $\beta$ used for tilting, but our conjecture is that the rarity of a datapoint in the training distribution does not necessarily determine how difficult that point is to predict. If rare but easy-to-predict datapoints are made more common due to exponential tilting, then this could explain why jackknife+ does not lose coverage in some cases as in Figure \ref{fig:wave_hists}, though this conjecture requires further investigation.

\section{Code and computational details}
\label{app:details}

\subsection{Code:}
https://github.com/drewprinster/jaws.git

\subsection{Computational details}

All experiments, aside from the runtime comparison described in Appendix \ref{app:runtime} Table \ref{tab:runtime} were performed on an institutional high performance computing cluster using 10 CPUs with a total of 50GB of memory.